\documentclass{article}

\usepackage{rlwrld_report}
\usepackage{algorithm}
\usepackage{algpseudocode}
\usepackage[utf8]{inputenc}
\usepackage[T1]{fontenc}
\usepackage{hyperref}
\usepackage{url}
\usepackage{booktabs}
\usepackage{amsfonts}
\usepackage{nicefrac}
\usepackage{amsmath}
\usepackage{microtype}
\usepackage{lipsum}
\usepackage{duckuments}
\usepackage{adjustbox}
\usepackage{array}
\usepackage{subcaption}
\usepackage{amsmath}
\usepackage{float}
\usepackage{multirow}
\usepackage{enumitem}
\usepackage{listings}
\usepackage{algpseudocode}
\usepackage{bibunits}
\usepackage{changepage}
\usepackage{makecell}
\usepackage{wrapfig}
\usepackage{graphicx}
\usepackage{amssymb}
\usepackage{mathtools}
\usepackage{amsthm}
\usepackage{tocloft}
\usepackage{cleveref}
\usepackage{booktabs}
\usepackage{xspace}
\usepackage{fontawesome5}
\usepackage{pifont}

\usepackage{amsmath,amsfonts,bm}

\def\eqref#1{equation~\ref{#1}}
\def\1{\bm{1}}

\def\rva{{\mathbf{a}}}

\def\rvc{{\mathbf{c}}}

\def\rvh{{\mathbf{h}}}
\def\rvu{{\mathbf{i}}}

\def\rvk{{\mathbf{k}}}
\def\rvl{{\mathbf{l}}}
\def\rvm{{\mathbf{m}}}

\def\rvo{{\mathbf{o}}}
\def\rvp{{\mathbf{p}}}
\def\rvq{{\mathbf{q}}}

\def\rvs{{\mathbf{s}}}

\def\rvu{{\mathbf{u}}}
\def\rvv{{\mathbf{v}}}

\def\rvx{{\mathbf{x}}}
\def\rvy{{\mathbf{y}}}
\def\rvz{{\mathbf{z}}}

\def\rmA{{\mathbf{A}}}

\def\rmC{{\mathbf{C}}}

\def\rmP{{\mathbf{P}}}
\def\rmQ{{\mathbf{Q}}}

\def\rmS{{\mathbf{S}}}

\DeclareMathAlphabet{\mathsfit}{\encodingdefault}{\sfdefault}{m}{sl}
\SetMathAlphabet{\mathsfit}{bold}{\encodingdefault}{\sfdefault}{bx}{n}

\def\gA{{\mathcal{A}}}

\def\gE{{\mathcal{E}}}
\def\gF{{\mathcal{F}}}

\def\gL{{\mathcal{L}}}
\def\gM{{\mathcal{M}}}
\def\gN{{\mathcal{N}}}

\def\gS{{\mathcal{S}}}

\newcommand{\ie}{\textit{i}.\textit{e}., }
\newcommand{\eg}{\textit{e}.\textit{g}., }
\newcommand{\cmark}{\ding{51}}

\definecolor{cornellred}{rgb}{0.7, 0.11, 0.11}
\definecolor{cadmiumgreen}{rgb}{0.0, 0.42, 0.24}
\definecolor{aliceblue}{rgb}{0.91, 0.94, 0.97}
\definecolor{darkblue}{rgb}{0.83, 0.89, 0.97}
\definecolor{Red7}{rgb}{0.941, 0.243, 0.243}
\definecolor{Green7}{RGB}{55, 178, 77}
\definecolor{Blue9}{rgb}{0.098,0.3,0.9}
\definecolor{SJViolet}{RGB}{105,100,171}
\definecolor{SJRed}{RGB}{237,109,107}

\hypersetup{
  colorlinks = true,
  linkcolor  = rlRefLink,   % \ref / \cref / \Cref → rlwrldDarkGreen
  citecolor  = rlHyperlink, % \citep → rlwrldDarkGreen
  urlcolor   = rlHyperlink  % \url / \href → rlwrldDarkGreen
}

\def \name{\textsc{RLDX-1}\xspace}
 
\colorlet{rlTitle}{rlwrldBlack}
\colorlet{rlSectionNum}{rlwrldBlack}
\colorlet{rlHyperlink}{rlwrldDarkGreen}
\colorlet{rlRefLink}{rlwrldDarkGreen}
\colorlet{rlTocLink}{rlwrldBlack}
\colorlet{rlAbsBorder}{rlwrldTextGreen}
\colorlet{rlAbsBg}{rlwrldBgGreenSoft}
\def\abstractstyle{background}
\def\abstractposition{intro}

\def\authordisplay{full}

\def\logotype{woMark}

\title{RLDX-1 Technical Report}

\ifdefstring{\authordisplay}{full}{% 
\author{% 
\normalsize \linespread{1.25}\selectfont 

    \textbf{Dongyoung~Kim}\textsuperscript{1,2\dag\S}%
    \quad \textbf{Huiwon~Jang}\textsuperscript{1,2\dag\S}%
    \quad \textbf{Myungkyu~Koo}\textsuperscript{1,2\dag}%
    \quad \textbf{Suhyeok~Jang}\textsuperscript{1,2\dag}%
    \quad \textbf{Taeyoung~Kim}\textsuperscript{1,2\dag}%
    \quad \textbf{Beomjun~Kim}\textsuperscript{2}%
    \quad \textbf{Byungjun~Yoon}\textsuperscript{2}%
    \quad \textbf{Changsung~Jang}\textsuperscript{1}%
    \quad \textbf{Daewon~Choi}\textsuperscript{2}%
    \quad \textbf{Dongsu~Han}\textsuperscript{2}%
    \quad \textbf{Donguk~Lee}\textsuperscript{1}%
    \quad \textbf{Heeseung~Kwon}\textsuperscript{1}%
    \quad \textbf{Hojin~Jeon}\textsuperscript{2}%
    \quad \textbf{Jaehyun~Kang}\textsuperscript{1}%
    \quad \textbf{Jaekyoung~Bae}\textsuperscript{1}%
    \quad \textbf{Jihyuk~Lee}\textsuperscript{2}%
    \quad \textbf{Jimin~Lee}\textsuperscript{2}%
    \quad \textbf{John~Won}\textsuperscript{2}%
    \quad \textbf{Joonwoo~Ahn}\textsuperscript{1}%
    \quad \textbf{Junhyeong~Park}\textsuperscript{2}%
    \quad \textbf{Junyoung~Sung}\textsuperscript{2}%
    \quad \textbf{Kyungmin~Lee}\textsuperscript{2}%
    \quad \textbf{Minseong~Han}\textsuperscript{1}%
    \quad \textbf{Minsung~Yoon}\textsuperscript{1}%
    \quad \textbf{Sejune~Joo}\textsuperscript{1}%
    \quad \textbf{Seonil~Son}\textsuperscript{1}%
    \quad \textbf{Seungcheol~Park}\textsuperscript{1}%
    \quad \textbf{Seunggeun~Cho}\textsuperscript{2}%
    \quad \textbf{Seungjun~Moon}\textsuperscript{1}%
    \quad \textbf{Seungku~Kim}\textsuperscript{2}%
    \quad \textbf{Yonghoon~Dong}\textsuperscript{2}%
    \quad \textbf{Yongjin~Cho}\textsuperscript{1}%
    \quad \textbf{Youngchan~Kim}\textsuperscript{2}%
    \quad \textbf{Jinwoo~Shin}\textsuperscript{1,2\S}%
    \par
    {\footnotesize
      \textsuperscript{1}RLWRLD
      \quad \textsuperscript{2}KAIST
      \quad \textsuperscript{\dag}Project Leads
      \quad \textsuperscript{\S}Research Leads
      }%
  }%
}{%
  \author{ALINLAB $\times$ RLWRLD}%
}
\begin{document}

\maketitle

\makeatletter
\newcommand{\rlfootertext}[1]{%
  \g@addto@macro\rl@pagefootnotes{%
    \par\noindent #1%
  }%
}
\makeatother

\rlfootertext{%
All project leads contributed equally, and authors with equal contribution are listed alphabetically by first name.\par
Names without additional markers indicate core contributors and additional contributors are listed in Appendix~\ref{sec:contributors}.
}

\vspace{0.2em}

\noindent{\raisebox{-1.5pt}{\includegraphics[height=1.05em]{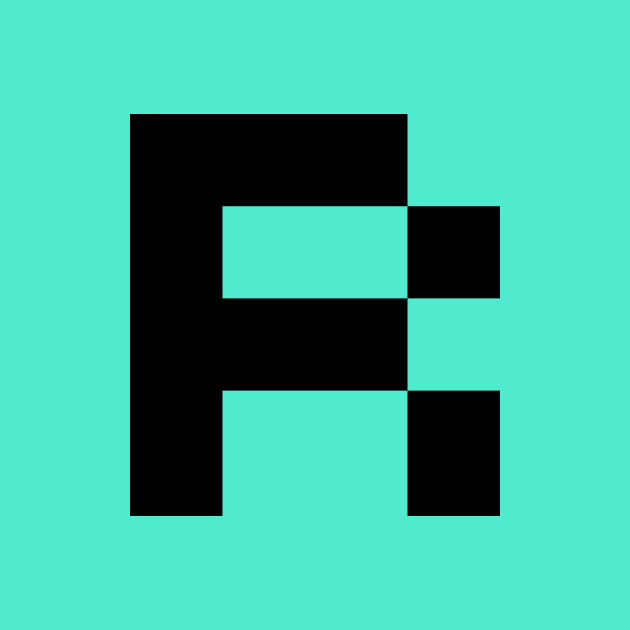}}~\href{http://rlwrld.ai/rldx-1}{\texttt{rlwrld.ai/rldx-1}}}\par

\noindent{\raisebox{-1.5pt}{\includegraphics[height=1.05em]{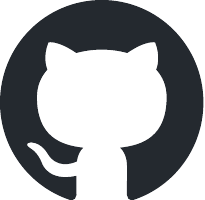}}~\href{https://github.com/RLWRLD/RLDX-1}{\texttt{github.com/RLWRLD/RLDX-1}}}\par%\vspace{0.4em}

\noindent{\raisebox{-1.5pt}{\includegraphics[height=1.05em]{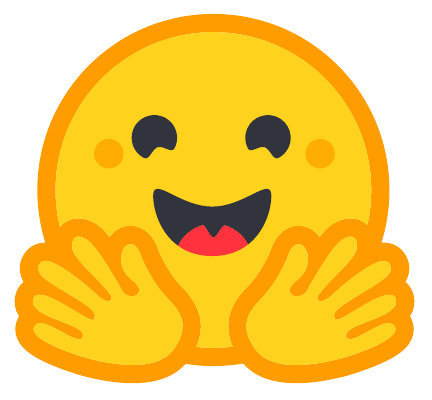}}~\href{https://huggingface.co/collections/RLWRLD/rldx-1}{\texttt{huggingface.co/collections/RLWRLD/rldx-1}}}\par\vspace{0.4em}

\ifdefstring{\abstractposition}{page1}{\begin{abstract}

While Vision-Language-Action models (VLAs) have shown remarkable progress toward human-like generalist robotic policies through the versatile intelligence (\ie broad scene understanding and language-conditioned generalization) inherited from pre-trained Vision-Language Models, they still struggle with complex real-world tasks requiring broader functional capabilities (\eg motion awareness, long-term memory, and physical sensing).
To address this, we introduce RLDX-1, a general-purpose robotic policy for dexterous manipulation built on the Multi-Stream Action Transformer (MSAT), an architecture that unifies these capabilities by integrating heterogeneous modalities through modality-specific streams with cross-modal joint self-attention.
RLDX-1 further combines this architecture with system-level design choices, including data synthesis for rare manipulation scenarios, learning procedures specialized for human-like manipulation, and inference optimizations for real-time deployment.
Through empirical evaluation, we show that RLDX-1 consistently outperforms recent frontier VLAs (\eg $\pi_{0.5}$ and GR00T N1.6) across both simulation benchmarks and real-world tasks that require broad functional capabilities beyond general versatility.
In particular, RLDX-1 shows superiority in ALLEX humanoid tasks by achieving success rates of 86.8\% while $\pi_{0.5}$ and GR00T N1.6 achieve around 40\%, highlighting the ability of RLDX-1 to control a high-DoF humanoid robot under diverse functional demands.
Together, these results position RLDX-1 as a promising step toward reliable VLAs for complex, contact-rich, and dynamic real-world dexterous manipulation.

\end{abstract}}{}

\begin{figure}[H]
  \centering
   \includegraphics[width=\textwidth]{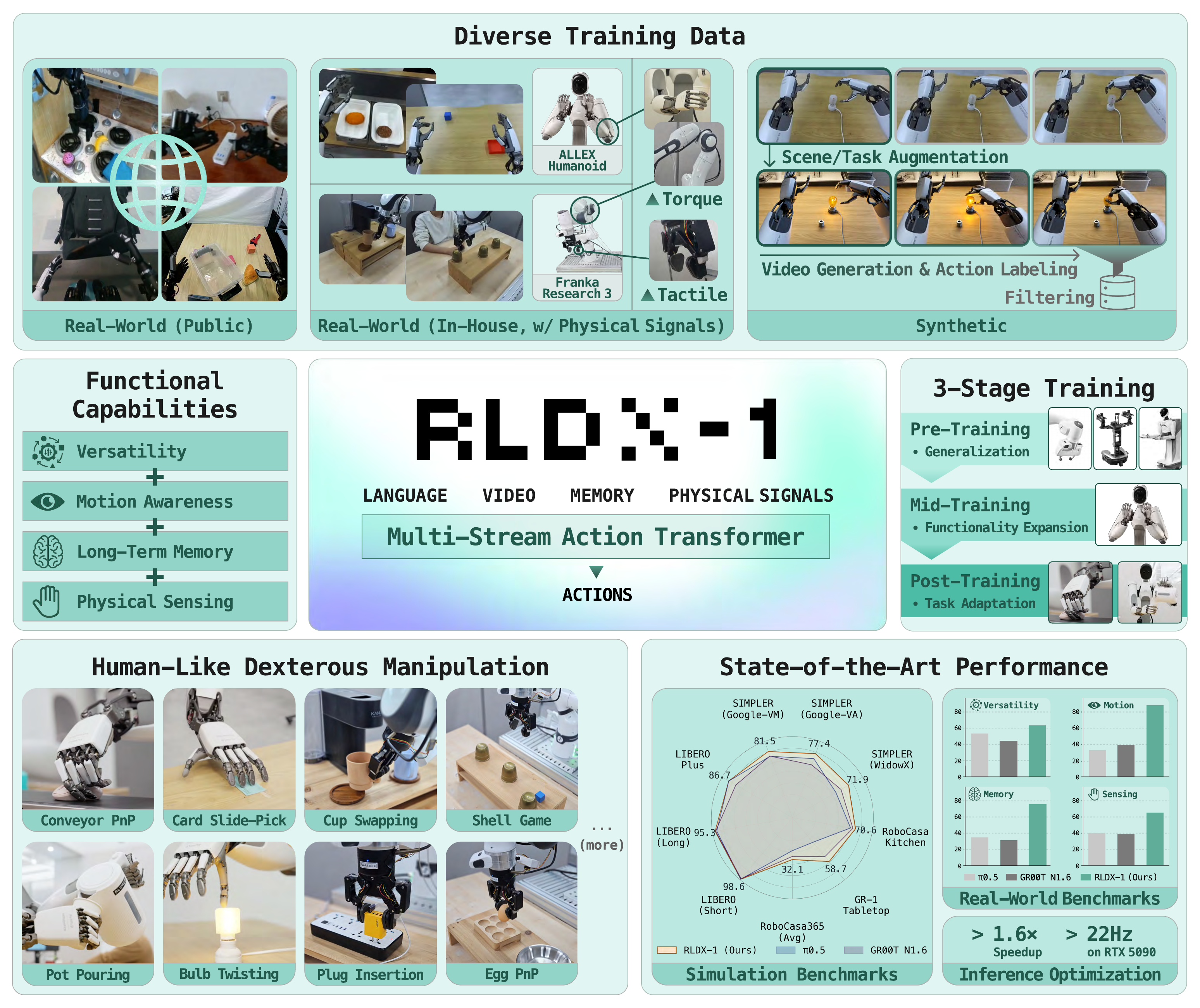}
  \caption{\textbf{Overview of \name.} \name{} is a Vision-Language-Action model (VLA) that integrates diverse functional capabilities for dexterous manipulation in real-world deployment.}
  \label{figure:teaser}
\end{figure}

\newpage

{\hypersetup{linkcolor=rlTocLink}\tableofcontents
\thispagestyle{fancy}}
\newpage

\ifdefstring{\abstractposition}{intro}{\begin{abstract}

While Vision-Language-Action models (VLAs) have shown remarkable progress toward human-like generalist robotic policies through the versatile intelligence (\ie broad scene understanding and language-conditioned generalization) inherited from pre-trained Vision-Language Models, they still struggle with complex real-world tasks requiring broader functional capabilities (\eg motion awareness, long-term memory, and physical sensing).
To address this, we introduce RLDX-1, a general-purpose robotic policy for dexterous manipulation built on the Multi-Stream Action Transformer (MSAT), an architecture that unifies these capabilities by integrating heterogeneous modalities through modality-specific streams with cross-modal joint self-attention.
RLDX-1 further combines this architecture with system-level design choices, including data synthesis for rare manipulation scenarios, learning procedures specialized for human-like manipulation, and inference optimizations for real-time deployment.
Through empirical evaluation, we show that RLDX-1 consistently outperforms recent frontier VLAs (\eg $\pi_{0.5}$ and GR00T N1.6) across both simulation benchmarks and real-world tasks that require broad functional capabilities beyond general versatility.
In particular, RLDX-1 shows superiority in ALLEX humanoid tasks by achieving success rates of 86.8\% while $\pi_{0.5}$ and GR00T N1.6 achieve around 40\%, highlighting the ability of RLDX-1 to control a high-DoF humanoid robot under diverse functional demands.
Together, these results position RLDX-1 as a promising step toward reliable VLAs for complex, contact-rich, and dynamic real-world dexterous manipulation.

\end{abstract}}{}
\section{Introduction}
\label{sec:introduction}

Learning generalist robot policies that achieve human-like dexterous manipulation in real-world environments remains a central goal in robotics.
Existing efforts have mainly focused on versatile intelligence, broadly defined as the ability to understand diverse visual scenes and language instructions, generalize across tasks and environments, and remain robust to unexpected perturbations.
Vision-Language-Action models (VLAs) are a representative framework for this approach~\citep{zitkovich2023rt,kim2024openvla,black2024pi_0,bjorck2025gr00t,team2025gemini,pertsch2025fast}, as they build robot policies on top of Vision-Language Models (VLMs;~\citealt{beyer2024paligemma,chen2025eagle,yang2025qwen3}) with strong world understanding and commonsense reasoning.
However, versatility alone is insufficient for many real-world manipulation tasks, which instead demand a broader range of \emph{functional capabilities} (see Figure~\ref{figure:teaser}).
For instance, in dynamic environments such as manipulation on moving conveyors, existing VLAs struggle to act appropriately, as static visual observations fail to capture object trajectories or temporal dynamics.
Similar limitations extend beyond dynamic environments to tasks that require physical sensing to infer contact forces under occlusion or subtle visual changes, and memory for decisions grounded in prior interactions. 
These observations suggest that human-like dexterous manipulation requires not only versatile intelligence, but also explicit capabilities for motion awareness, long-term memory, and physical sensing.

To address these challenges, \textbf{RLDX-1} combines four key components:
a unified neural architecture integrating diverse functional capabilities;
a synthetic data generation pipeline that augments rare manipulation scenarios via motion-consistency filtering;
a three-stage training procedure bridging internet-scale pre-trained priors with embodiment-specific deployment;
and an inference optimization pipeline that enables real-time control through static graph conversion and operator fusion.
Together, these components enable RLDX-1 to go beyond versatile intelligence toward human-like dexterous manipulation that operates effectively in real-world environments.

\paragraph{Neural Architecture}
Real-world dexterous manipulation requires diverse functional capabilities beyond the versatile intelligence provided by a pre-trained VLM.
We focus on three such capabilities, including motion awareness, long-term memory, and physical sensing, and address each with a tailored architectural module built on top of a standard flow-matching VLA architecture~\citep{black2024pi_0,bjorck2025gr00t}:
\vspace{-0.5em}
\begin{itemize}[leftmargin=*,itemsep=0mm]
    \item {\bf Motion awareness.}
    To operate on dynamic environments, RLDX-1 processes videos with a vision encoder integrated with a motion learning module to capture temporal dynamics effectively~\citep{kim2026exploring}.
    We further compress the past video frames into a single token within intermediate layers of the VLM, allowing the model to efficiently capture temporal context from prior observations~\citep{jang2025contextvla}.
    \item {\bf Long-term memory.}
    To capture long-term historical information beyond short-term multi-frame observations, we employ an explicit memory module for long-term temporal reasoning~\citep{koo2025hamlet}, which maintains a queue of past observation features and integrates them with the current ones to produce memory features for the decoder.
    \item {\bf Physical sensing.}
    To capture contact-rich information that visual observation alone cannot provide (\eg tactile and torque), we feed physical signal inputs into the action module~\citep{lee2026modular}.
    We train this module to predict the future physical sensory signals.
\end{itemize}
\vspace{-0.5em}
To handle the diverse modalities arising from these capabilities, we propose the Multi-Stream Action Transformer (MSAT), an extension of the Multi-Modal Diffusion Transformer (MM-DiT;~\citealt{esser2024scaling,black2024flux}) to action modeling.
MSAT assigns a dedicated stream to each modality and couples them through joint self-attention, allowing each modality to retain its own representation while still contributing to action generation.
Together, these architectural components yield strong performance on tasks where these capabilities are decisive: \eg on catching fast-moving objects on conveyor-belt manipulation, RLDX-1 reaches a success rate of over 87.5\% while $\pi_{0.5}$ remains below 29.2\% (see Section~\ref{sec:exp-real-world-allex} for details).

\paragraph{Training Data}
We use three complementary data sources to train RLDX-1:
(a) large-scale public robot datasets spanning single-arm, dual-arm, and humanoid robots;
(b) in-house demonstrations collected on the ALLEX humanoid\footnote{https://wi-robotics.vercel.app/allex} and a sensor-augmented Franka Research 3 platform (FR3)\footnote{https://franka.de/franka-research-3} that provide tactile and torque supervision absent from public data;
and (c) synthetic data generated by video generative models.
Our synthetic data pipeline augments rare dexterous manipulation scenarios difficult to scale: it increases scene and task diversity by generating scene-augmented frames with off-the-shelf image editing models~\citep{black2024flux} and novel task instructions with VLMs.
It then generates videos with image-to-video models~\citep{nvidia2025cosmospredict2, ali2025world}, optionally diversifies them via video-to-video transfer, and annotates the resulting trajectories with robot actions using an inverse dynamics model~\citep{baker2022video}.
To improve the quality of the generated samples, we further introduce video quality filtering and motion-consistency filtering~\citep{kim2026robocurate}: the former focuses on the quality of generated videos, while the latter focuses on the quality of annotated actions by replaying predicted actions in a simulator and comparing the rollout against the generated video using a learned consistency classifier.
Consequently, the proposed synthetic data results in, \eg improving success rate by 9.1\% on \textit{GR-1 Tabletop} over training on real data alone (see \Cref{sec:ablation-and-analysis} for details).

\paragraph{Training Procedure}
Our training data is structured around three distinct regimes, namely broad multi-embodiment priors, embodiment-specific functional supervision, and task-specific deployment data, each of which requires different optimization signals.
Accordingly, we develop a three-stage training pipeline that progressively specializes the policy from a generalist backbone to a task-specialist deployment model. 
We first pre-train the base model on diverse vision-based embodied data spanning single-arm, dual-arm, and humanoid, equipping temporal modeling capability shared across embodiments and broad embodied action priors.
We then mid-train the model on embodiment-specific data by combining in-house demonstrations with synthetic trajectories.
This stage injects functional capabilities, including motion awareness, long-term memory, and physical sensing, that are absent from public pre-training data, and produces specialized variants for the ALLEX humanoid and FR3.
Finally, we post-train each variant on task-specific data, optionally integrating RECAP-style reinforcement learning~\citep{intelligence2025pi} when needed to further improve success rates on challenging tasks. 
Our training pipeline yields a general pre-trained model together with embodiment-specific variants obtained through mid-training.

\paragraph{Inference Strategy}
In real-robot deployment, high inference latency causes the scene to change between observation and action execution, leading to a mismatch between the observed state and the action moment. 
Off-the-shelf inference stacks leave both graph-level and kernel-level overheads unoptimized. 
Under PyTorch Eager~\citep{paszke2019pytorch}, the resulting per-step latency reaches 71.2 ms for RLDX-1 on an NVIDIA RTX 5090. 
At the graph level, we eliminate launch overhead by converting the model into a static graph, precomputing constant tensors and capturing the entire forward pass as a single CUDA Graph.
At the kernel level, Torch Compile fails to fully exploit cross-operator fusion under the short-prefill execution pattern.
Inspired by state-of-the-art tensor optimization techniques~\citep{park2026trinity}, we design custom kernels for RLDX-1 that fuse critical operator groups and reduce unnecessary memory traffic.
Together, the two stages reduce the per-step latency of the all-modality RLDX-1 to 43.7 ms, achieving a 1.63× speedup.

\paragraph{Evaluation \& Analysis}
For evaluation, we combine diverse simulation benchmarks with real-world manipulation tasks across humanoid and single-arm embodiments.
The simulation benchmarks assess broad VLA capabilities, while the real-world tasks evaluate versatile intelligence and functional capabilities.
As strong baselines, we include recent state-of-the-art VLA models, such as GR00T N1.6 and $\pi_{0.5}$.
For simulation-based evaluation, we consider a broad suite of benchmarks, including conventional benchmarks such as \textit{LIBERO} and \textit{SIMPLER}, robustness benchmarks such as \textit{LIBERO-Plus}, and more challenging evaluation suites such as \textit{RoboCasa Kitchen}, \textit{GR-1 Tabletop}, and \textit{RoboCasa365}.
Across all benchmarks, RLDX-1 consistently outperforms baselines by a significant margin (see Table~\ref{Tab:main_sim}).
Notably, on \textit{GR-1 Tabletop}, RLDX-1 achieves 58.7\%, outperforming GR00T N1.6, which achieves 47.6\%, demonstrating particularly strong performance in humanoid manipulation tasks.
For real-robot experiments, we first evaluate the versatile intelligence of RLDX-1 on an OpenArm humanoid equipped with Inspire Hands, and RLDX-1 consistently outperforms the major baselines.
Specifically, RLDX-1 substantially outperforms $\pi_{0.5}$ in \textit{Unseen Object} (37.5\% to 54.2\%) and \textit{Unseen Task} (45.8\% to 54.2\%) in versatile intelligence tasks.
After that, we evaluate functional capability on the ALLEX humanoid and the Franka Research 3 platform (FR3), including tasks that require motion awareness, long-term memory, and physical sensing, and the performance gap becomes even more pronounced. 
For example, on the ALLEX \textit{Object-in-Box Selection} task, which requires long-term memory, both GR00T N1.6 and $\pi_{0.5}$ achieve success rates in the 30\% range, whereas RLDX-1 achieves a substantially higher success rate of 91.7\%.
These results suggest that existing VLA models remain limited on real-world tasks requiring fine-grained functional capabilities, whereas RLDX-1 effectively addresses these challenges.

\subsection{RLDX-1 Overview}
\begin{figure*}[t]
    \centering
    \includegraphics[width=0.85\textwidth]{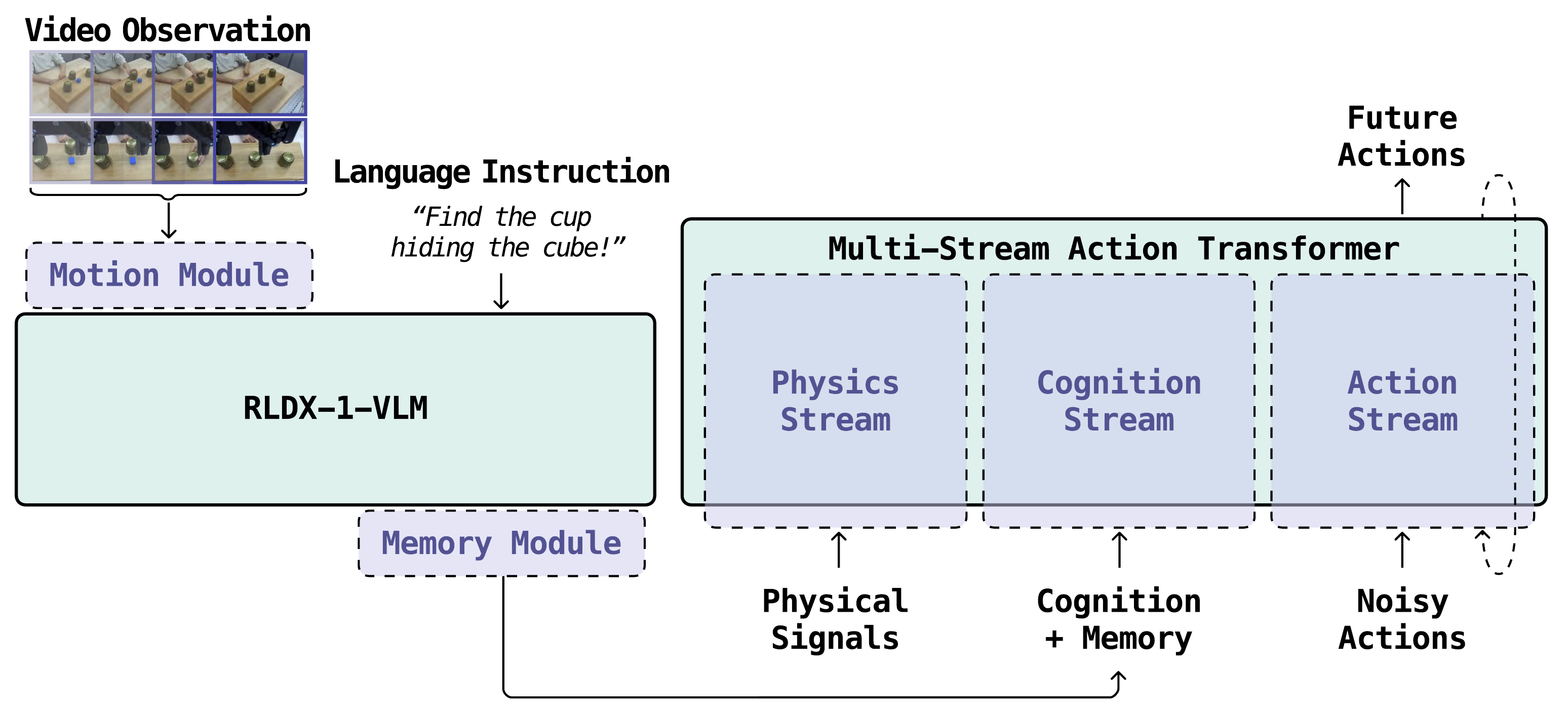}
    \vspace{-0.5em}
    \caption{\textbf{Overview of RLDX-1.} 
    Given video observations and a language instruction, RLDX-1 predicts future actions through three key functionalities: \emph{motion awareness} via the Motion Module, \emph{long-term memory} via the Memory Module, and \emph{physical sensing} via the Physics Stream that ingests torque and tactile signals. 
    A VLM backbone grounds vision and language into a cognition representation, which is jointly denoised with physics and action tokens by the Multi-Stream Action Transformer to produce the final action.
    }
    \label{figure:overview:rldx-model}
\end{figure*}

RLDX-1 is a Vision-Language-Action model (VLA) that integrates diverse functional capabilities for dexterous manipulation in real-world deployment. RLDX-1 covers diverse embodiments including single-arm, dual-arm, and humanoid robots, supporting motion awareness, long-term memory, and perception of physical sensory signals (\eg tactile and torque).
Concretely, given multimodal inputs at the timestep $t$, including language instruction $\rvc_t$, $K+1$-frame video observations $\rvo_{t-K:t}$, proprioceptive state $\rvs_t$, and physical sensory signals $\rvp_{t}$, RLDX-1 generates a sequence of $H+1$ future actions $\rva_{t:t+H}$, \ie an action chunk \citep{zhao2023learning,chi2023diffusionpolicy}.
To integrate these capabilities, RLDX-1 provides a unified framework spanning architecture, data, training, and inference optimization. 
We provide an overview of the RLDX-1 model in \Cref{figure:overview:rldx-model}, and the corresponding sections of the framework below.

\vspace{-0.5em}
\begin{itemize}[leftmargin=*,itemsep=0mm]
\item In \Cref{sec:architecture}, we present the RLDX-1 architecture, consisting of a Vision-Language Model (VLM) augmented with a memory module that encodes video and language into the history-aware cognition features (\Cref{sec:arch-vlm}), and a flow-matching action model that integrates these features with proprioceptive state and physical signals to generate actions (\Cref{sec:arch-policy-module}).

\item In \Cref{sec:Training-data}, we describe the training data for RLDX-1, including public real-world robot datasets spanning diverse embodiments (\Cref{subsec:real-data}) and in-house datasets of the ALLEX humanoid and sensor-augmented Franka Research 3 platform (\Cref{subsec:in-house-datasets}).
We further present synthetic robot datasets generated via our generation pipeline (\Cref{subsec:synthetic-data}).

\item In \Cref{sec:training}, we describe the three-stage training pipeline of RLDX-1. We first pre-train RLDX-1 on a large-scale multi-embodiment dataset to learn general-purpose manipulation and temporal understanding capabilities  (\Cref{sec:training:pretraining}). We then mid-train the model on embodiment-specific datasets to enhance motion awareness, long-term memory, and physical sensing (\Cref{sec:training:midtraining}). Finally, we post-train RLDX-1 for downstream tasks, optionally combined with Adaptive data collection or reinforcement learning when needed  (\Cref{sec:training:posttraining}).

\item In \Cref{sec:inference}, we describe the inference optimization pipelines of RLDX-1 for real-time control. We introduce an inference optimization pipeline based on graph capture (\Cref{subsec:graph-capture-optimization}) and kernel optimization (\Cref{subsec:kernel-optimization}). 
\end{itemize}
\begin{figure*}[t]
    \centering
    \includegraphics[width=\textwidth]{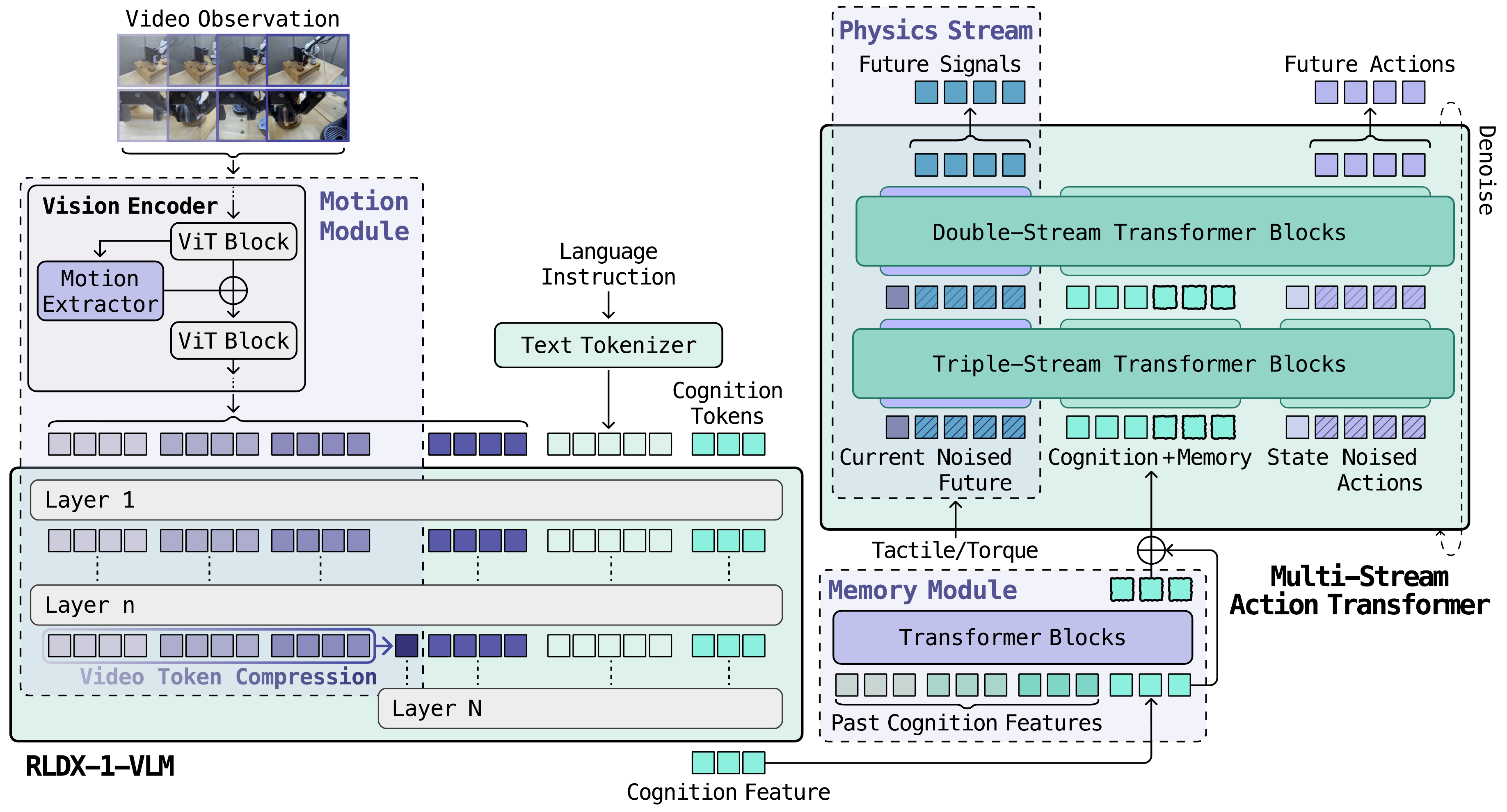}
    \caption{\textbf{Overview of the RLDX-1 architecture.} RLDX-1 consists of two main components: a Vision-Language Model (VLM) and an action model. The VLM takes video observations as input, captures motion-aware visual-language representations, and converts the extracted representations into cognition features that are passed to the action model.
    These cognition features are further augmented with memory features through a memory module.
    The action model then takes the memory-augmented cognition features, physical signals, robot states, and previous actions as inputs to a Multi-Stream Action Transformer (MSAT), which predicts future physical signals and actions for downstream manipulation.}
    \label{figure:overview:architecture}
\end{figure*}

\section{Neural Architecture}
\label{sec:architecture}

In this section, we describe the RLDX-1 architecture, designed to support diverse functionalities by effectively processing heterogeneous inputs.
We describe the two main components: a temporally aware Vision-Language Model (VLM) in \Cref{sec:arch-vlm} and the multimodal action model in \Cref{sec:arch-policy-module}.
We illustrate an overview of the architecture in \Cref{figure:overview:architecture}.

\subsection{Vision-Language Model}
\label{sec:arch-vlm}

The Vision-Language Model (VLM) encodes visual observations and language instruction into action-relevant features for action generation.
By leveraging rich scene understanding and common-sense reasoning, it enables the versatile intelligence capability in RLDX-1.
To make these representations more useful for robotic manipulation, we adapt the VLM with additional robot-related VQA training and introduce cognition tokens to effectively extract action-relevant information for the action decoder.
Then, we further extend the VLM to support additional functional capabilities for real-world manipulation.
In particular, the VLM processes multi-frame observations to better capture temporal dynamics, and we introduce a memory module for long-term reasoning over past observations.

\paragraph{RLDX-1-VLM}
RLDX-1 leverages a pre-trained Vision-Language Model (VLM) to encode video observations and language instructions. 
We build the RLDX-1-VLM upon Qwen3-VL 8B \citep{bai2025qwen3}, a strong open-sourced model offering strong visual perception and multimodal reasoning capabilities.
However, despite its strong performance on general visual reasoning, it often lacks embodied grounding for robot manipulation, including subtask inference for goal completion \citep{chen2025training}, spatial relations between objects in the scene \citep{yuan2025seeing,kim2025robot,jeon2026spatialboost}, and grounding to low-level control actions \citep{kim2025contrastive,kim2026roboalign}. 
To address this, we construct a Visual Question Answering (VQA) dataset tailored to robotic scenarios and fine-tune Qwen3-VL 8B on this dataset.
Specifically, we derive VQA samples from robot trajectory observations that capture three complementary aspects:
(1) spatial relationships between the robot end-effector and target objects to improve spatial reasoning;
(2) intermediate subtasks to enhance task understanding;
and (3) low-level actions associated with the current robot frame to better align the VLM’s understanding with action execution.
The resulting model is used as RLDX-1-VLM.
For action decoding, we use hidden states from an intermediate layer of the model rather than the final LLM layer, since higher layers are typically more specialized for language generation \citep{bjorck2025gr00t}.

\paragraph{Cognition Tokens}
To extract action-relevant representations from the VLM, we introduce cognition tokens $\rvq$, learnable query tokens that are appended to the input token sequence.
Formally, given a video observation $\rvo_{t-K:t}$ and language instruction $\rvl_t$ at timestep $t$, we first process the video observation through a vision encoder $\gE_\theta$ to obtain video features $\rvv_{t}=\gE_\theta(\rvo_{t-K:t})$.
We then use $\rvx=[\rvv_{t},\rvl_t,\rvq]$, \ie a concatenation of $\rvv_t$, $\rvl_t$, and $\rvq$, as input tokens to the VLM backbone $\gF_\theta$.
The output features corresponding to the cognition token are retained as cognition features $\rvh_t$, while the remaining outputs are discarded.
This design allows the cognition tokens to attend to both the visual and linguistic contexts, aggregating information most relevant to downstream action prediction \citep{li2024cogact,pan2025transfer}.
In practice, we use 64 cognition tokens.

\paragraph{Functionality 1: Motion Awareness}
In real-world scenarios, it is essential to perceive diverse dynamic situations, including interactions with moving objects or egocentric camera motions \citep{kaelbling1998planning,zheng2024tracevla,torne2025learning}.
To achieve this, we incorporate multi-frame observations $\rvo_{t-K:t}$ into our VLM and introduce a \emph{motion module} that explicitly models temporal dynamics across frames.
We then extend both the vision encoder and the LLM backbone of RLDX-1-VLM to support temporal reasoning from the multi-frame observations.

First, for the vision encoder, we integrate a module~\citep{kim2026exploring} into its intermediate layers via a residual connection.
This module explicitly captures temporal dynamics by computing space-time self-similarity (STSS; \citealt{kwon2021learning}) of the video features.
Specifically, let $\rvv_t^{(i)}$ denote the video features obtained by processing the video observation $\rvo_{t-K:t}$ through the first $i$ layers of the vision encoder.
Then, the module computes correlations between each spatio-temporal feature of $\rvv_t^{(i)}$ and its local neighbors to obtain a space-time self-similarity tensor $\rmS_t$.
Then, we obtain motion features by processing $\rmS_t$ through the STSS encoder $\gS_\theta$, and use them to residually update the video features as $\tilde{\rvv}_{t}^{(i)} = \rvv_{t}^{(i)} + \gS_\theta(\rmS_t)$.
By integrating motion features, the vision encoder produces motion-aware visual representations through subsequent layers, enabling effective modeling of dynamic changes across frames.
In practice, we integrate the module after the 9th layer of the vision encoder (out of 27 layers), motivated by the observation that physically relevant cues are richly represented at around 30\% depth~\citep{joseph2026interpreting}.

Second, for the LLM backbone, we leverage the temporal reasoning capability while compressing multi-frame observations into a compact representation for efficiency \citep{jang2025contextvla}.
Specifically, in the early layers, we feed multi-frame observation tokens in temporal order and leverage the LLM's causal structure to accumulate temporal context within the current frame and the cognition tokens.
After this, we retain the current frame while compressing past observations into a single context token via average pooling, significantly reducing computational complexity.
In the remaining blocks, we replace the hidden states of past observations with the average pooled context token, which is processed jointly with the hidden states of current observations, language instruction, and cognition tokens, through the blocks.
These modifications enable our model to operate effectively and efficiently in dynamic environments. In practice, we apply the compression after the 4th layer, rather than after the 2nd layer as in \citet{jang2025contextvla}, to use the DeepStack design of Qwen3-VL \citep{bai2025qwen3} without compression, where multi-level vision encoder features are fused into the first 4 LLM layers.

\paragraph{Functionality 2: Long-Term Memory}
While events occurring within a few seconds can be handled using multi-frame observations, sequential and long-horizon tasks often require long-term memory for successful action execution \citep{sridhar2025memer,koo2025hamlet,shi2025memoryvla}.
When short-term visual observations alone are insufficient to reveal the task state or progress, reliable execution may not be possible (\eg the shell game).
To this end, we employ an explicit memory module to enable long-term temporal reasoning \citep{koo2025hamlet}.

The memory module is inserted directly after the RLDX-1-VLM.
To efficiently manage historical observations, we maintain a cache of the most recent $n_\text{mem}$ cognition features stored at an interval of $H+1$ timesteps, where $H+1$ denotes the action chunk horizon.
Formally, given the cognition features of the current timestep $\rvh_t$, we maintain a \emph{memory queue} $\rmQ_t$ that stores the last $n_\text{mem}$ cached cognition tokens, \ie $\rmQ_t = [\rvh_{t-n_\text{mem}H},\cdots,\rvh_{t-2H},\rvh_{t-H}]$.
Then, we obtain a sequence of $n_{\mathrm{mem}}+1$ features by concatenating the memory queue $\rmQ_t$ with the current cognition feature $\rvh_t$, and process it through a Transformer $\gM_\theta$ to obtain the memory feature $\rvm_t$, \ie $\rvm_t=\gM_\theta([\rmQ_t,\rvh_t])$.
Specifically, we use causal attention so that cognition tokens of later timesteps attend only to themselves and earlier ones, preserving temporal ordering.
We feed both the memory features $\rvm_t$ and the original cognition tokens $\rvh_t$ into the action model, enabling the integration of long-term context with current observations for sequential decision making.
In practice, we use a lightweight Transformer module and a memory queue of size $n_\text{mem}=3$.

\subsection{Action Model}
\label{sec:arch-policy-module}

The action model $\gA_\theta$ generates a chunk of $H+1$ future actions $\rva_{t:t+H}$, conditioned on the history-aware cognition feature $\rvh_t$ (and its memory-augmented counterpart $\rvm_t$), the proprioceptive state $\rvs_t$, and, when available, physical sensory signals $\rvp_t$.
We implement the action model as a flow-matching Diffusion Transformer (DiT; \citealt{peebles2023scalable}), where the action model learns a denoising velocity field over action trajectories.
Formally, during training, we sample a denoising timestep $\tau \in [0,1]$ and noise $\bm{\epsilon} \sim \gN(\mathbf{0}, \mathbf{I})$, and construct a noisy action chunk $\rva_{t:t+H}^\tau = \tau\rva_{t:t+H} + (1-\tau)\bm{\epsilon}$.
Given this noisy action chunk $\rva_{t:t+H}^\tau$ and the conditioning inputs $\rvc_t = [\rvh_t,\rvm_t,\rvs_t,\rvp_t]$, we parameterize the action model $\gA_\theta$ through a neural vector field $\rvu_\theta$, which is trained
to predict the velocity that moves the noisy sample toward the clean action chunk, namely $\rva_{t:t+H} - \bm{\epsilon}$, using the following flow-matching objective \citep{lipman2022flow}:
\begin{equation}
\gL(\theta;t,\tau,\bm{\epsilon}) = \left\| \rvu_\theta(\rva_{t:t+H}^{\tau},\tau,\rvc_t) - (\rva_{t:t+H} - \bm{\epsilon})\right\|_2^2.
\end{equation}
During inference, we sample denoising timesteps $\{\tau_i\}_{i=1}^T$ with $0 = \tau_1 < \cdots < \tau_T = 1$ and use Euler's method to generate action chunks over $T$ denoising steps:
\begin{equation}
\rva_{t:t+H}^{\tau_{i+1}}
=
\rva_{t:t+H}^{\tau_i}
+
(\tau_{i+1}-\tau_i)\,
\rvu_\theta(\rva_{t:t+H}^{\tau_i}, \tau_i, \rvc_t),
\quad i = 1, \ldots, T-1.
\end{equation}
where $\rva_{t:t+H}^0 \sim \gN(\mathbf{0}, \mathbf{I})$ is a random noise.
However, the inputs are heterogeneous in both semantics and data scale: cognition features summarize high-dimensional visual–linguistic context, proprioceptive states provide low-dimensional but high-fidelity kinematic information, and tactile and torque signals are data-scarce but essential for contact-rich control.

\paragraph{Multi-Stream Action Transformer (MSAT)}
To handle heterogeneous modality inputs for action generation, we introduce the Multi-Stream Action Transformer (MSAT), an architecture that processes each modality through a dedicated stream while enabling cross-modal interaction via joint self-attention. 
MSAT extends the two-stage (double-then-single-stream) architecture of the Multi-Modal Diffusion Transformer (MM-DiT; \citealt{esser2024scaling,black2024flux}) to action modeling. 
In the early double-stream blocks, we process cognition features $[\rvh_t, \rvm_t]$ through the cognition ($\rmC$) stream and proprioceptive states paired with noisy actions $[\rvs_t, \rva_{t:t+H}^\tau]$ through the action ($\rmA$) stream.
In the subsequent single-stream blocks, we merge the $\rmC$ and $\rmA$ streams into a single sequence for joint processing.
When physical signals are available, MSAT augments this architecture with an additional physics ($\rmP$) stream.
Specifically, we extend the early double-stream blocks into triple-stream blocks over $\rmC$, $\rmA$, and $\rmP$, and the later single-stream blocks into double-stream blocks over the merged $\rmC$-$\rmA$ stream and the $\rmP$ stream.
Within each multi-stream block, every stream applies its own normalization and attention input (QKV) projections.
The resulting queries, keys, and values are then concatenated along the token dimension and processed by joint self-attention, after which the outputs are split back into their corresponding streams, followed by stream-wise residual updates. 
This extensible design enables cross-modal information exchange while preserving modality-specific parameters.
Following \citet{bjorck2025gr00t}, we share the parameters of MSAT across embodiments and use lightweight embodiment-specific projection layers at the MSAT input and output.

\paragraph{Functionality 3: Physical Sensing}
While vision-based observations are sufficient for many manipulation tasks, they are often insufficient for contact-rich tasks, which frequently involve occlusions or subtle task-relevant visual changes, \eg grasping deformable objects, regulating grip force, detecting incipient slip, and inserting a plug \citep{zhang2025ta,su2024roformer,lee2026modular}.
However, physical sensory signals are much scarcer than visual observations and action labels, \eg Franka Research 3 arm equipped with AnySkin tactile sensor \citep{bhirangi2025anyskin} is limited to a small set of internal data (see \Cref{fig:midtrain-data}).
Therefore, to incorporate such signals only when available, RLDX-1 introduces a \emph{physics module} with a decoupled physics ($\rmP$) stream, inspired by the stream-wise physical signal modeling~\citep{lee2026modular}.

The $\rmP$ stream processes physical sensory signals $\rvp_t$ separately from the cognition ($\rmC$) and action ($\rmA$) streams, while allowing cross-modal interaction through joint self-attention.
This allows us to disable the $\rmP$ stream when physical signals are unavailable by masking out the attention operations of the $\rmP$ stream.
In addition, motivated by \citet{zhang2025ta,lee2026modular}, we incorporate an auxiliary objective for predicting $L$ future physical signals $\rvp_{t+1:t+L}$ during training.
Specifically, given current physical signals $\rvp_t$ and future physical signals $\rvp_{t+1:t+L}$, we first sample noises $\bm{\epsilon}_\rvp \sim \gN(\mathbf{0}, \mathbf{I})$, and construct an interpolated noisy future physical signals $\rvp_{t+1:t+L}^\tau = \tau \rvp_{t+1:t+L} + (1-\tau)\bm{\epsilon}_\rvp$.
We then use $\rmP$ stream to predict a velocity field that moves the noisy sample toward the clean future physical signals, namely $\rvp_{t+1:t+L} - \bm{\epsilon}_\rvp$, using the flow-matching objective.
In summary, the action model $\gA_\theta$ jointly denoises two sequences: (a) the noised future action chunk $\rva_{t:t+H}^{\tau}$ in the $\rmC$ stream, and (b) the noised future physical signals $\rvp_{t+1:t+L}^{\tau}$ in the $\rmP$ stream.
This encourages the model to internalize physical interaction dynamics and utilize physical feedback for action generation.

\paragraph{Further Design Choices for MSAT}
We advance the MSAT architecture with three design choices tailored to action generation.
First, we apply rotary positional embeddings (RoPE; \citealt{su2024roformer}) to the action ($\rmA$) stream to capture the relative temporal structure within the action chunk better.
Second, we inject the flow-matching timestep $\tau$ as an in-context token rather than through feature-wise modulation (\eg adaLN; \citealt{peebles2023scalable}).
Specifically, $\tau$ is encoded via sinusoidal embedding and an MLP and prepended to the $\rmA$ sequence as a single token that participates in attention like any other $\rmA$ tokens, allowing the timestep signal to propagate to all modalities through joint self-attention without per-block affine modulation.
Finally, we adopt RMSNorm (including on queries and keys; \citealt{zhang2019root}) and a feed-forward network with SwiGLU activation function \citep{shazeer2020glu}, following common practices in modern Transformer architectures \citep{touvron2023llama,liu2024deepseek,yang2025qwen3,bai2025qwen3}.
\section{Training Data}
\label{sec:Training-data}

In this section, we describe the training data used to develop RLDX-1, which consists of two complementary sources: real-world and synthetic robot data.
\Cref{subsec:real-data} describes the real-world data, which comprises public robot datasets spanning diverse embodiments, and \Cref{subsec:in-house-datasets} presents in-house datasets collected on the ALLEX humanoid and tactile-augmented Franka Research 3 platform (FR3).
To further cover specialized scenarios that are difficult to scale through direct collection, \Cref{subsec:synthetic-data} introduces synthetic robot data produced by our generation and filtering pipeline, which we use to augment the training set.

\paragraph{Data Preprocessing}
We apply a unified visual preprocessing pipeline to all training data.
For training efficiency, we resize each image so that each frame yields at most $64$ vision tokens while preserving its original aspect ratio.
This is enabled by the native-resolution vision encoder of Qwen3-VL~\citep{bai2025qwen3} inherited by RLDX-1, which can process images of arbitrary aspect ratio and resolution without cropping or padding (see~\Cref{appendix:data:image-preprocessing} for more details).

\subsection{Public Real-World Data}
\label{subsec:real-data}

To equip RLDX-1 with a strong action prior across diverse embodiments, we curate public robot manipulation datasets spanning single-arm gripper, dual-arm, and humanoid platforms.
We summarize the detailed decomposition below.
\vspace{-0.5em}
\begin{itemize}[leftmargin=*,itemsep=0mm]
    \item \textbf{Open-X-Embodiment} (OXE;~\citealt{o2024open}) is a widely adopted dataset that aggregates over 1M real-robot trajectories across 20 embodiments. 
    We follow the curated mixture of~\citet{team2024octo} and~\citet{kim2024openvla}, which includes BridgeV2, Fractal, Kuka, and others (see~\Cref{tab:oxe-mixture} for the detailed composition).

    \item \textbf{DROID}~\citep{khazatsky2024droid} provides 92K in-the-wild manipulation trajectories collected on a Franka Research 3 platform (FR3) with a Robotiq gripper, covering 1,417 third-person camera viewpoints with stereo calibration across 564 scenes and 86 tasks.

    \item \textbf{Galaxea Open-World}~\citep{jiang2025galaxea} presents 100K dual-arm manipulation trajectories collected with the Galaxea R1 Lite, a 23-DoF bimanual mobile robot, across 150 task categories in 50 real-world scenes, with subtask-level language annotations.

    \item \textbf{Agibot World}~\citep{bu2025agibot} contribute over 1M trajectories from $100{+}$ homogeneous AgiBot G1 mobile base humanoid robots across 217 tasks and 106 scenes, equipped with parallel grippers, or 6-DoF hands. 
    We sample 275K episodes from this dataset including all hand manipulation episodes.

    \item \textbf{Fourier ActionNet}~\citep{fourier2025actionnet} provides 30K bimanual manipulation trajectories ($\sim$140 hours) collected on Fourier GR-1 and GR-2 humanoids equipped with 6-DoF or 12-DoF dexterous hands, covering tabletop tasks such as pick-and-place, pouring, and insertion via VR-based egocentric teleoperation.

    \item \textbf{Humanoid Everyday}~\citep{zhao2025humanoid} provides 10.3K trajectories across 260 tasks in 7 categories spanning dexterous manipulation, human-humanoid interaction, and loco-manipulation, collected on Unitree H1 and G1 humanoids.

\end{itemize}

\subsection{In-house Real-World Data}
\label{subsec:in-house-datasets}

For embodiment-specific integration and physical modality expansion, we collect in-house datasets on two real-world robot platforms: an in-house Franka Research 3 platform (FR3) setup augmented with tactile and torque sensors, and the ALLEX humanoid platform augmented with torque feedback for high-DoF dexterous manipulation. We describe the detailed hardware specifications of both platforms in~\Cref{figure:hardware_overview}.

\vspace{-0.5em}
\begin{itemize}[leftmargin=*,itemsep=0mm]
    \item \textbf{In-house Franka} Our FR3 setup follows the DROID configuration~\citep{khazatsky2024droid}, consisting of a FR3 arm, a parallel-jaw gripper, a wrist-mounted camera, and third-person cameras. 
    We further augment the gripper with an AnySkin tactile sensor~\citep{bhirangi2025anyskin} and additionally record joint torque measurements.
    We collect data through a teleoperation via a Meta Quest VR controller that commands the end-effector pose.

    \item \textbf{In-house ALLEX} ALLEX is an upper-body humanoid robot designed for human-like dexterous manipulation. 
    It is equipped with 7-DoF arms and 15-DoF five-finger hands.
    A 2-DoF waist extends the robot's workspace, while a 2-DoF neck and stereo egocentric cameras provide a wide perceptual range.
    We further leverage joint torques estimated from motor currents via the current-to-torque constant~\citep{zhang2025ta}.
    We use a multi-device teleoperation system that assigns different control interfaces to different parts of the robot: a Meta Quest VR device controls the head and waist, Vive Trackers capture wrist poses, and inverse kinematics computes the arm joint angles.
    We further capture finger-tip positions using Manus Pro gloves, from which the finger joint angles are computed via inverse kinematics.
\end{itemize}

\begin{figure*}[t]
    \centering
    \includegraphics[width=0.95\textwidth]{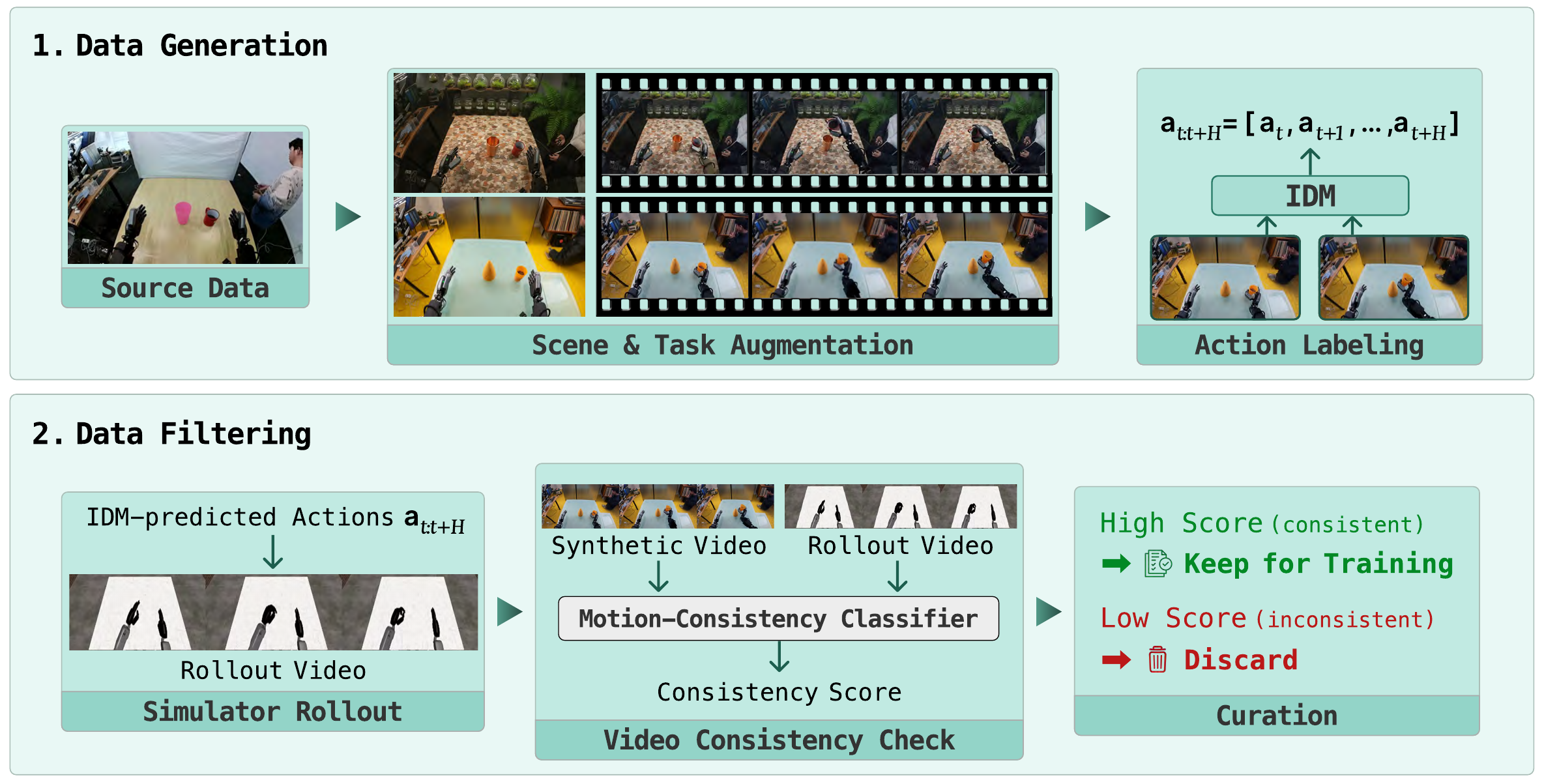}
%    \vspace{-1.0em}
    \caption{\textbf{Overview of the synthetic data framework.} (1) \textit{Data Generation}: a source demonstration is diversified via scene and task augmentation, and an inverse dynamics model (IDM) annotates action labels for the generated videos. (2) \textit{Data Filtering}: IDM-predicted actions are replayed in a simulator, and a motion-consistency classifier compares the rollout against the synthetic video to retain only consistent samples.}
    \label{figure:synthetic-data-overview}
\end{figure*}

\subsection{Synthetic Data}
\label{subsec:synthetic-data}

Generalist robot policies need to address specialized manipulation scenarios, yet scaling such data is challenging due to the need for specialized hardware (\eg ALLEX) or dexterous teleoperation. 
We address this by augmenting specialized robot datasets using video generative models~\citep{yang2024cogvideox, wan2025wan}, to both the public GR-1 humanoid dataset~\citep{fourier2025actionnet} and our in-house ALLEX humanoid dataset. 
Given a source robot video, we use its initial frame and task instruction to generate a new robot video with an image-to-video (I2V) model. 
We then label actions with an inverse dynamics model (IDM)~\citep{baker2022video}, that predicts the action sequence from current and future observations (see \Cref{sec:synth-data-details} for details).
We illustrate an overview of the proposed pipeline in \Cref{figure:synthetic-data-overview}.

Since videos generated directly from source frames and instructions remain too similar to the source trajectories, we augment the data along two complementary axes:
(1) \emph{task instructions} for diversifying intended manipulation behaviors via VLMs,
and (2) \emph{scene visuals} for diversifying initial frames or generated videos through image-to-image (I2I) editing and video-to-video (V2V) transformations. 
We further apply a two-stage filtering pipeline of \emph{video quality filtering} at video level, and \emph{motion-consistency filtering} at action level, removing noisy samples and improving the reliability of the synthetic data.

\begin{figure*}[t]
    \centering
    \includegraphics[width=0.95\textwidth]{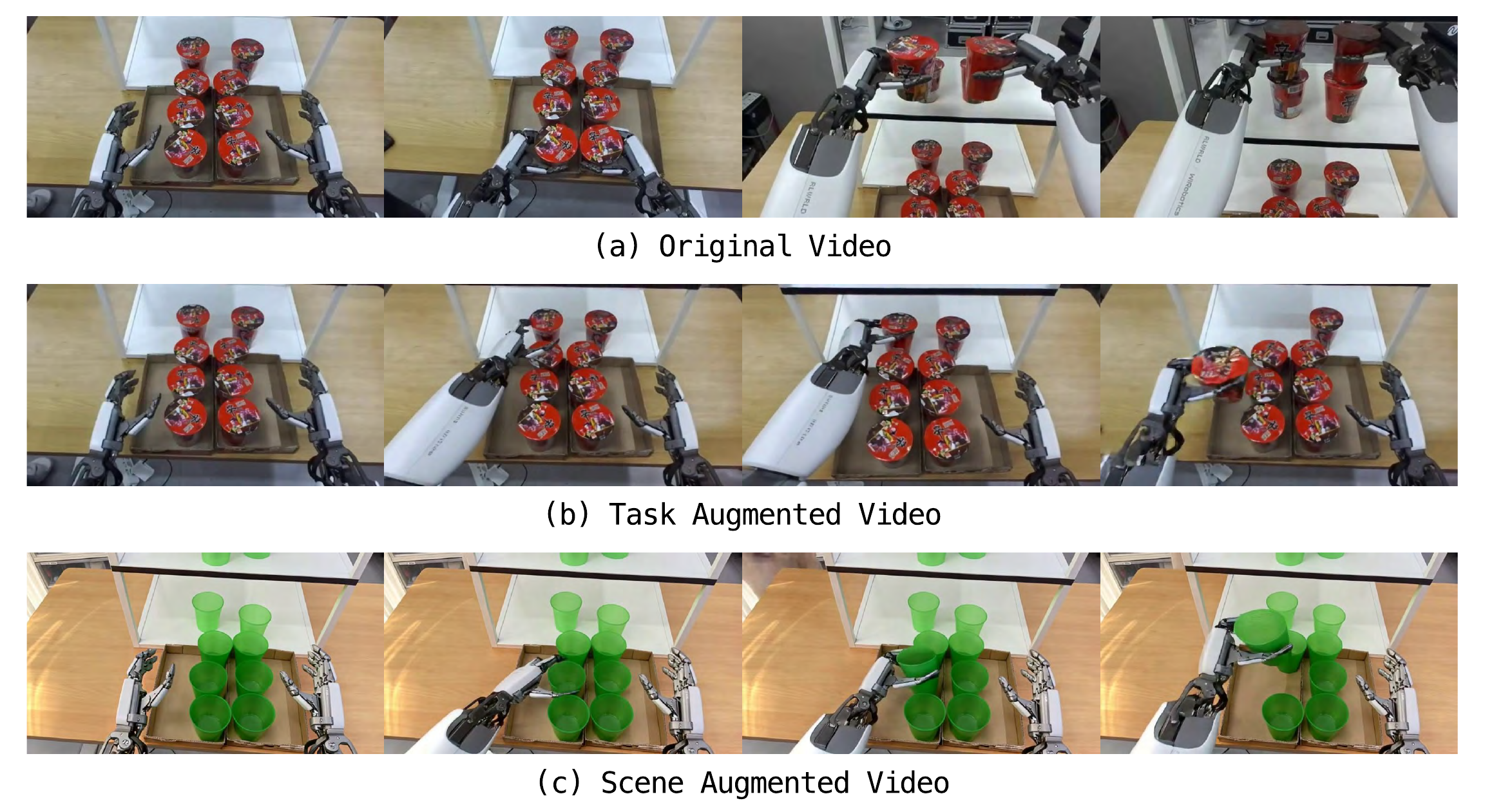}
    \vspace{-1.0em}
    \caption{\textbf{Examples of synthetic data.} We visualize one example of our synthetic data: (a) original in-house ALLEX \textit{stack cup noodles} demonstration, (b) task-augmented variant with a VLM-generated instruction, and (c) scene-augmented variant via I2I editing of the initial frame followed by I2V generation.}
    \vspace{-1.0em}
    \label{figure:synthetic-data-examples}
\end{figure*}

\paragraph{Task Augmentation}
Task augmentation synthesizes executable task instructions conditioned on an initial scene frame using a VLM \citep{bjorck2025gr00t}.
To improve generation quality, we prompt the VLM with few-shot examples drawn from the training data, along with a system prompt that describes the manipulation context.
We use two complementary strategies: (1) \emph{factorized instruction composition} and (2) \emph{skill-primitive-conditioned instruction variation}.
Factorized instruction composition decomposes task instructions into four factors corresponding to \emph{behavior}, \emph{target object}, \emph{placement}, and \emph{hand type}, and recombines them to synthesize plausible yet unseen instructions.
However, when a scene supports only limited manipulation scenarios, such factorized composition may yield infeasible behaviors.
We therefore introduce skill-primitive-conditioned instruction variation, \ie we first extract the underlying skill (such as \textit{pick}, \textit{pour}, or \textit{push}) from the source instruction and use it to condition task generation.
To this end, we leverage the extracted skill to either preserve the original behavior while substituting the target object or location, or replace it with another executable skill drawn from a pre-defined skill set spanning the dataset.
Together, these strategies systematically increase instruction diversity while preserving scene-level feasibility (see \Cref{figure:synthetic-data-examples} (b) for examples).

\paragraph{Scene Augmentation}
To increase scene diversity, we inject visual variation at both the image and video level.
At the image level, we apply I2I editing to the initial frame using FLUX.2-dev~\citep{black2025flux2}, varying four factors: \emph{table appearance}, \emph{target object identity and appearance}, \emph{lighting}, and \emph{background}.
We additionally condition on a Canny edge map for editing to preserve the underlying scene structure and maintain a plausible starting state for video generation \citep{ali2025world}.
These scene variations are then propagated through the subsequent I2V generation process (see \Cref{figure:synthetic-data-examples} (c) for examples).
At the video level, we further apply V2V transfer using Cosmos-Transfer2.5-2B~\citep{ali2025world} to synthetic data to diversify appearance while preserving motion dynamics, thereby maintaining the validity of the annotated actions.
Specifically, we again condition on a Canny edge map to keep object identity and shape unchanged while modifying only texture and color.

\paragraph{Video Quality Filtering}
Video quality filtering uses a VLM to evaluate generated videos along two axes: \emph{instruction following} and \emph{trajectory plausibility}.
For instruction following, we verify whether the robot motion in the generated video aligns with the instruction used for generation.
We first use a VLM to judge whether the generated motion deviates from the original instruction, and then re-caption the deviating videos based on the observed behavior using the VLM \citep{bjorck2025gr00t}.
For trajectory plausibility, we focus on depth perception and spatial orientation, as common manipulation failures include incorrect approach distances of robot hands and implausible object interactions.
In particular, we prompt the VLM to assign a plausibility score from 1 to 5, retaining only videos above a pre-defined threshold.

\paragraph{Motion-Consistency Filtering}
While video quality filtering removes visually implausible or instruction-inconsistent videos, the IDM-predicted action labels can still be misaligned with the motion in the generated video.
To detect such cases, we reformulate action verification as a motion-matching problem between the generated video and a simulator rollout that provides action-consistent reference motion \citep{kim2026robocurate}.
In particular, given a synthetic data sample, we replay the IDM-predicted actions in a simulator to render a rollout video and compare it against the synthetic video.
To perform this comparison, we train a lightweight attentive probe on top of a frozen V-JEPA2~\citep{assran2025v} video encoder (see \Cref{sec:synth-data-details} for details).
The probe consists of a single cross-attention layer with a learnable query token attending to the concatenated embeddings of the two video clips, followed by a linear head that predicts an alignment logit.
We train the probe with positive and negative pairs from available real-world demonstrations to capture fine-grained motion discrepancies.
Specifically, positive pairs consist of a real clip paired with a simulator rollout replayed from its ground-truth action, whereas negative pairs are obtained either by shifting the time window within the same episode or by pairing clips from different episodes that share the same task instruction.
At inference time, the probe filters synthetic samples by retaining those whose alignment probability exceeds a pre-defined threshold, yielding a curated synthetic dataset that is both visually plausible and action-consistent.
\section{Training Procedure}
\label{sec:training}

RLDX-1 is trained in three stages.
\Cref{sec:training:pretraining} describes the first stage, in which we pre-train the model on large-scale multi-embodiment data to learn general action-prediction capabilities.
\Cref{sec:training:midtraining} then presents embodiment-specific mid-training, which develops expert skills and introduces new capabilities beyond pre-training, such as long-term memory and physical sensing.
Finally, \Cref{sec:training:posttraining} details task-specific post-training, which further refines the model for state-of-the-art performance.

\begin{figure}[t]
\centering
\begin{minipage}{0.49\linewidth}
    \centering
    \includegraphics[width=\linewidth]{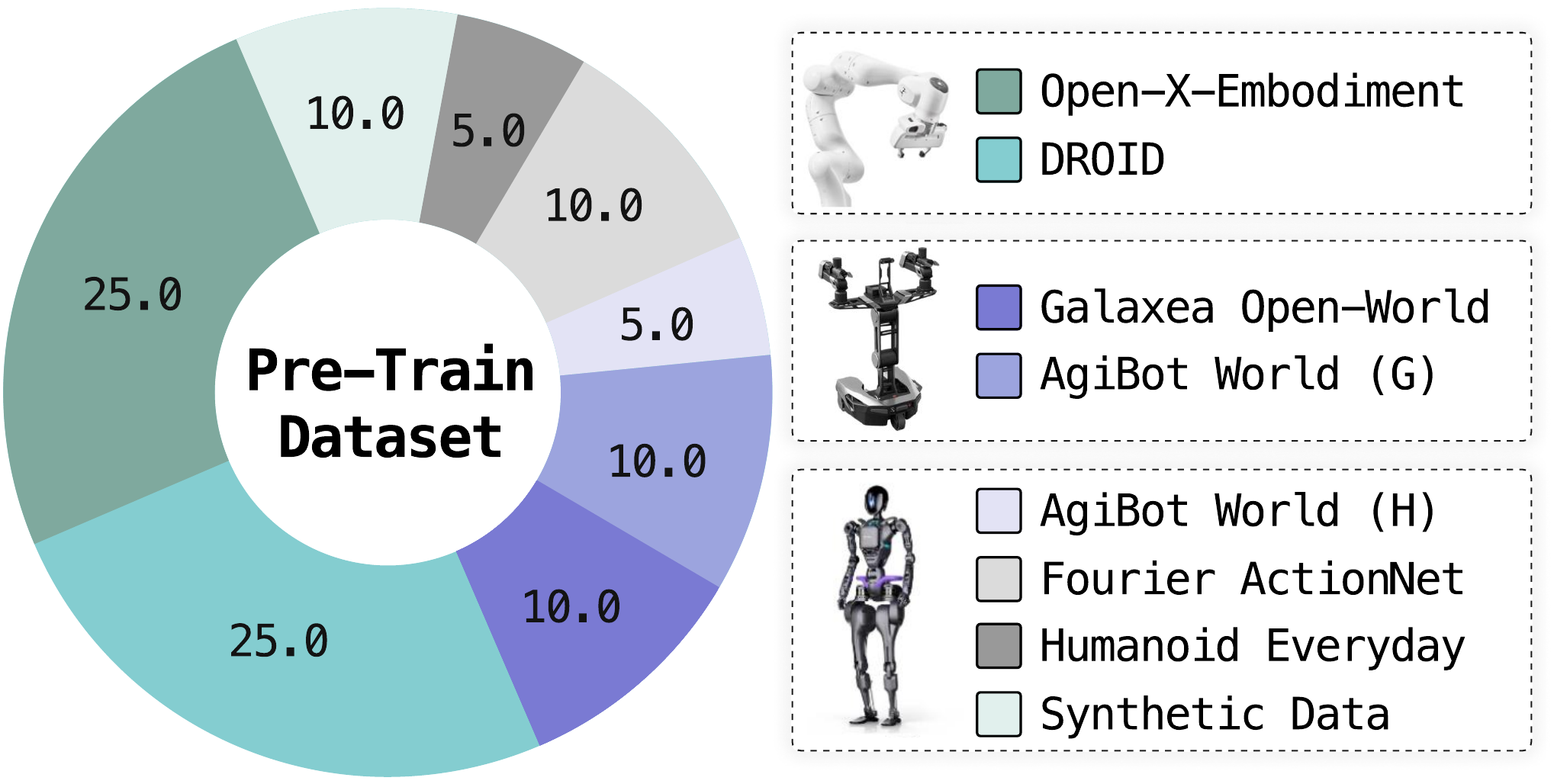}
\end{minipage}%
\hfill
\begin{minipage}{0.478\linewidth}
    \centering
    \resizebox{\linewidth}{!}{%
    \begin{tabular}{lccc}
    \toprule
    \textbf{Dataset} & \textbf{Embodiment} & \textbf{End-Effector} & \textbf{Episodes} \\
    \midrule
    Open-X-Embodiment & Single-arm  & Gripper & \phantom{0,}870K \\
    DROID             & Single-arm  & Gripper & \phantom{0,0}92K \\
    Galaxea Open-World & Dual-arm   & Gripper & \phantom{0,}114K \\
    AgiBot World (G)  & Humanoid    & Gripper & \phantom{0,}239K \\
    AgiBot World (H)  & Humanoid    & Hand    & \phantom{0,0}36K \\
    Fourier ActionNet & Humanoid    & Hand    & \phantom{0,0}30K \\
    Humanoid Everyday & Humanoid    & Hand    & \phantom{0,00}9K \\
    Synthetic Data    & Humanoid    & Hand    & \phantom{0,}150K \\
    \midrule
    \textbf{Total}    &             &         & \textbf{1.5M} \\
    \bottomrule
    \end{tabular}%
    }
\end{minipage}
\vspace{-0.5em}
\caption{
\textbf{Overview of dataset composition for pre-training RLDX-1.} 
RLDX-1 pre-train dataset covers multiple embodiments spanning single-arm grippers, dual-arm grippers, and humanoid platforms equipped with dexterous hands, including synthetic GR-1 humanoid data (\Cref{subsec:synthetic-data}).
}
\label{fig:pretrain-data}
\end{figure}

\subsection{Pre-Training}
\label{sec:training:pretraining}

We first pre-train RLDX-1 on a large-scale, multi-embodiment dataset to learn a general-purpose manipulation policy.
The model takes multi-frame video observations as input and is trained with the flow-matching objective.

\paragraph{Dataset Composition}
The pre-training corpus spans three embodiment categories: single-arm robots with grippers, dual-arm robots with grippers, and humanoids with dexterous hands. 
This heterogeneous mixture covers diverse morphologies, scene environments, and task semantics, encouraging the model to learn embodiment-agnostic representations.
To mitigate the data scarcity of humanoid embodiments, we further incorporate 150K episodes of synthetic GR-1 humanoid data generated by our pipeline (\Cref{subsec:synthetic-data}).
We summarize the overall composition in \Cref{fig:pretrain-data}, with per-dataset details deferred to \Cref{sec:Training-data}.

\paragraph{Generalization to New Embodiments}
RLDX-1 uses embodiment-specific projection layers that map each embodiment's inputs and outputs to a shared latent space, accommodating heterogeneous state and action spaces.
To facilitate adaptation to unseen embodiments, we additionally maintain an embodiment-agnostic projection layer, applied to a small fraction of samples in each batch regardless of source embodiment, providing a strong initialization for downstream fine-tuning.
Inputs to this shared module are zero-padded to a fixed size along the leading dimensions.

\paragraph{Implementation Details}
We use multi-frame video observations consisting of four frames at relative temporal offsets $\{-6, -4, -2, 0\}$ from the current observation.
We normalize proprioceptive states and actions to $[-1, 1]$ using per-dataset $1$st and $99$th percentiles.
We freeze the VLM backbone except for its top four layers.
We pre-train RLDX-1 for $100$K steps with a global batch size of $8192$ using the AdamW optimizer \citep{loshchilov2017decoupled} at a learning rate of $1 \times 10^{-4}$, with a constant schedule preceded by a linear warmup over the first $5\%$ of training.
For each batch, we randomly sample $256$ trajectories and route them through the shared, embodiment-agnostic encoder-decoder.
Pre-training takes approximately $195$ hours on $64$ NVIDIA H200 GPUs.

\begin{figure}[t]
\centering
\small
\centering
\includegraphics[width=0.9\linewidth]{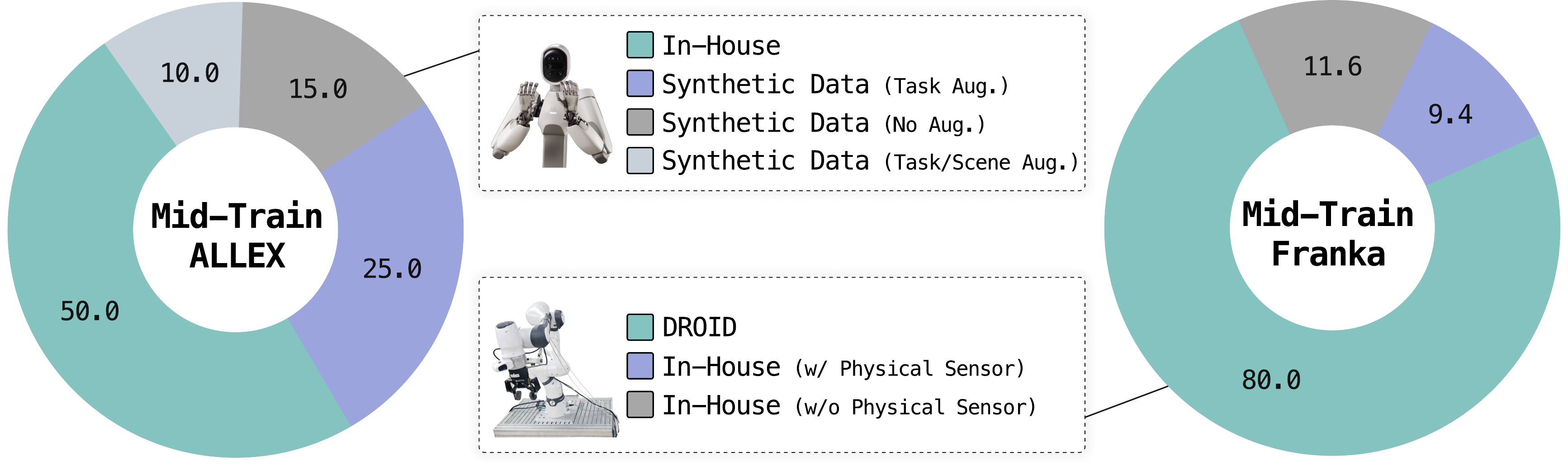}
\caption{\textbf{Overview of data compositions of RLDX-1 mid-training.} 
The mid-training data covers two target platforms: ALLEX and Franka Research 3 platform (FR3). 
The ALLEX composition combines in-house teleoperation data with synthetic data from our generation pipeline (\Cref{subsec:synthetic-data}), while the FR3 composition combines in-house teleoperation data with the public DROID dataset~\citep{khazatsky2024droid}.}
\label{fig:midtrain-data}
\end{figure}

\subsection{Mid-Training}
\label{sec:training:midtraining}

Mid-training has two goals: \emph{embodiment specialization}, which adapts the generalist pre-trained policy into embodiment-expert policies, and \emph{functionality expansion}, which extends the policy 
with three new capabilities corresponding to enhanced motion awareness, long-term memory, and physical sensing.
We perform mid-training on two target platforms: ALLEX, a 48-DoF humanoid, and Franka Research 3 platform (FR3), a single-arm gripper robot.

\paragraph{Dataset Composition}
For ALLEX, we combine in-house teleoperated episodes with $72$K synthetic episodes, sampled at a $5$:$5$ ratio during training to balance the two sources given the scale imbalance. 
For FR3, we combine the $92$K episodes from DROID~\citep{khazatsky2024droid} with in-house teleoperated episodes collected with the target modalities, sampled at an $8$:$2$ ratio. 
In both cases, the in-house data is the sole source of supervision for the newly added memory, torque, and (for FR3) tactile inputs, while the larger-scale source preserves the broad manipulation coverage inherited from pre-training.
We summarize the overall composition used to mid-train RLDX-1 in~\Cref{fig:midtrain-data}.

\paragraph{Functionality Expansion}
We enhance \emph{motion awareness} by integrating the space-time self-similarity (STSS) module into the vision encoder.
We also provide \emph{long-term memory} by attaching the memory module after the VLM, maintaining a queue of the last $n_\text{mem}=3$ cognition features sampled at intervals of $H+1$ timesteps, where $H+1$ matches the action chunk horizon.
We use chunk horizons of $40$ for ALLEX and $16$ for FR3, and the memory module covers temporal windows of $120$ and $48$ past timesteps, respectively.
We incorporate \emph{physical sensing} by predicting the future sensor trajectory over the action chunk horizon (\ie $L=H+1$).
ALLEX uses joint torque feedback, while FR3 uses both joint torque feedback and tactile signals from the AnySkin sensor mounted on the gripper.

\paragraph{Implementation Details}
We mid-train RLDX-1 for $25$K steps with a global batch size of $1024$ using AdamW at a learning rate of $5 \times 10^{-5}$, with the same schedule as pre-training. 
To stabilize the integration of newly introduced modalities, we apply a dropout of $0.3$ independently to each expanded modality input and prepend a $2$K-step alignment warmup during which all pre-trained parameters are frozen and only the newly added modality-specific parameters are updated; afterward, all parameters are jointly trained.
The physics stream's parameters, including its projection layers and attention, are initialized with near-zero output weights.
Mid-training takes $15$ hours on $64$ NVIDIA H200 GPUs.

\subsection{Post-Training}
\label{sec:training:posttraining}
Imitation learning (IL) on embodiment- and task-specific demonstration data is a common practice for real-world deployment, and we adopt this approach in almost all experiments throughout the paper.
However, adapting the model to real-world tasks still requires collecting task-specific data, which is particularly challenging for tasks involving complex embodiments.
For instance, humanoid manipulation requires control over many degrees of freedom, inducing a large action space, making it difficult to learn effective policies from limited demonstrations.
Fine-grained manipulation, such as manipulating complex objects, further demands precise contacts and accurate object interactions, making high-quality demonstrations difficult to obtain.
To address this, we develop an adaptive data collection protocol and further leverage reinforcement learning (RL) to improve deployment performance beyond imitation learning.

\begin{figure}[t]
    \centering
    \includegraphics[width=\linewidth]{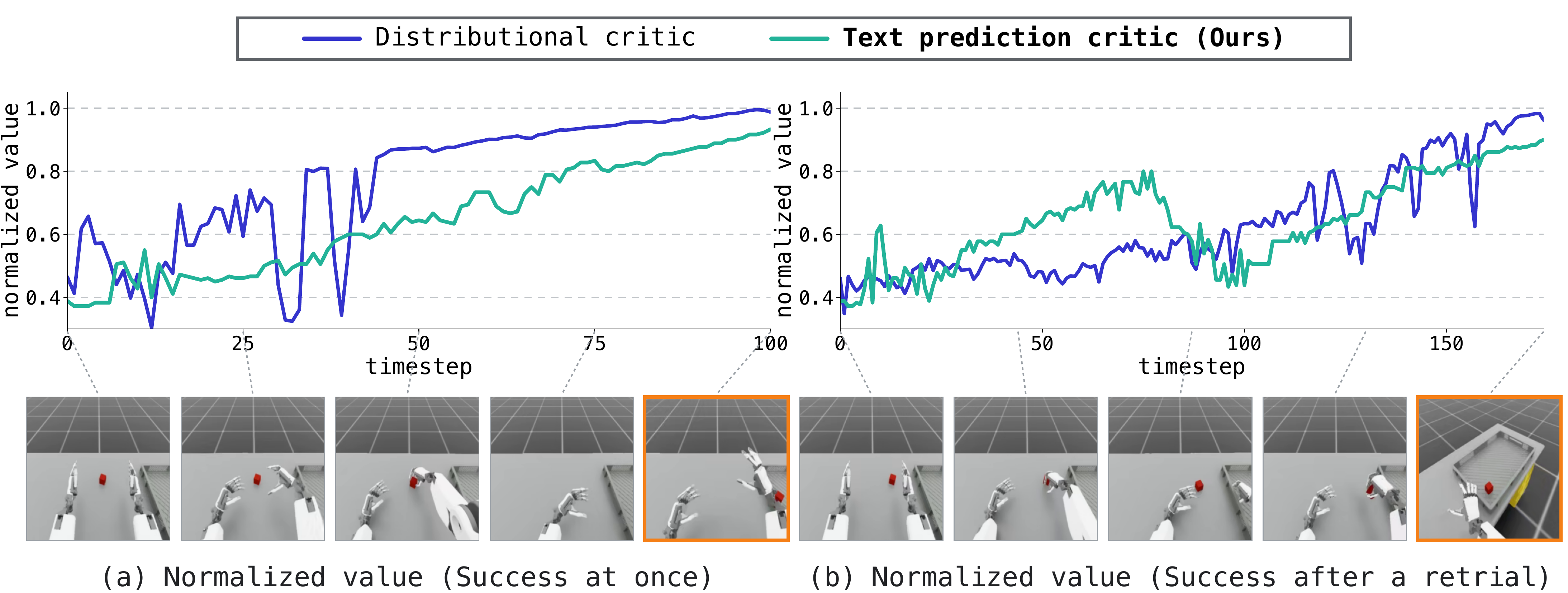}
    \vspace{-0.8em}
    \caption{\textbf{Normalized value over timesteps for a cube pick-and-place task.} The text prediction critic (Ours) better reflects the task progress than the distributional critic: 
    (a) It produces more monotonically increasing values for episodes that succeed on the first attempt, and (b) captures both failure and recovery in episodes that succeed after a retry.}
    \label{figure:rl_value}
\end{figure}

\paragraph{Adaptive Data Collection} To specialize the model for target real-world tasks, we adopt an adaptive data collection protocol consisting of two stages: base data collection and refinement data collection.
The base data collection stage constructs an initial demonstration dataset that balances consistency and task diversity.
Specifically, we first define a teleoperation scenario by decomposing the full task into temporally ordered atomic motion primitives, such as reaching, grasping, moving, placing, and waiting.
For each scenario, we specify consistency factors, which remain fixed across demonstrations to preserve coherent state-action patterns, and variance factors, which are intentionally diversified to improve generalization.
For example, consistency factors can include grasp poses, motion trajectories, and execution order, while variance factors include object poses, object locations, initial configurations, and waiting time during human-robot interaction.
After defining these scenario-level factors, we collect initial demonstrations by sampling along the specified variance axes.
This procedure allows the base dataset to cover task-relevant variations while maintaining a consistent policy structure, improving learnability from limited real-world data.

The refinement data collection stage improves the policy by targeting failure cases observed after training on the base dataset.
We train the model using the base demonstrations, deploy the resulting policy on the target task, and identify conditions under which the policy fails.
For such cases, we expand the scenario definition or variance factors and collect additional demonstrations focused on the observed weaknesses.
For example, if the policy fails to handle objects placed in diverse positions, we increase the spatial coverage of object placements in the refinement dataset.
This process is repeated until the policy reaches the desired level of performance.
By iteratively focusing data collection on failure modes, the proposed protocol enables efficient acquisition of task-specific demonstrations and improves real-world deployment performance.

\begin{figure*}[t]
    \small
    \centering\includegraphics[width=0.95\textwidth]{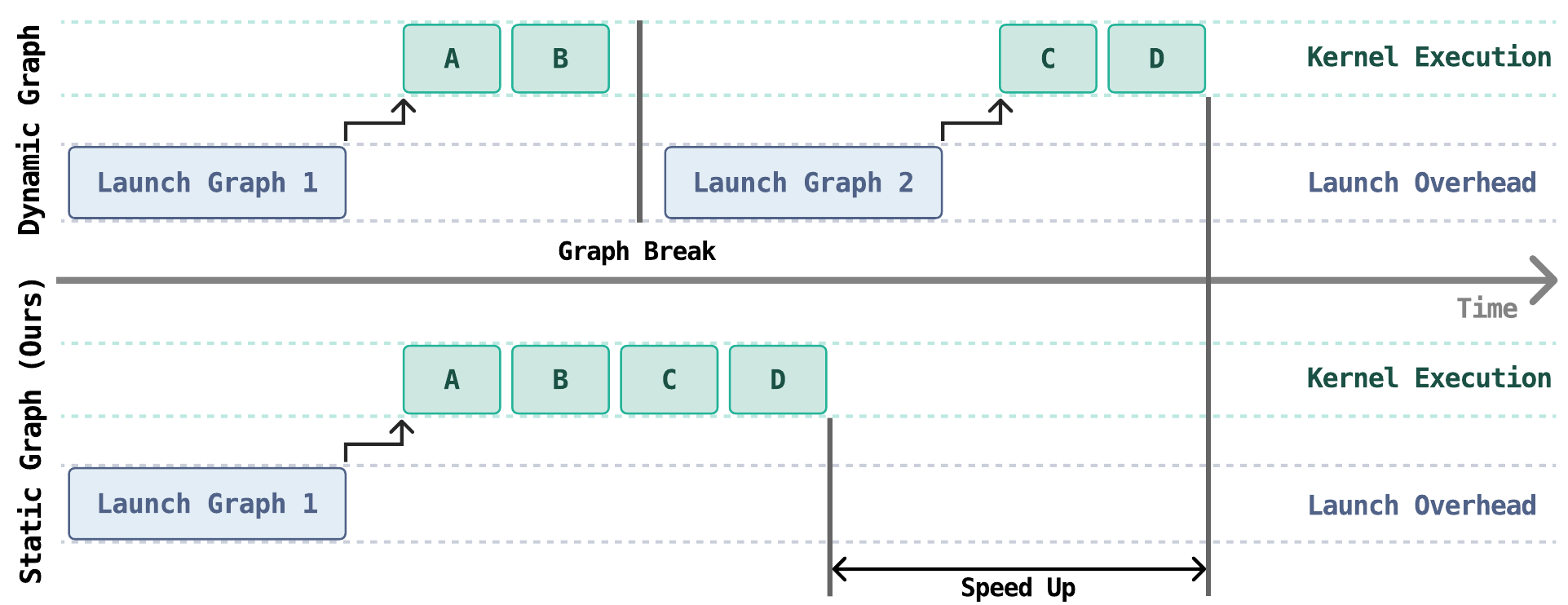}
    \vspace{-0.8em}
    \caption{
    \textbf{Dynamic graph vs.\ Static graph (Ours).} 
    Dynamic graph execution accumulates launch overhead across repeated graph launches. 
    Static graph conversion captures the forward pass as a single CUDA Graph, reducing launch overhead.
    }
    \label{figure:static_graph}
\end{figure*}

\paragraph{Reinforcement Learning (RL)}
To improve RLDX-1 on challenging dexterous manipulation tasks, we introduce an RL-based policy refinement stage that complements imitation learning.
We build our framework on RECAP~\citep{intelligence2025pi}, which decouples critic training from policy optimization to avoid expensive joint optimization and mitigate the instability of RL.
Specifically, we first train a VLM-derived critic to predict values over offline data, and then train the policy with advantage-conditioned supervision derived from the critic.
After that, we iteratively improve both components: at each iteration, we roll out the current policy to collect additional trajectories, merge them into the training dataset, and use the expanded dataset to further improve them.

While RECAP provides an efficient framework for policy optimization in dexterous manipulation settings, learning a reliable critic remains challenging.
To address this issue, we introduce a \emph{text-based} VLM critic that predicts values autoregressively using the native number tokens of the VLM.
Prior VLM critics typically rely on newly initialized prediction heads for value prediction~\citep{tan2025robo,intelligence2025pi,liang2026robometer}, which require large-scale training for adaptation and can suffer from limited transfer for target-domain tasks due to distributional mismatch.
In contrast, our critic avoids introducing a new prediction head and instead reuses the VLM's native text-prediction interface for value prediction.
To be specific, the VLM predicts an unnormalized integer value as a text given the current observation, task instruction, and discretized state.
This design allows the critic to directly leverage the VLM's internal knowledge, enabling reliable value estimation from limited data and efficient adaptation to new target-domain tasks (see~\Cref{figure:rl_value}).
We describe more implementation details, including detailed procedures of RECAP, in \Cref{appendix:RL}.
\section{Inference Strategy}
\label{sec:inference}

Inference latency is critical when deploying on a real robot.
The robot operates in a closed loop of perception, inference, and actuation. 
Each step introduces a delay between capturing an observation and executing the corresponding action on the motors.
Especially in dynamic environments, the scene keeps changing during the delay, and the action for the observation state gets outdated when executed.
The mismatch grows as latency increases.
To reduce the per-step latency, we present an inference optimization pipeline tailored for RLDX-1. 
We optimize at the graph capture and kernel levels.
\Cref{subsec:graph-capture-optimization} removes kernel launch overhead at the graph capture level.
\Cref{subsec:kernel-optimization} further reduces inefficiency at the kernel level.

\subsection{Graph Capture Optimization}
\label{subsec:graph-capture-optimization}

We eliminate the kernel launch overhead that bottlenecks the RLDX-1 forward pass by resolving graph fragmentation through static graph conversion.
Off-the-shelf execution stacks such as PyTorch eager~\citep{paszke2019pytorch} and torch.compile~\citep{ansel2024pytorch} cannot fully utilize CUDA Graphs, so they fail to eliminate the kernel launch overhead entirely.
The reason is that these stacks often fragment the forward pass into multiple subgraphs, preventing end-to-end graph capture.
By converting the model into a static graph, we remove this fragmentation and eliminate the resulting kernel launch overhead.

\paragraph{Kernel Launch Overhead}
Under PyTorch eager and Torch Compile, kernel launch overhead is the main source of the RLDX-1 inference latency.
PyTorch eager launches each operator separately, and the overhead accumulates across the hundreds of operators in the forward pass. 
Torch Compile reduces part of the kernel launch overhead by capturing portions of the RLDX-1 forward pass with CUDA Graphs, but it does not eliminate the overhead entirely.
As a result, the forward pass is split into multiple subgraphs instead of being captured as a single CUDA Graph (Figure~\ref{figure:static_graph}, top).

\paragraph{Graph Fragmentation}
The remaining launch overhead is caused by graph fragmentation during CUDA Graph capture.
In the RLDX-1 forward pass, some parts of the rotary position embedding and attention mask construction still depend on runtime configuration.
Under a fixed deployment setting, these values do not change across inference runs.
However, Torch Compile does not fully resolve these configuration-dependent parts at compile time.
As a result, the forward pass is split into multiple subgraphs instead of being captured as a single CUDA Graph.

\paragraph{Static Graph Conversion}
Static graph conversion reduces this fragmentation by moving configuration-dependent computations out of the forward pass.
We compute the required rotary position embeddings and attention masks in advance and reuse them during execution.
As a result, the forward pass is no longer split into multiple subgraphs and can be captured as a single CUDA Graph.
The forward pass is then launched once per inference step (Figure~\ref{figure:static_graph}, bottom).

\subsection{Kernel Optimization}
\label{subsec:kernel-optimization}

We optimize RLDX-1 inference with hand-designed kernels tailored to its workload.
Each RLDX-1 iteration runs as a short prefill, where latency depends not only on the matmuls but also on how surrounding operators are grouped and executed.
Torch Compile does not fully optimize this workload, because its fixed optimization path misses some useful operator groupings in this setting.
Inspired by state-of-the-art tensor optimization techniques~\citep{park2026trinity}, we hand-design the critical kernels, jointly determining which operators to fuse and how each computation should execute.

\paragraph{Short-Prefill Workload}
RLDX-1 runs every inference iteration as a short prefill.
The full forward pass is executed in one shot, without autoregressive token generation.
Across both the VLM backbone and MSAT, the sequence lengths remain relatively short for an attention-based workload.
In this short-sequence prefill setting, compute-bound matmuls in the VLM backbone and MSAT are interleaved with memory-bound operators (\eg RMSNorm, RoPE, and residual updates).
As a result, end-to-end latency depends not only on the cost of individual operators, but also on how data movement and computation are coordinated across the full operator block.

\begin{figure*}[t]
    \centering
    \includegraphics[width=\textwidth]{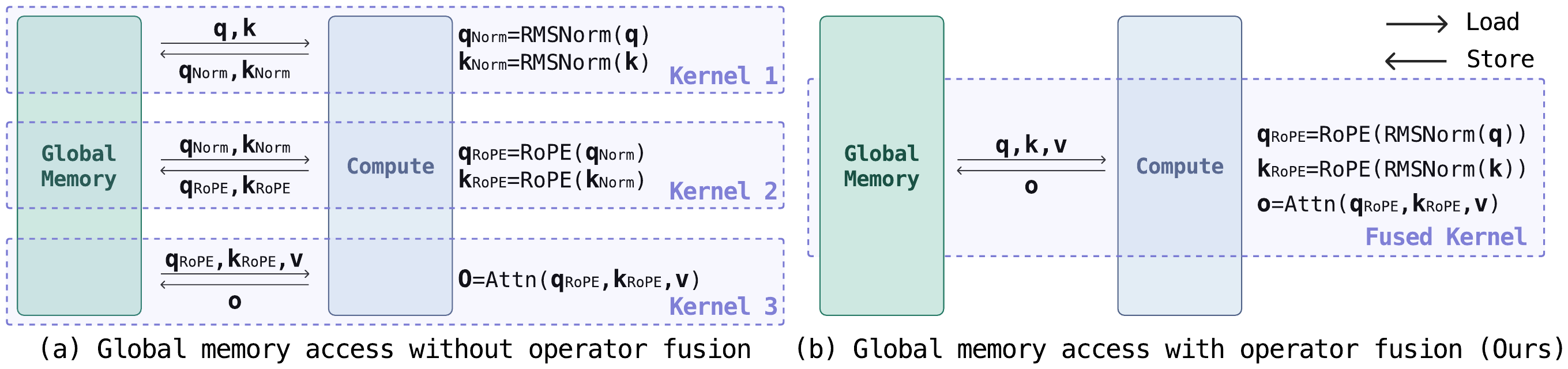}
    \caption{
    \textbf{Effect of operator fusion on memory access.}
    (a) Without fusion, each kernel writes its output to memory and the next kernel reads it back, and these memory round-trips dominate the runtime.
    (b) With fusion, the operators access memory only once for the input load and once for the output store, minimizing memory traffic.
    }
    \label{figure:kernel_fusion}
\end{figure*}

\paragraph{Fixed Fusion Path}
Torch Compile follows a graph-driven fusion strategy, where fusion decisions are determined by the operator graph and compiler pass ordering.
While this approach works for general workloads, it limits the fusion space under the short-prefill execution pattern of RLDX-1, where performance depends on coordinating data movement across operators.
This limitation arises from two factors.
First, the graph-driven fusion strategy exposes only a fixed set of fusion patterns, determined by the structure of the computation graph.
Second, some operators, such as attention, are implemented as external fused kernels (\eg FlashAttention) and are treated as opaque by the compiler.
These kernels introduce hard boundaries in the graph, preventing surrounding operators from being fused across them.
As a result, Torch Compile cannot form cross-operator fusion patterns that are beneficial for this workload.
Figure~\ref{figure:kernel_fusion} (a) shows one such missed opportunity.
When RMSNorm, RoPE, and attention are executed as separate kernels, intermediate Q/K tensors are repeatedly written to and read from global memory, causing data movement to be decoupled from computation.

\paragraph{Workload-Aware Kernel Design}
To address these limitations, we design custom kernels tailored to the RLDX-1 workload.
Building on this principle, our design combines operator fusion with execution-level decisions on on-chip memory reuse and compute ordering, rather than relying solely on Torch Compile.
We profile the Torch Compile-generated implementation using NVIDIA Nsight Compute~\citep{nvidia_nsight_compute} to identify inefficient operator groups.
We replace these groups with manually designed fused kernels while leaving the rest under Torch Compile.
Figure~\ref{figure:kernel_fusion} (b) shows a representative fused implementation.
Fusing these operators keeps intermediate tensors on-chip, allowing data movement and computation to be coordinated within a single kernel, reducing memory traffic and improving execution efficiency.
We implement the selected kernels manually, and we summarize the full list in \Cref{appendix:kernel_opt_list}.
\begin{figure*}[t]
    \centering
    \includegraphics[width=\textwidth]{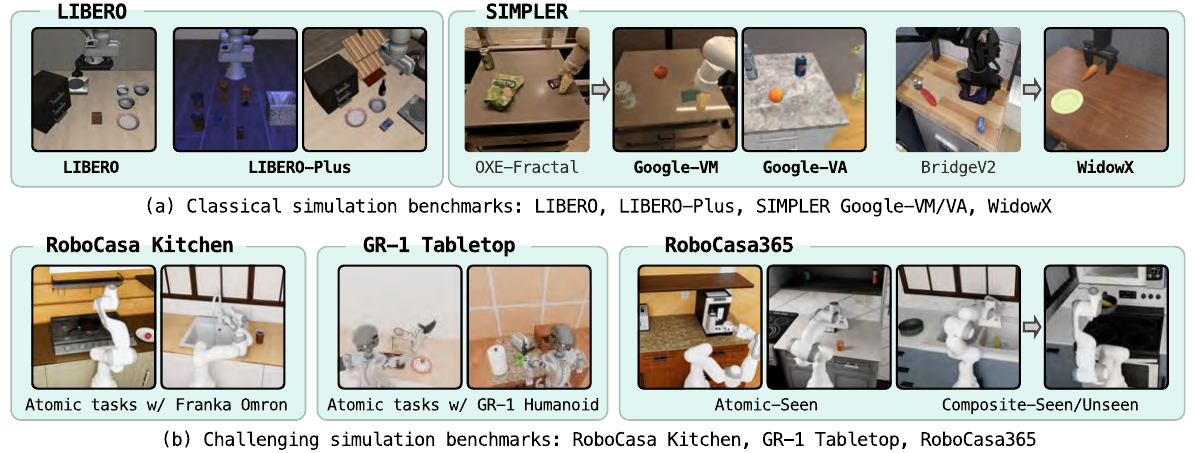}
    \caption{\textbf{Overview of the simulation benchmarks.} (a) We consider established benchmarks, including LIBERO \citep{liu2023libero}, SIMPLER \citep{li2024evaluating} with Google Robot and WidowX for evaluating RLDX-1 on a single-arm robot, and consider LIBERO-Plus \citep{fei2025libero}, SIMPLER Google-VA for evaluating robustness to diverse variations. (b) We further consider more challenging benchmarks, including RoboCasa Kitchen \citep{nasiriany2024robocasa}, GR-1 Tabletop \citep{bjorck2025gr00t}, and RoboCasa365 \citep{nasiriany2026robocasa365} for evaluating a more comprehensive assessment of RLDX-1.
    }
    \label{figure:sim-bench-overview}
\end{figure*}

\section{Evaluation and Analysis}
\label{sec:Evaluation}

In this section, we evaluate the effectiveness of RLDX-1 in various manipulation scenarios.
\Cref{sec:exp-sim} first evaluates the broad capabilities of the pre-trained RLDX-1 model as a Vision-Language-Action model (VLA) across multiple simulated robotic manipulation benchmarks spanning different embodiments, and \Cref{sec:exp-real-world-openarm} extends this evaluation to a real-world OpenArm humanoid benchmark.
\Cref{sec:exp-real-world-allex,sec:exp-real-world-franka} then assess the functional capabilities of the mid-trained RLDX-1 models on real-world benchmarks using the ALLEX humanoid and Franka Research 3 platform (FR3), focusing on tasks that require motion awareness, long-term memory, and physical sensing.
Finally, \Cref{sec:ablation-and-analysis} presents ablation and analysis studies that examine the contributions of key system components, including the architecture, post-training procedure, and inference optimization.

\paragraph{Baselines}
To comprehensively evaluate the performance of RLDX-1, we compare RLDX-1 to recent frontier VLAs, as described in detail below:

\vspace{-0.5em}
\begin{itemize}[leftmargin=*,itemsep=0mm]
    \item \textbf{$\boldsymbol{\pi}_0$-FAST}~\citep{pertsch2025fast} is an autoregressive VLA initialized from PaliGemma 3B VLM~\citep{beyer2024paligemma} that utilizes discrete cosine transform for robot action tokenization.
    
    \item \textbf{$\boldsymbol{\pi}_0$}~\citep{black2024pi_0} is a diffusion-based VLA that pairs PaliGemma 3B VLM with a 300M-parameter action expert through shared self-attention layers, jointly processing multimodal observations and actions via flow-matching.
    
    \item \textbf{$\boldsymbol{\pi}_{0.5}$}~\citep{intelligence2025pi_} extends $\pi_0$ with  Knowledge Insulation~\citep{driess2025knowledge}, which jointly trains the action model on continuous actions via flow-matching and the backbone VLM with discretized action tokens via next-token prediction.
    
    \item
    \textbf{GR00T N1.5}~\citep{nvidia2025gr00t} is a diffusion-based VLA with Eagle 2.5 2B VLM~\citep{chen2025eagle} as backbone, where intermediate VLM hidden states are injected into the action model through cross-attention layers, with a learnable adapter bridging the VLM and the action model.
    
    \item
    \textbf{GR00T N1.6}~\citep{nvidia2025gr00t16} follows the cross-attention design of GR00T N1.5, but alternates between hidden states derived from visual observations and those derived from language instructions. 

\end{itemize}

\subsection{Simulation Experiments}
\label{sec:exp-sim}

To investigate the general capabilities of RLDX-1 as a Vision-Language-Action model (VLA), we evaluate the pre-trained RLDX-1 model on a diverse set of simulated robotic manipulation benchmarks. 
We first consider established benchmarks and their variants, including LIBERO \citep{liu2023libero} for lifelong task execution with a single-arm robot; SIMPLER \citep{li2024evaluating}, with Google Robot and WidowX, for evaluating zero-shot policy transfer under sim-to-real visual domain shifts using a single-arm robot; and LIBERO-Plus \citep{fei2025libero} and SIMPLER Google-VA for robustness under perturbations and variations.
Furthermore, to provide a more rigorous assessment of the model's capabilities, we evaluate RLDX-1 on more challenging benchmarks, including RoboCasa Kitchen \citep{nasiriany2024robocasa} for complex single-arm lifelong manipulation; GR-1 Tabletop \citep{bjorck2025gr00t} for humanoid lifelong manipulation; and RoboCasa365 \citep{nasiriany2026robocasa365} for long-horizon compositional manipulation with a single-arm robot. 
We provide a visualization of each benchmark in \Cref{figure:sim-bench-overview} and describe the details of each benchmark in 
\Cref{appendix:simul_setup}.
We provide full results in \Cref{appendix:full_results}.

\paragraph{Implementation Details}
We evaluate the pre-trained RLDX-1 by fine-tuning it on each benchmark.
Following the pre-training implementation, we freeze the vision encoder and the Large-Language Model (LLM) backbone, except for the top four layers of the LLM backbone.
Unless otherwise specified, we train the model for 60K steps with a global batch size of 1024 using AdamW optimizer \citep{loshchilov2017decoupled} with a learning rate of $1\times10^{-4}$ and a cosine schedule preceded by linear warmup over the first 5\% of training.
We use benchmark-specific settings for three cases: LIBERO (and LIBERO-Plus) uses a global batch size of 256, SIMPLER Google-VM/VA is trained for 20K steps, and RoboCasa365 follows the official implementation with 250K training steps and a global batch size of 196. We describe more details in \Cref{appendix:simul_setup}.

\begin{table}[t]
\centering\small
\caption{\textbf{Results on simulation benchmarks.} We report the success rates (\%) of VLAs on LIBERO \citep{liu2023libero}, LIBERO-Plus \citep{fei2025libero}, SIMPLER  \citep{li2024evaluating}, RoboCasa Kitchen \citep{nasiriany2024robocasa}, GR-1 Tabletop \citep{bjorck2025gr00t}, and RoboCasa365 \citep{nasiriany2026robocasa365} benchmarks, fine-tuned on the training dataset of each benchmark. LIBERO-Short reports the average success rate of the spatial, goal, and object suite.
}
\vspace{-0.8em}
\label{table:simul}
\begin{subtable}{\textwidth}
\centering
\caption{Results on classical simulation benchmarks.}
\vspace{-0.7em}
% \vspace{-0.1in}
%\scalebox{0.95}{%
\begin{tabular}{lccc c ccc}
\toprule
 & \multicolumn{3}{c}{LIBERO} & & \multicolumn{3}{c}{SIMPLER} \\
 \cmidrule(lr){2-4} \cmidrule(lr){6-8}
Method & Short & Long & Avg. & LIBERO-Plus & Google-VM & Google-VA & WidowX \\
\midrule
$\pi_0$-FAST & 93.9 & 60.2 & 85.5 & 64.2 & 61.9 & 59.0 & 48.3 \\
$\pi_0$ & 97.1 & 85.2 & 94.1 & 54.6 & 58.8 & 54.8 & 27.1 \\
$\pi_{0.5}$ & 98.0 & 92.0 & 96.9 & 86.5 & 72.7 & 68.4 & 46.9 \\
GR00T N1.5 & 90.0 & 76.0 & 86.5 & 66.3 & 52.4 & 43.7& 62.0 \\
GR00T N1.6 & 97.4  & 94.4 & 96.7 & 72.6 & 76.1 & 57.1 & 57.1 \\
\midrule
\textbf{RLDX-1 (Ours)} & \textbf{98.6} & \textbf{95.3} & \textbf{97.8} & \textbf{86.7} & \textbf{81.5} & \textbf{77.4} & \textbf{71.9} \\
\bottomrule
\end{tabular}%}
\end{subtable}

\begin{subtable}{\textwidth}
\centering
\vspace{0.05in}
\caption{Results on more challenging simulation benchmarks.}
\vspace{-0.7em}
%\scalebox{0.95}{%
\begin{tabular}{l c c cccc}
\toprule
& & & \multicolumn{4}{c}{RoboCasa365} \\
\cmidrule(lr){4-7}
Method & RoboCasa Kitchen & GR-1 Tabletop & Atomic-S & Comp.-S & Comp.-U & Avg. \\
\midrule
$\pi_0$-FAST & 63.6 & - & 51.7 &  \phantom{0}8.0 &  \phantom{0}1.8 &  21.7 \\
$\pi_0$ & 62.5 & 13.6 & 34.6 & \phantom{0}6.1 & \phantom{0}1.1 & 14.8 \\
$\pi_{0.5}$ & 62.1 & 15.4 & 39.6 & \phantom{0}7.1 & \phantom{0}1.2 & 16.9\\
GR00T N1.5 & 65.7 & 48.0 & 43.0 & \phantom{0}9.6 & \phantom{0}4.4 & 20.0\\
GR00T N1.6 & 66.2 & 47.6 & 61.1 & 12.6 & \phantom{0}2.6 & 26.9 \\
\midrule
\textbf{RLDX-1 (Ours)} & \textbf{70.6} & \textbf{58.7} & \textbf{67.3} & \textbf{19.0} & \phantom{0}\textbf{5.6} & \textbf{32.1} \\
\bottomrule
\end{tabular}%}
\end{subtable}
\label{Tab:main_sim}
\end{table}

\paragraph{Result}
In \Cref{Tab:main_sim}, we find that RLDX-1 consistently outperforms recent frontier models across all simulated robotic manipulation benchmarks.
On LIBERO, RLDX-1 achieves the highest success rate of 97.8\%, demonstrating strong performance on single-arm manipulation tasks.
RLDX-1 also achieves strong results on SIMPLER Google-VM and WidowX, with success rates of 81.5\% and 71.9\%, respectively, demonstrating successful transfer to real-world visual conditions.
Notably, these gains are consistent across robustness benchmarks.
RLDX-1 achieves the best performance on LIBERO-Plus and SIMPLER Google-VA, with success rates of 86.7\% and 77.4\%, respectively.
In contrast, GR00T N1.6 shows substantial performance drops under robustness shifts, decreasing from 96.7\% on LIBERO to 72.6\% on LIBERO-Plus and from 76.1\% on SIMPLER Google-VM to 57.1\% on SIMPLER Google-VA.
This highlights the stronger robustness of RLDX-1.

The advantage of RLDX-1 becomes more pronounced on challenging benchmarks.
On RoboCasa Kitchen, every baseline achieves success rates between 62\% and 67\%, whereas RLDX-1 exceeds 70\%.
The gap is even larger on the GR-1 Tabletop, where RLDX-1 achieves 58.7\% while most baselines remain below 50\%, indicating a clear advantage in humanoid manipulation tasks.
Another notable result is observed on RoboCasa365. While the average scores of baselines range from 14.8\% to 26.9\%, RLDX-1 achieves 32.1\%.
This improvement is particularly pronounced on composite tasks, where the previous state-of-the-art model, GR00T N1.6, remains below 12.6\% on seen tasks and around 2.6\% on unseen tasks. In contrast, RLDX-1 achieves 19.0\% and 5.6\% on seen and unseen composite tasks, respectively.
These results suggest that the performance gap becomes more pronounced as tasks shift from single atomic skills to complex long-horizon behaviors.
Importantly, this comparison highlights the strength of RLDX-1, since GR00T N1.6 is pre-trained with RoboCasa simulation data \citep{nvidia2025gr00t16}, whereas RLDX-1 achieves stronger performance without using any simulation data during pre-training.

\begin{figure*}[t]
    \centering
    \includegraphics[width=0.9\textwidth]{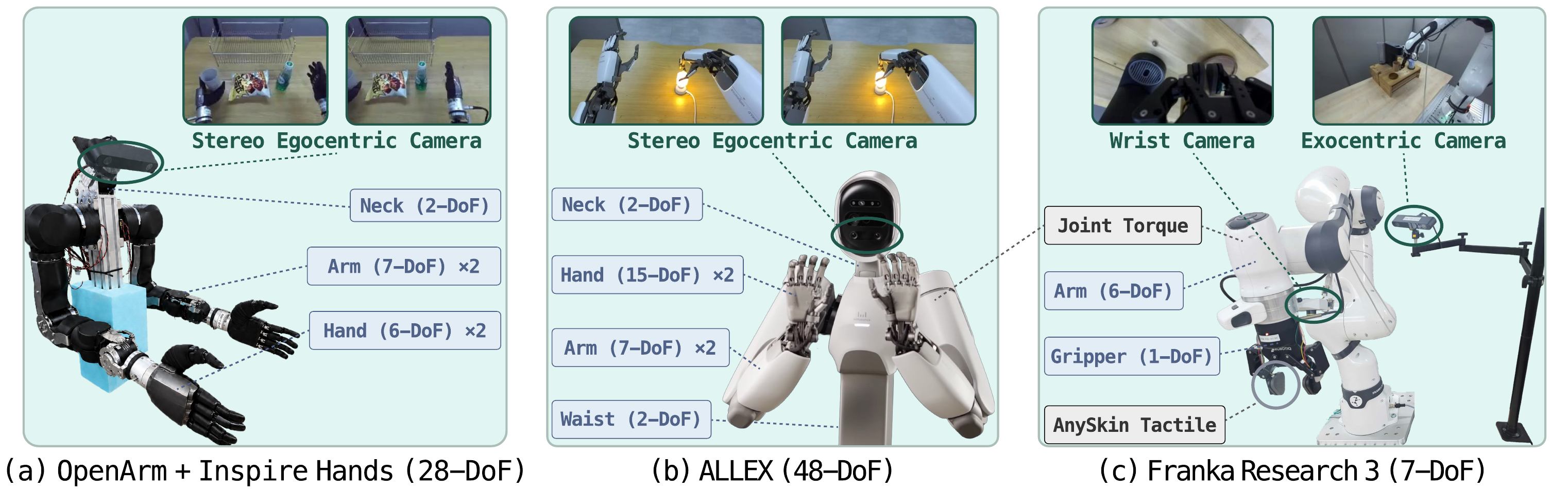}
   % \vspace{-0.8em}
    \caption{
        \textbf{Real-robot platforms.} We use (a) \emph{OpenArm with Inspire RH56F1 Hands}, a 28-DoF upper-body humanoid with stereo egocentric cameras; (b) \emph{ALLEX}, a 48-DoF upper-body humanoid with stereo egocentric cameras; (c) \emph{Franka Research 3 platform (FR3)}, a 7-DoF single-arm robot with an AnySkin tactile sensor, wrist and third-person cameras.
    }
    \label{figure:hardware_overview}
\end{figure*}

\subsection{Real-World Experiments: OpenArm Humanoid}
\label{sec:exp-real-world-openarm}

We evaluate the versatility of RLDX-1 in real-world humanoid manipulation using OpenArm, a whole upper-body humanoid, equipped with 6-DoF Inspire RH56F1 Hands (see \Cref{figure:hardware_overview} (a) for the hardware details).
The benchmark consists of simple and spatially instructed pick-and-place tasks designed to evaluate generalization across object instances, task instructions, and object grounding.
This setup enables us to evaluate basic manipulation ability as well as key aspects of versatile intelligence, including instruction following, object and task understanding, and generalization to unseen variations.
We describe the task details below (see \Cref{fig:openarm_task} for visualization) and provide more details in \Cref{appendix:openarm_details}.

\begin{figure*}[!t]
    \centering
    \includegraphics[width=0.95\linewidth]{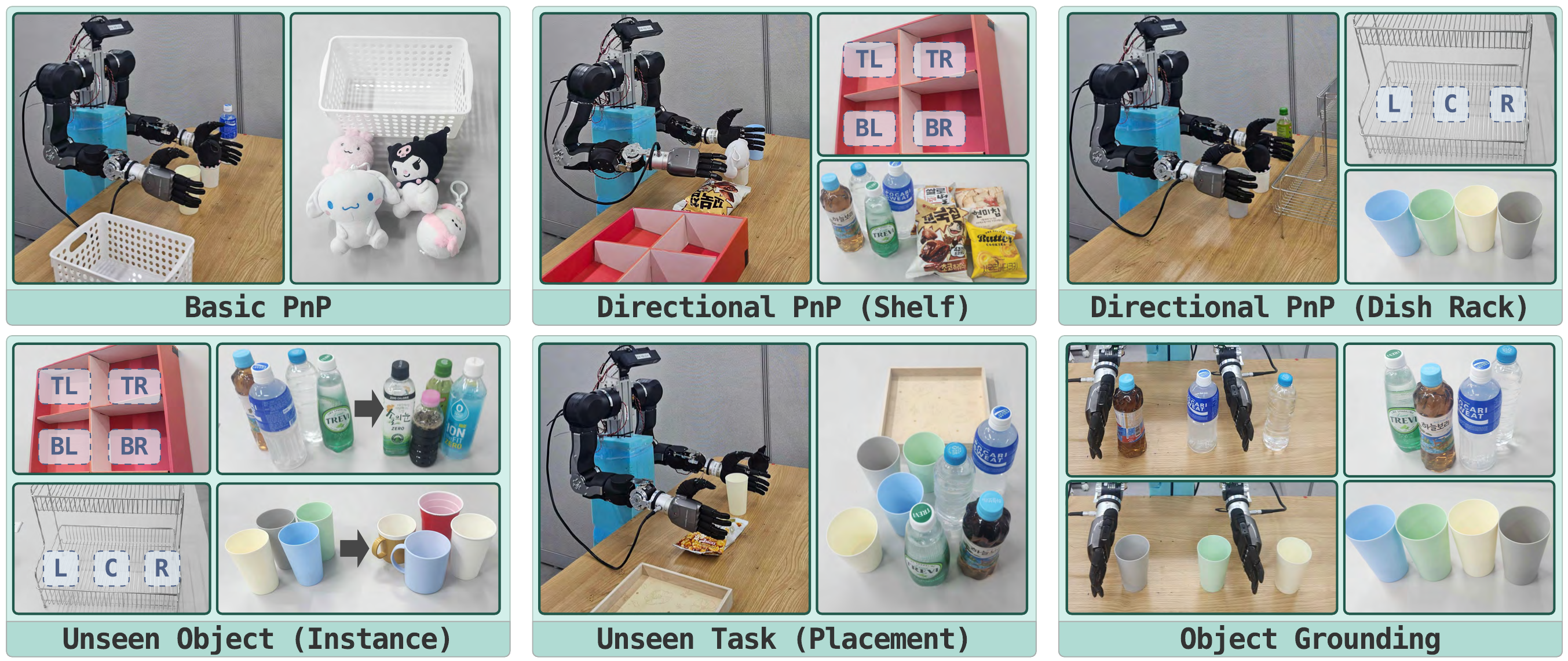}
    \caption{\textbf{OpenArm humanoid benchmark.} We visualize the initial setup for six tasks for evaluating versatility in humanoid manipulation: \textit{Basic PnP}, \textit{Directional PnP (Shelf)}, \textit{Directional PnP (Dish Rack)}, \textit{Unseen Object (Instance)}, \textit{Unseen Task (Placement)}, and \textit{Object Grounding}.}
    \label{fig:openarm_task}
    
    \vspace{0.25em}
    
    \includegraphics[width=0.95\linewidth]{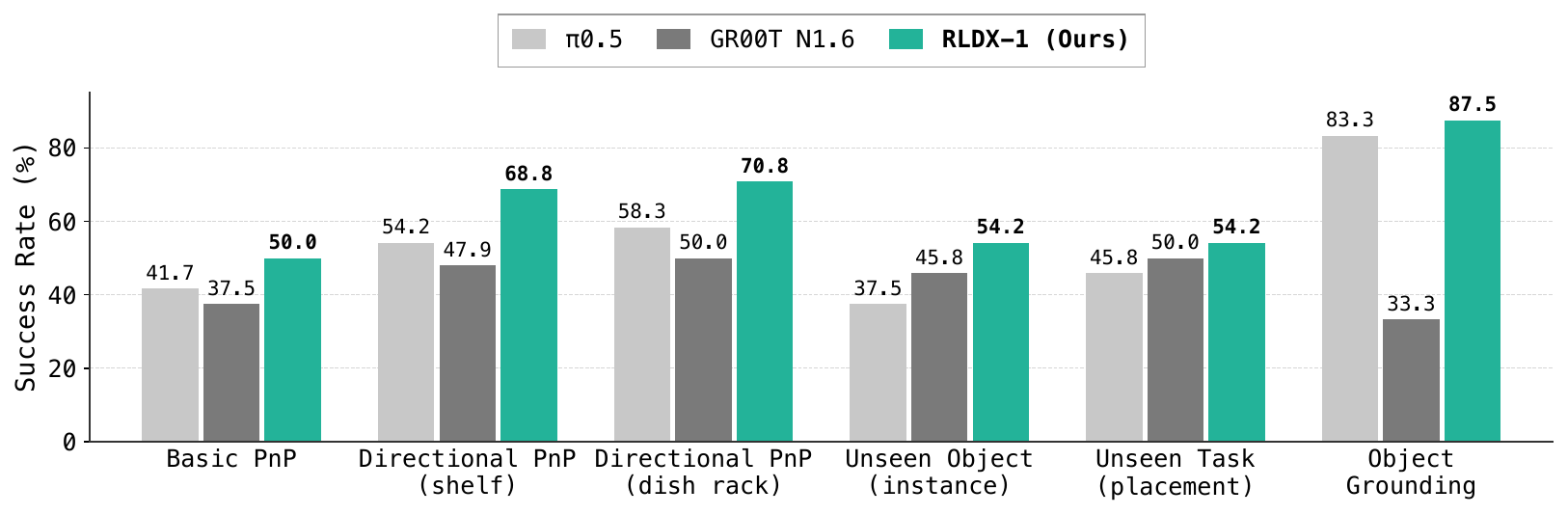}
    \vspace{-0.5em}
    \caption{\textbf{OpenArm humanoid benchmark results.} We report the success rates (\%) of fine-tuned VLAs.
    RLDX-1 substantially improves performance across all tasks, spanning both seen and unseen settings during training.
    }
    \label{fig:openarm_results}
\end{figure*}

\vspace{-0.5em}
\begin{itemize}[leftmargin=*,itemsep=0mm]

    \item \textbf{Basic Pick-and-Place.} 
    This task evaluates basic humanoid manipulation, where the robot identifies the doll among three objects and moves it into a box placed on the right side of the robot.
    The task requires target-category recognition, single-hand manipulation, and bimanual coordination when the object and target place lie on opposite sides of the workspace.
    \item \textbf{Directional Pick-and-Place (Shelf / Dish Rack).}
    This task evaluates instruction-guided humanoid manipulation. 
    The robot identifies the target object among distractors and moves it into one of several target locations:
    in the \emph{Shelf} variant, it picks the snack or bottle and places it into one of four shelf locations (top-right, top-left, bottom-right, bottom-left);
    in the \emph{Dish Rack} variant, it picks the cup and places it into one of three dish rack slots (left, center, right).
    Both variants require spatial instruction, in addition to the capabilities required for Basic Pick-and-Place.
    
    \item \textbf{Unseen Object (Instance).} 
    This task evaluates generalization to unseen object instances in humanoid manipulation.
    The robot performs the Directional Pick-and-Place tasks with novel bottles and cups unseen during training, requiring object-instance generalization in addition to the capabilities required for Directional Pick-and-Place.

    \item \textbf{Unseen Task (Placement).} 
    This task evaluates generalization to an unseen task configuration in humanoid manipulation.
    The robot performs Basic Pick-and-Place with bottles and cups, despite training demonstrations containing only dolls.
    The task also uses a novel target receptacle that is not included in the training demonstrations.
    This task requires task generalization to new objects and placing positions, in addition to the capabilities required for Basic Pick-and-Place.

    \item \textbf{Object Grounding.} 
    This task evaluates fine-grained object grounding in humanoid manipulation.
    The robot must identify and pick the correct object instance specified by the language instruction among three objects from the same category.
    Since this task is absent from the training demonstrations, it requires both instance-level object grounding and task generalization.
\end{itemize}

\paragraph{Evaluation Protocol}
We conduct each evaluation trial in a three-object tabletop scene consisting of one target object and two distractors, and vary the initial object layout across three predefined configurations.
For all tasks except \textit{Object Grounding}, we count a trial as successful if the robot identifies and grasps the correct target object and places it at the target location specified in the language instruction.
For \textit{Object Grounding}, we count a trial as successful once the robot identifies and grasps the object instance specified in the language instruction.

\paragraph{Results}
As shown in \Cref{fig:openarm_results}, RLDX-1 achieves the best overall performance across all benchmark settings.
On \textit{Basic Pick-and-Place}, despite the challenge of grasping a deformable object with limited training data, RLDX-1 reaches 50\% success, outperforming $\pi_{0.5}$ (41.7\%) and GR00T N1.6 (37.5\%).
The gap widens when instruction following is required: RLDX-1 surpasses the strongest baseline $\pi_{0.5}$ by 14.6\% on \textit{Directional Pick-and-Place (Shelf)} and 12.5\% on \textit{(Dish Rack)}, indicating stronger instruction-following capability in addition to basic manipulation.
The gains also persist on \textit{Unseen Object}, \textit{Unseen Task}, and \textit{Object Grounding}, showing that RLDX-1 generalizes beyond the training distribution.
We also identify distinct failure modes in the two baselines. 
$\pi_{0.5}$ degrades significantly on unseen settings (\eg 37.5\% vs.\ RLDX-1's 54.2\% on \textit{Unseen Object}) and is the only baseline that frequently becomes stuck during unseen tasks, likely due to its weaker VLM backbone and full VLM fine-tuning, which encourages overfitting.
GR00T N1.6, in contrast, recognizes object categories but struggles with instance-level grounding, scoring only 33.3\% (equivalent to random selection) on \textit{Object Grounding} versus RLDX-1's 87.5\%.
RLDX-1 avoids both failure modes and improves consistently across all settings, demonstrating stronger generalization for real-world humanoid manipulation.

\subsection{Real-World Experiments: ALLEX Humanoid}
\label{sec:exp-real-world-allex}

\begin{figure*}[t]
    \centering
    \includegraphics[width=0.95\textwidth]{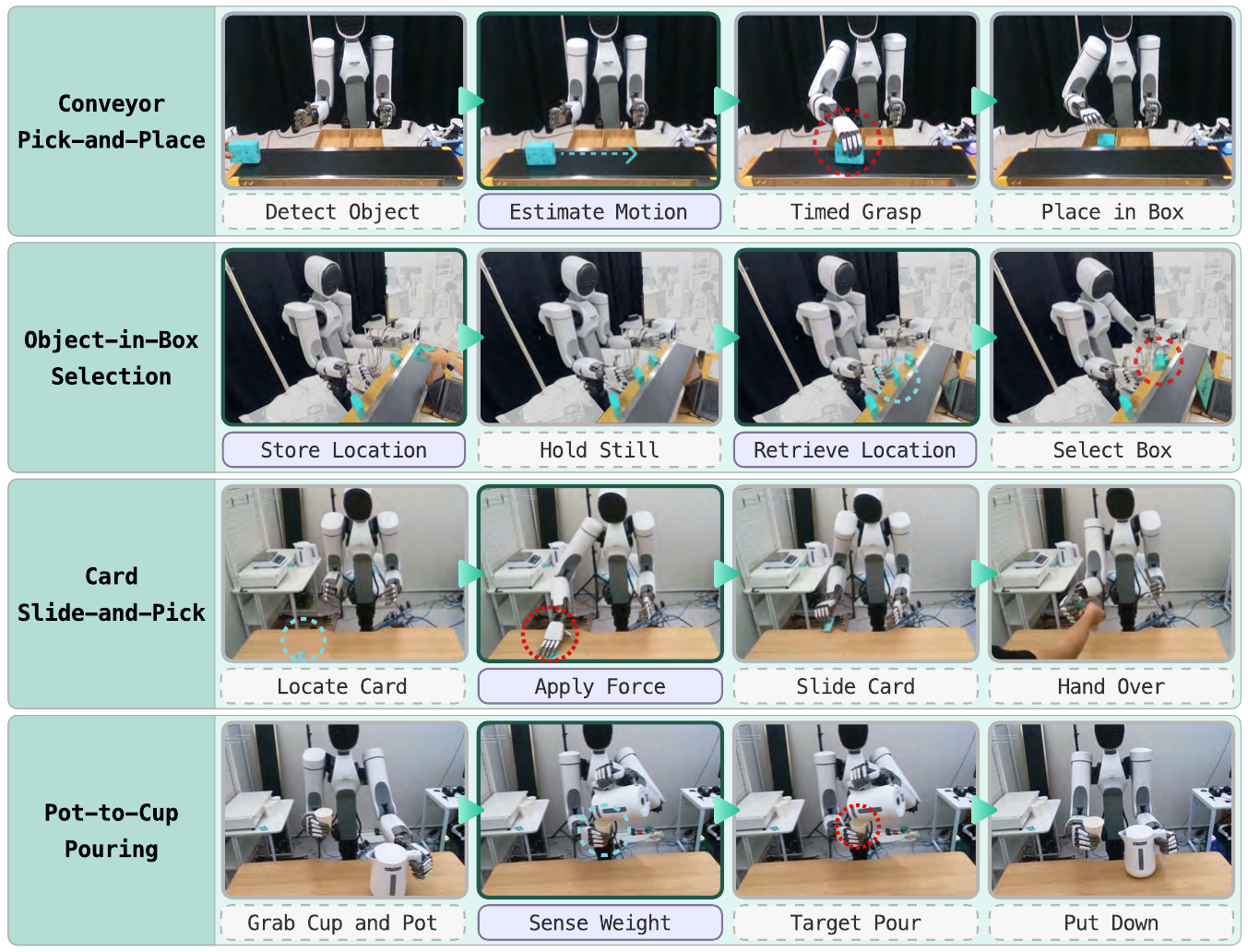}
    \caption{\textbf{ALLEX humanoid benchmark.}
    We design four tasks for evaluating functional capabilities in dexterous humanoid manipulation: \textit{Conveyor Pick-and-Place}, \textit{Object-in-Box Selection}, \textit{Card Slide-and-Pick}, and \textit{Pot-to-Cup Pouring}.
    }
    \label{figure:allex_task_overview}
\end{figure*}

\begin{figure*}[t]
    \centering
    \includegraphics[width=0.95\textwidth]{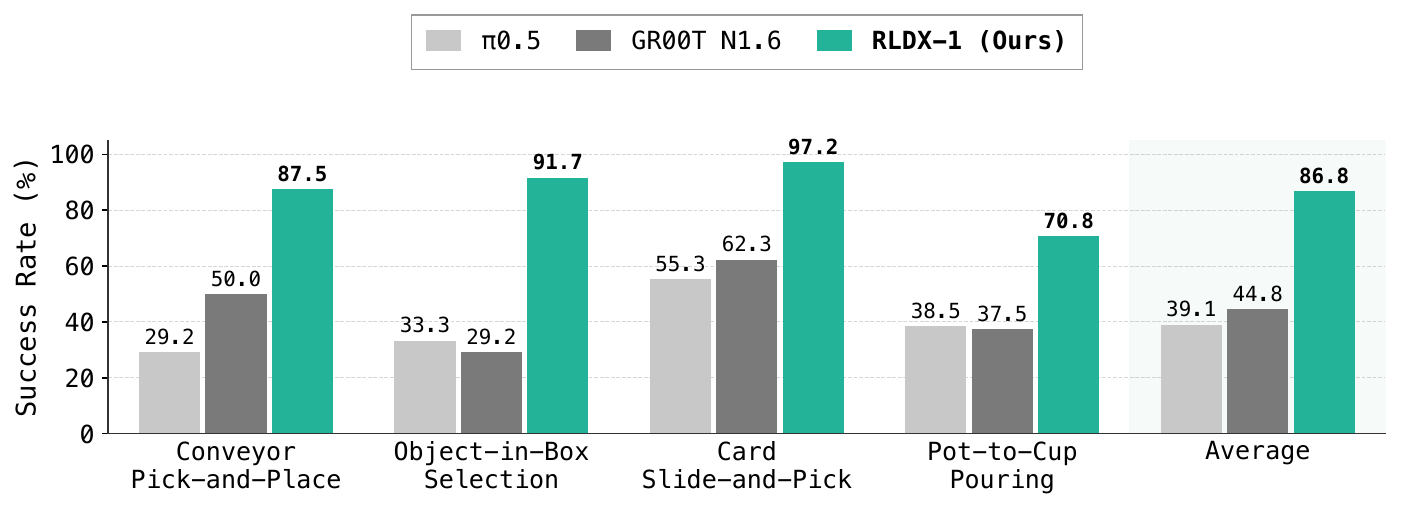}
    \caption{\textbf{ALLEX humanoid benchmark results.} We report the success rates (\%) of VLAs fine-tuned on the training dataset of each task. RLDX-1 substantially outperforms the baselines across all task categories, including motion awareness (\textit{Conveyor Pick-and-Place}), long-term memory (\textit{Object-in-Box Selection}), and physical sensing (\textit{Card Slide-and-Pick, Pot-to-Cup-Pouring}). 
    }
    \label{figure:allex_results}
\end{figure*}

To evaluate the functional capabilities of RLDX-1 in real-world dexterous manipulation, we curate a set of realistic, practical task scenarios for the ALLEX humanoid. 
As a whole upper-body humanoid equipped with high-DoF hands, ALLEX provides a suitable platform for evaluating human-like functional capabilities (see \Cref{figure:hardware_overview} (b) for the hardware details).
The benchmark includes tasks that require \textit{motion awareness}, \textit{long-term memory}, and \textit{physical sensing}, allowing us to assess whether RLDX-1 can handle dexterous manipulation scenarios beyond general visual-language understanding.
We describe the task details below (see \Cref{figure:allex_task_overview} for visualization) and provide more details in~\Cref{appendix:allex_details}.

\vspace{-0.5em}
\begin{itemize}[leftmargin=*,itemsep=0mm]
\item \textbf{Conveyor Pick-and-Place.}
This task evaluates robot behavior in dynamic environments through interaction with moving objects.
The robot is instructed to pick a box from a conveyor belt moving at varying speeds and place it on a shelf in front.
The varying conveyor speed introduces temporal dynamics, making it difficult to predict the next object position from static observations alone.
The policy needs to anticipate object motion and adapt its action accordingly.

\item \textbf{Object-in-Box Selection.}
This task requires leveraging long-term observation history rather than relying solely on the current observation.
A human picks up one of three boxes placed in front of the robot, puts an object inside, and returns it to its original position.
After a start signal, the robot is instructed to select the box containing the object.
The policy requires long-term reasoning over past observations to generate accurate actions.

\item \textbf{Card Slide-and-Pick.}
This task evaluates precise control of contact forces in contact-rich scenarios.
A card is placed on a table in front, and the robot slides it to the edge of the table, picks it up, and hands it over to a person in front.
The policy requires fine-grained control of contact forces to apply appropriate pressure to the card.
We vary the table height slightly, making the appropriate contact depth difficult to determine from visual observations alone and thus requiring physical signals to detect contact and modulate the applied force.

\item \textbf{Pot-to-Cup Pouring.}
This task evaluates understanding of the weights of manipulated objects through physical signals.
A pot containing balls and a cup are placed on a table in front, and the robot grasps both and tilts the pot to pour the balls into the cup.
The robot returns the pot and extends the cup forward when the cup reaches a target weight.
During pouring, visual cues change little, making it difficult for the policy to determine when pouring should stop based on vision alone.
Instead, it should estimate the cup's weight from the joint torques of the arm.

\end{itemize}

\paragraph{Evaluation Protocol}
We adopt task-specific evaluation criteria.
For \textit{Conveyor Pick-and-Place}, we evaluate performance at 4 different conveyor belt speeds, denoted as S1 through S4.
S1 and S4 are seen during training, while S2 and S3 are unseen speeds that lie between them, assessing the policy's capability to interpolate across motion conditions.
We measure the pick-and-place success rate for each speed.
For \textit{Object-in-Box Selection}, we measure the selection success rate per position across three box positions.
For \textit{Card Slide-and-Pick}, we evaluate performance using a three-stage progress score corresponding to sliding the card, picking it up, and handing it over, with scores of 0.33, 0.66, and 1.0, respectively.
For \textit{Pot-to-Cup Pouring}, we use a four-stage progress score corresponding to grasping both objects, pouring, returning the pot, and bringing the cup forward, with scores of 0.25, 0.5, 0.75, and 1.0, respectively.

\paragraph{Results}
As shown in \Cref{figure:allex_results}, RLDX-1 achieves the best performance across all tasks by a significant margin.
In \textit{Conveyor Pick-and-Place}, the baseline models collapse to nearly fixed-conveyor-belt-speed actions aligned with one of the seen speeds, regardless of the actual conveyor motion, indicating the lack of motion awareness.
GR00T N1.6 succeeds only at the lower speeds S1 and S2 (50.0\% on average), and $\pi_{0.5}$ succeeds at the faster seen speed S4 but largely fails at the unseen S3 (29.2\% on average).
In contrast, RLDX-1 maintains strong performance across both seen (100\%) and unseen (75\%) speeds, validating the effectiveness of the motion module.
At unseen speeds, RLDX-1 adaptively switches its action tempo to match the conveyor, enabling successful pick-and-place execution.
In \textit{Object-in-Box Selection}, baselines that rely only on the current observation fail to generate accurate choices.
GR00T N1.6 selects a box at random (29.2\%) regardless of the prior human demonstration, while $\pi_{0.5}$ repeatedly selects the same box throughout evaluation (33.3\%).
By contrast, RLDX-1 selects the correct target box with 91.7\% success, demonstrating the effectiveness of the memory module.
In \textit{Card Slide-and-Pick}, both GR00T N1.6 and $\pi_{0.5}$ exhibit diverse failure modes such as inaccurate sliding, failure to grasp the thin card, or dropping it during handover, indicating limited fine-grained control.
RLDX-1, on the other hand, achieves a near-perfect progress score of 97.2, showing that joint torque feedback is beneficial for contact-rich, fine-grained manipulation.
In \textit{Pot-to-Cup Pouring}, neither baseline completes the full task in any trial.
Both struggle to pour accurately into the cup, and even when pouring succeeds, they remain stuck in the pouring pose due to the lack of awareness of weight changes in the cup.
RLDX-1, in contrast, achieves a progress score of 70.8, outperforming the baselines by over 30 points.
Once pouring is complete, RLDX-1 perceives the cup's weight change and proceeds to complete the task smoothly, highlighting the effectiveness of the physics stream.
Overall, these results demonstrate the functional capabilities of RLDX-1 for dexterous humanoid manipulation, spanning motion awareness, long-term memory, and physical sensing, where existing VLAs fall short.

\subsection{Real-World Experiments: Franka Research 3}
\label{sec:exp-real-world-franka}

\begin{figure*}[t]
    \centering
    \includegraphics[width=\textwidth]{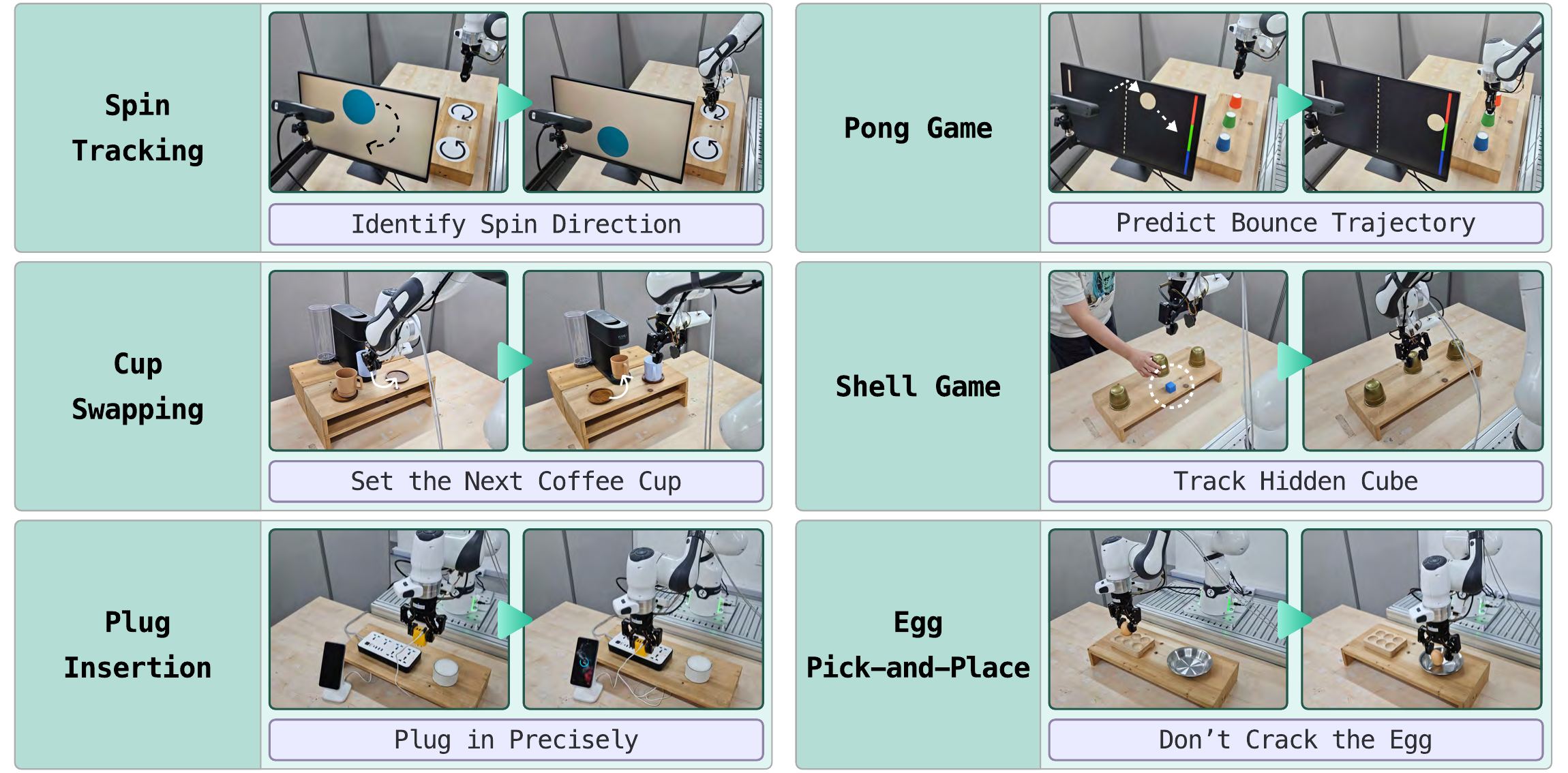}
    \caption{\textbf{Franka Research 3 benchmark.} We design six tasks for evaluating functional capabilities in dexterous single-arm robot manipulation: \textit{Spin Tracking}, \textit{Pong Game}, \textit{Cup Swapping}, \textit{Shell Game}, \textit{Plug Insertion}, and \textit{Egg Pick-and-Place}.
}
    \label{figure:franka_task_overview}
\end{figure*}
\

\begin{figure*}[t]
    \centering
    \includegraphics[width=\linewidth]{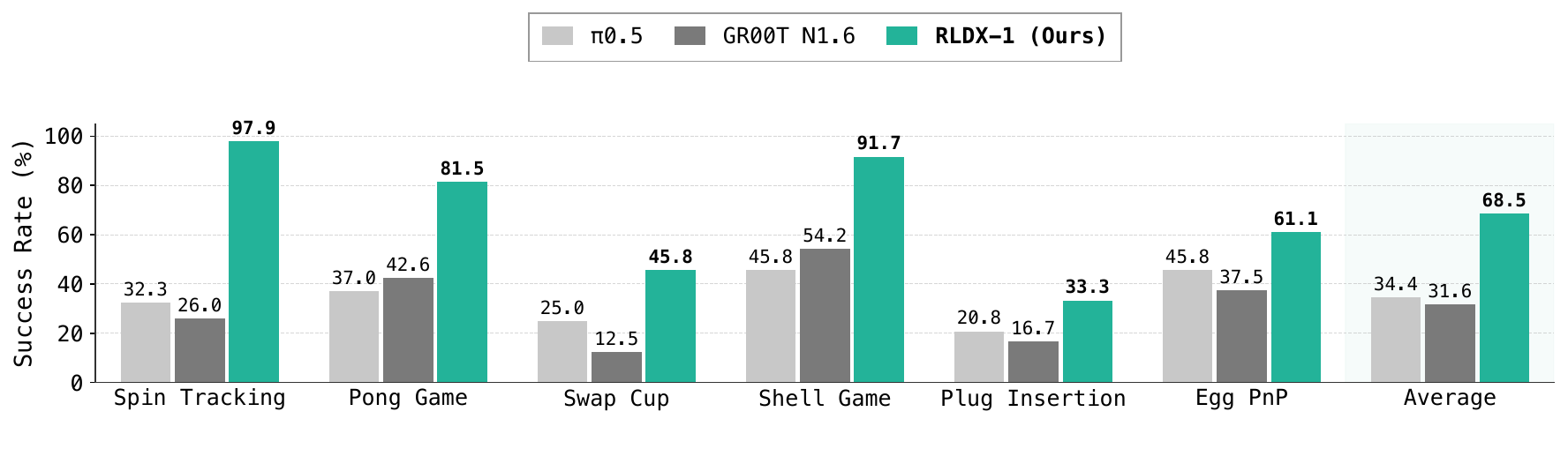}
    \vspace{-1.5em}
    \caption{\textbf{Franka Research 3 benchmark results.} We report the success rates (\%) of VLAs fine-tuned on the training dataset of each task. RLDX-1 substantially outperforms the baselines across all task categories, including motion awareness (\textit{Spin Tracking, Pong Game}), long-term memory (\textit{Cup Swapping, Shell Game}), and physical sensing (\textit{Plug Insertion, Egg PnP}).}
    \label{fig:franka_results}
\end{figure*}

To further evaluate the same functional capabilities on a different real-robot embodiment, we conduct experiments on the Franka Research 3 platform (FR3).
FR3 is a single-arm gripper robot equipped with an AnySkin tactile sensor \citep{bhirangi2025anyskin}, enabling evaluation of both visuomotor and contact-rich manipulation capabilities (see \Cref{figure:hardware_overview} (c) for the hardware details).
The benchmark includes video- and memory-demanding tasks, where short-horizon motion understanding or past-state memory is essential, as well as contact-rich manipulation tasks, where physical feedback is crucial for successful execution.
We describe the task details below (see \Cref{figure:franka_task_overview} for visualization) and provide more details in~\Cref{appendix:franka_details}.

\vspace{-0.5em}
\begin{itemize}[leftmargin=*,itemsep=0mm]
\item \textbf{Spin Tracking}. This task evaluates awareness of simple motion patterns.
Based on the camera view, the robot should point to ``Clockwise" when the object on the display moves clockwise, and to ``Counter-Clockwise" otherwise.
With varying initial positions, the policy cannot rely on positional cues and should estimate motion patterns.

\item \textbf{Pong Game}.
This task evaluates the prediction of more complex motion patterns.
The robot acts as a player in a Pong game on the display.
The right wall is divided into red, green, and blue regions, and the robot should point to the corresponding cup before the ball reaches the wall.
Due to frequent changes in the ball's trajectory, the policy needs to precisely capture motion dynamics.

\item \textbf{Cup Swapping}.
This task requires leveraging long-term observation history rather than relying solely on the current observation.
The robot should first move the cup currently on the coffee machine to the empty coaster, and then move the cup from the other coaster onto the coffee machine.
Therefore, it should remember which cup was originally on the coaster after moving the cup from the coffee machine to the empty coaster.

\item \textbf{Shell Game}.
This task also requires leveraging long-term observation history rather than relying
solely on the current observation.
The robot should identify the cup under which the cube is hidden among three cups.
The cube is initially shown to the policy, after which one cup is placed over the cube to hide it.
Since the cube is no longer visible after being covered, the policy must remember which cup contains the cube.

\item \textbf{Plug Insertion}.
This task evaluates precise control of contact forces in contact-rich scenarios, where the robot must insert a plug into a socket, which is completely occluded from the camera views.
Therefore, the robot requires torque and force cues to detect the contact and guide its alignment.

\item \textbf{Egg Pick-and-Place}.
This task also evaluates precise control of contact forces in contact-rich scenarios, where the robot must pick up a fragile egg and place it into the bowl without breaking it.
Therefore, force cues are essential for carefully gripping the object.

\end{itemize}

\paragraph{Evaluation Protocol} We adopt task-specific evaluation criteria and evaluate each policy over 24 trials per task unless otherwise specified.
For \textit{Spin Tracking}, we evaluate across 4 different initial positions and measure the success rate in responding to spin motion.
For \textit{Pong Game}, we measure the success rate based on correct pointing before the ball reaches the wall.
For \textit{Cup Swapping}, we evaluate performance at 2 different initial positions, where each position involves redundant trajectories that can be confused without memory.
For \textit{Shell Game}, we measure the selection success rate per position across 3 shell positions.
For \textit{Plug Insertion}, we evaluate the performance at 3 different target positions of plugging, and measure the insertion success rate.
For \textit{Egg Pick-and-Place}, we evaluate the performance at 3 different initial positions and measure the pick-and-place success rate (without cracking the egg).

\paragraph{Results}
In Figure~\ref{fig:franka_results}, we find that RLDX-1 achieves substantially higher performance than the baselines across all tasks.
In \textit{Spin Tracking}, both $\pi_{0.5}$ and GR00T N1.6 struggle to even detect the start of motion, obtaining only 32.3\% and 26.0\%, respectively.
However, RLDX-1 reaches 97.9\%, validating the motion perception capability of the motion module.
In \textit{Pong Game}, RLDX-1 predicts and responds to motions from diverse directions, achieving an 81.5\% success rate and demonstrating the practical potential of motion understanding.
In \textit{Cup Swapping}, both $\pi_{0.5}$ and GR00T N1.6 remain below 25\% success rate, whereas RLDX-1 achieves 45.8\%, showing a large performance improvement. 
This advantage becomes even more pronounced in \textit{Shell Game}. Notably, $\pi_{0.5}$ and GR00T N1.6 achieve around 50.0\%, whereas RLDX-1 achieves 91.7\%, demonstrating its history-aware decision-making ability.
A similar trend is observed in \textit{Plug Insertion}. 
$\pi_{0.5}$ and GR00T N1.6 achieve only 20.8\% and 16.7\%, respectively, showing that the baselines struggle with tasks involving subtle visual changes. 
Qualitatively, both baselines often fail to distinguish whether the plug is properly inserted into the target position, and continue applying force around the socket until failure. 
In contrast, RLDX-1 achieves a 33.3\% success rate and qualitatively applies downward force once the plug reaches the insertion position, unlike the variant without physical sensory inputs. 
These results suggest that tactile and force/torque information effectively improves performance in contact-rich manipulation tasks.
RLDX-1 also achieves strong performance in \textit{Egg Pick-and-Place}, reaching a 61.1\% success rate compared with 45.8\% for $\pi_{0.5}$ and 37.5\% for GR00T N1.6.
Qualitatively, models without tactile sensing often fail even when the gripper reaches a proper grasping position, because the grasping force is insufficient to securely hold the object. 
By contrast, when tactile sensing is used, the robot succeeds with high probability once the gripper reaches the correct grasping position. 
These observations show that RLDX-1 effectively improves contact-rich manipulation by incorporating physical sensing.

\subsection{Ablation and Analysis}
\label{sec:ablation-and-analysis}

\begin{figure}[t]
\centering
\includegraphics[width=0.6\linewidth]{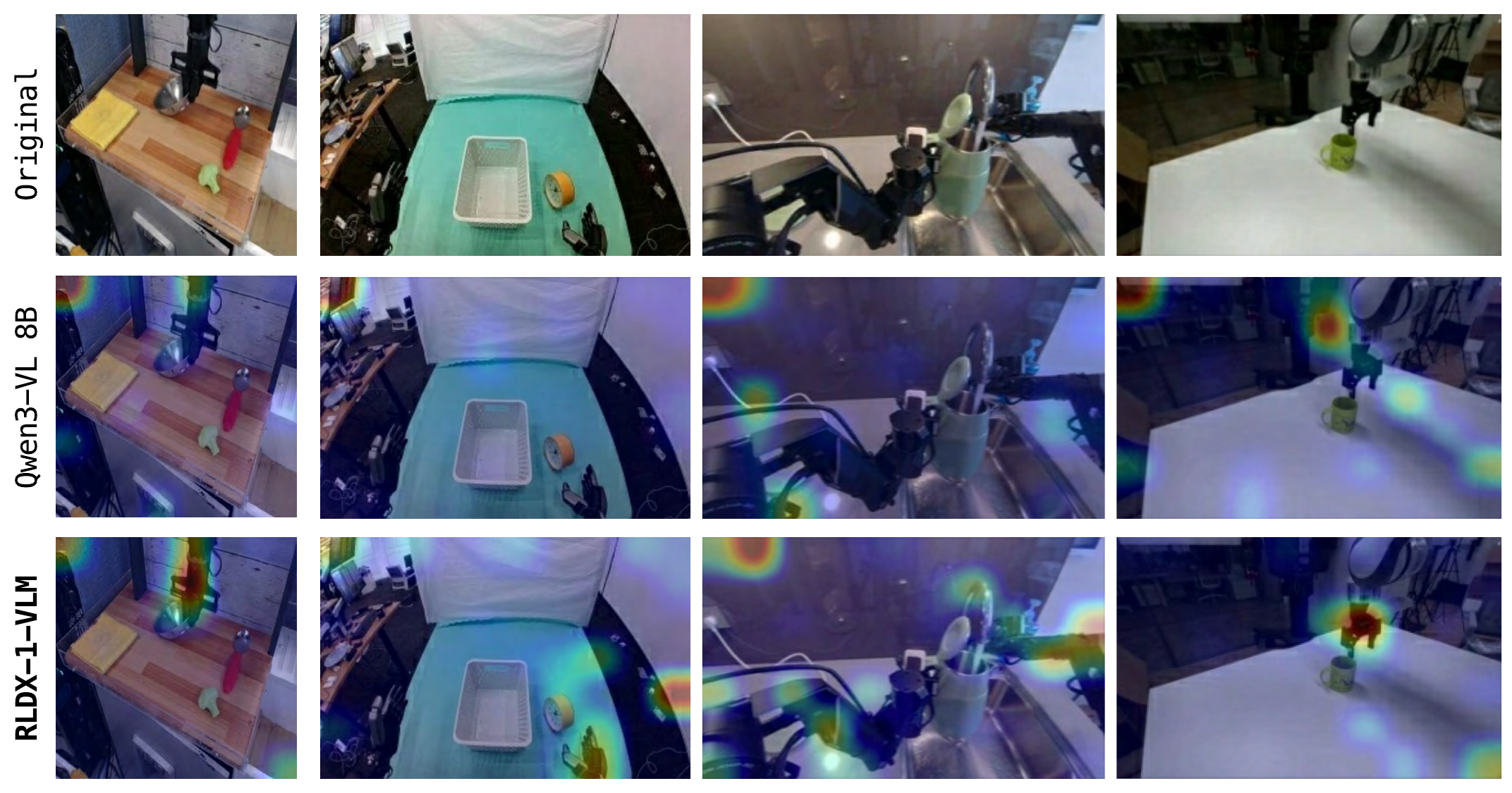}
\vspace{-0.5em}
\caption{\textbf{Attention-based analysis.} We visualize the self-attention scores from the last prompt token to the image patch tokens during task execution.}
\label{fig:vlm_atten}
\vspace{-1.0em}
\end{figure}

\paragraph{Vision-Language Model Analysis}
We first analyze how Vision-Language Model (VLM) feature extraction and robot-specific VQA adaptation affect downstream manipulation performance.
Specifically, we evaluate the effects of the VLM feature extraction layer and robot-specific VQA training through quantitative ablations, complemented by qualitative attention map visualizations. 
For the ablation study, we use the base architecture following the pre-training setup, where the action model takes VLM hidden states extracted from Layer 18 as input.
Each variant is trained from scratch without pre-training and evaluated on the \textit{RoboCasa Kitchen} benchmark.
As shown in \Cref{tab:layer_ablation}, we find that the VLM feature extraction layer has the largest impact on performance.
Using features from an earlier layer, such as Layer 8, reduces the success rate from 60.9\% to 51.1\%, while using features from a later layer, such as Layer 28, also degrades performance to 56.3\%.
We attribute this to the fact that intermediate VLM representations provide a more effective balance between visual grounding and semantic abstraction for action generation, whereas earlier layers lack sufficient semantics and later layers may become less aligned with fine-grained visual details required for manipulation.
Meanwhile, robot-specific VQA training also plays an important role in adapting VLM representations to downstream manipulation.
As shown in \Cref{tab:vlm_vqa_ablation}, we find that removing this training stage reduces the success rate from 60.9\% to 57.5\%, indicating that robot-specific VQA supervision helps refine VLM features into more action-relevant representations (see \Cref{fig:vlm_atten}).
To further examine this effect, we visualize the attention maps before and after VQA training.
Given robot observations and task instructions, the VLM before robot-specific VQA training exhibits limited attention to the robot embodiment and target objects.
After VQA training, however, the attention becomes more concentrated on the robot body and the objects to be manipulated.
These qualitative results support the quantitative findings, suggesting that robot-specific VQA training helps specialize the VLM for robotics tasks.

\begin{figure*}[t]
\centering
\captionsetup[table]{skip=1pt}
\captionsetup[sub]{skip=2pt}
\captionsetup[figure]{skip=1pt}

\begin{minipage}[t]{0.48\textwidth}
\vspace{0pt}
\centering
\captionof{table}{\textbf{Analysis of Vision-Language Model design choices} on the RoboCasa Kitchen benchmark \citep{nasiriany2024robocasa}.}
\label{tab:architecture_ablation}
\vspace{0.08em}
\subcaptionbox{Effect of layer selection.\label{tab:layer_ablation}}[\linewidth]{
\centering
\small
\begin{tabular}{lc}
\toprule
\textbf{Selected Layer} & \textbf{Success Rate (\%)} \\
\midrule
8  & 51.1 \\
\textbf{18 (Ours)}  & \textbf{60.9} \\
28 & 56.3 \\
\bottomrule
\end{tabular}
}
\\[1.0em]
\subcaptionbox{Effect of robot-specific VQA dataset.\label{tab:vlm_vqa_ablation}}[\linewidth]{
\centering
\small
\begin{tabular}{lc}
\toprule
\textbf{Method} & \textbf{Success Rate (\%)} \\
\midrule
Qwen3-VL 8B & 57.5 \\
\textbf{RLDX-1-VLM (Ours)} & \textbf{60.9} \\
\bottomrule
\end{tabular}
}
\end{minipage}
\hfill
\begin{minipage}[t]{0.50\textwidth}
\vspace{0pt}
\centering
\captionof{table}{\textbf{Analysis of synthetic data scale} on the GR-1 Tabletop benchmark \citep{bjorck2025gr00t}. We pre-train RLDX-1 with ActionNet \citep{fourier2025actionnet} as the real GR-1 humanoid dataset, augmented with our filtered synthetic data at varying proportions. We report the average success rate (\%) over 50 trials across 24 tasks, after fine-tuning with 1,000 demonstrations per task.}
\label{tab:syn_data_ablation_results}
\vspace{0.7em}
\small
\begin{tabular}{ccc}
\toprule
\multicolumn{2}{c}{\textbf{Pre-training Data}} & \multirow{2.5}{*}{\textbf{Success Rate (\%)}} \\
\cmidrule(lr){1-2}
Real & Synthetic & \\
\cmidrule(lr){1-2} \cmidrule(lr){3-3}
\textcolor{SJViolet}{\cmark} & \phantom{00}0\%  & 41.0 \\
\textcolor{SJViolet}{\cmark} & \phantom{0}25\%  & 45.6 \\
\textcolor{SJViolet}{\cmark} & \phantom{0}50\%  & 46.6 \\
\textcolor{SJViolet}{\cmark} & 100\% & \textbf{50.1} \\
\bottomrule
\end{tabular}
\end{minipage}
\end{figure*}

\paragraph{Effect of Synthetic Data}
To investigate the effectiveness of our synthetic data, we compare RLDX-1 models pre-trained with varying proportions of synthetic data.
Specifically, we pre-train RLDX-1 with a global batch size of 1024 and 60K gradient steps, restricted to the GR-1 humanoid embodiment for a controlled comparison. 
We fix the real GR-1 humanoid dataset \citep{fourier2025actionnet} and gradually scale the amount of synthetic GR-1 humanoid data to 25\%, 50\%, and 100\% of its full size, while keeping the real-to-synthetic sampling ratio fixed at 1:1. 
We then evaluate each model on the GR-1 Tabletop benchmark~\citep{bjorck2025gr00t} and report the results in \Cref{tab:syn_data_ablation_results}, where the success rate consistently improves as the synthetic proportion increases, rising from 41.0\% with real data alone to 50.1\% at the full synthetic scale. 
This scaling trend indicates that our task and scene augmentation strategies (see \Cref{subsec:synthetic-data}) effectively diversify the synthetic data and lead to consistent gains in downstream policy performance, thereby equipping RLDX-1 with broader coverage of manipulation scenarios that real data alone cannot sufficiently provide. 
Moreover, our synthetic data further improves spatial generalization on a specific task (see \Cref{appendix:additional_allex_exp}), motivating its use in mid-training for ALLEX humanoid (see \Cref{fig:midtrain-data}).
Overall, these results demonstrate that our synthetic data provides a scalable and effective complement to real robot data, improving downstream policy performance.

\begin{figure}[t]
    \centering
    \begin{minipage}[t]{0.32\linewidth}
        \centering
        \includegraphics[width=\linewidth]{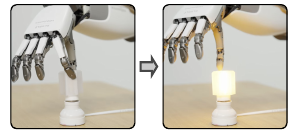}
        %\vspace{0.01in}
        \caption{\textbf{Overview of the \textit{Light Bulb Twisting} task.} In this task, the robot should twist the bulb using its hand until it turns on the light.}
        \label{fig:figure_lightbulbtwist_task}
    \end{minipage}
    \hfill
    \begin{minipage}[t]{0.65\linewidth}
        \centering
        \begin{subfigure}[t]{0.49\linewidth}
            \centering
            \includegraphics[width=\linewidth]{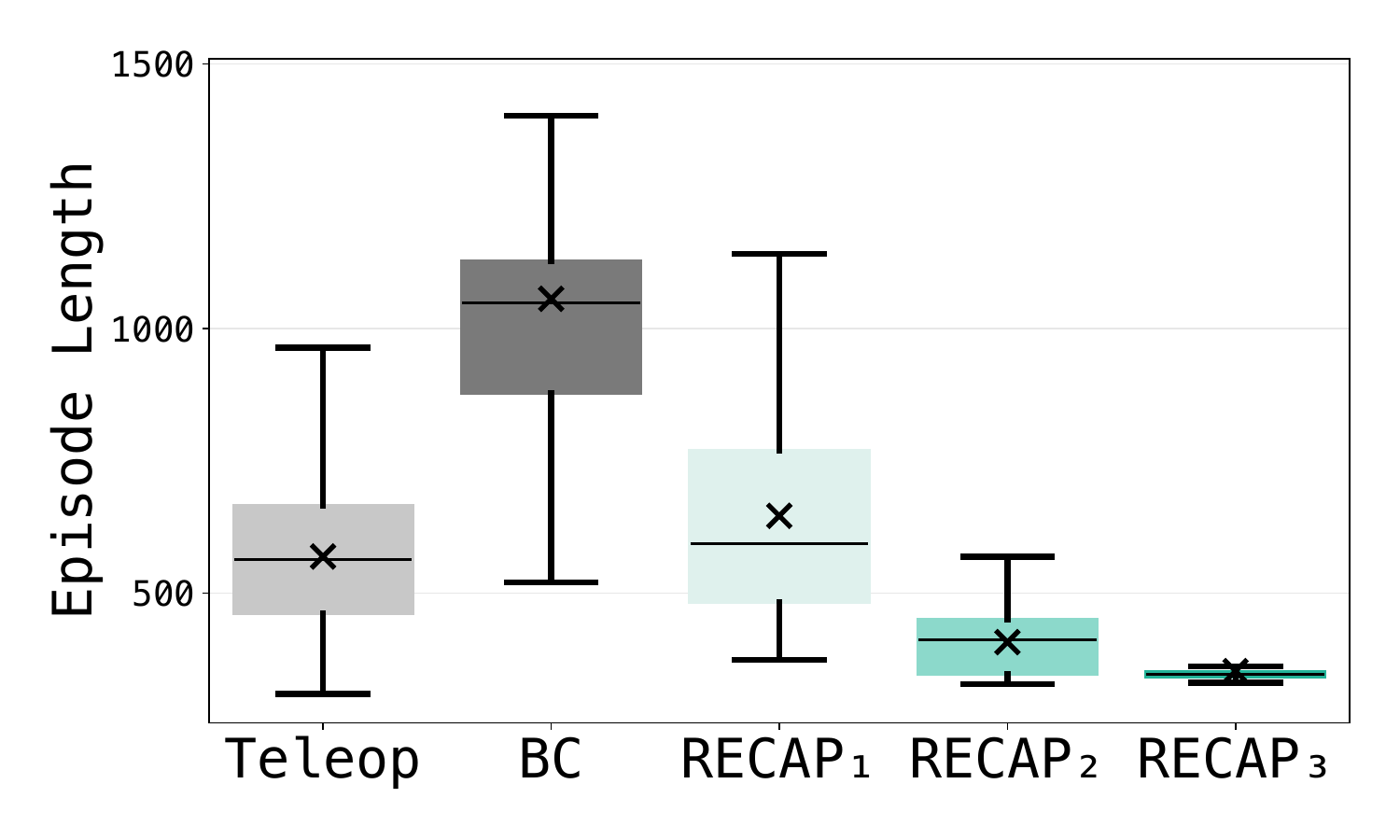}
            \vspace{-0.3in}
            \subcaption{Episode length by frames.}
            \label{fig:RECAP_result_frames}
        \end{subfigure}
        \hfill
        \begin{subfigure}[t]{0.49\linewidth}
            \centering
            \includegraphics[width=\linewidth]{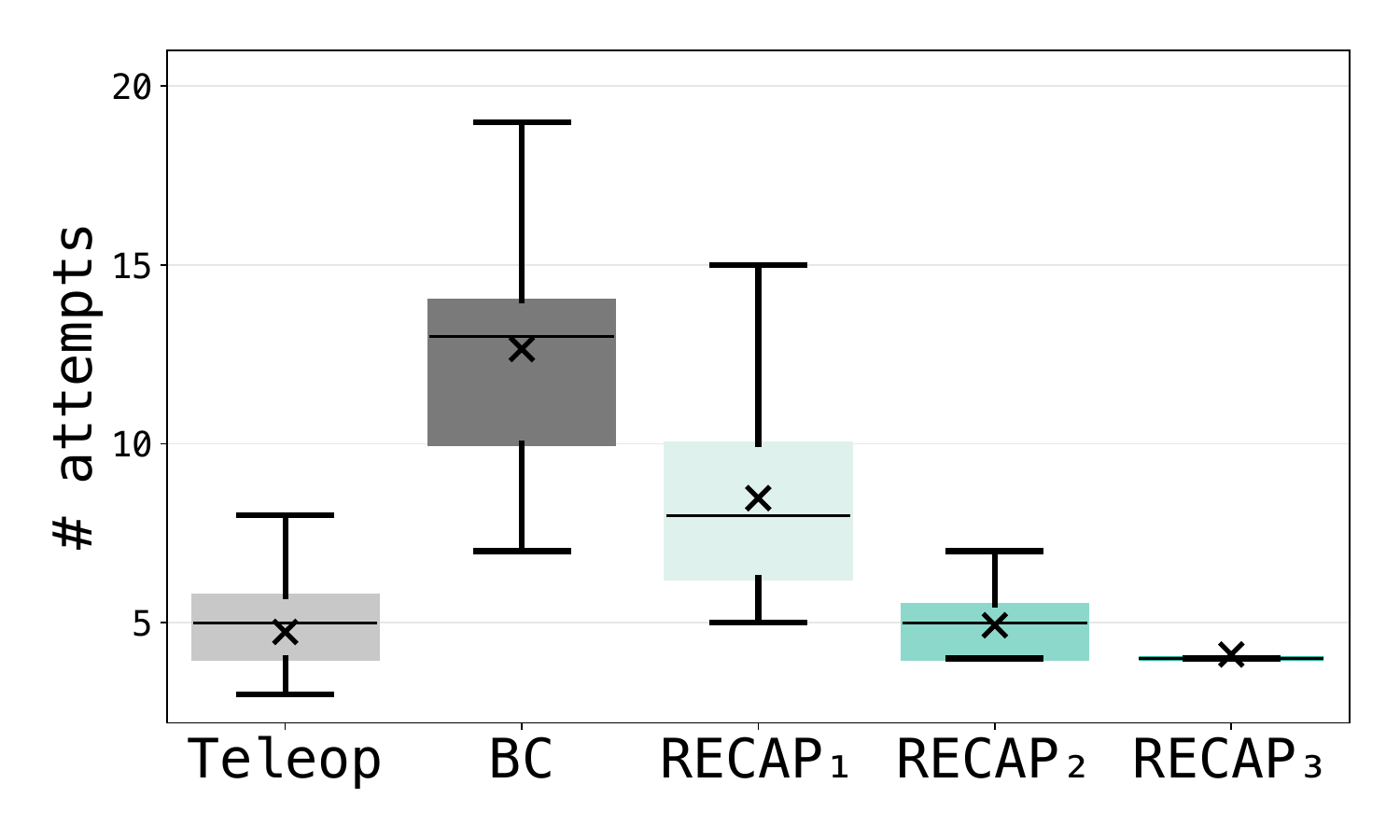}
            \vspace{-0.3in}
            \subcaption{Number of attempts.}
            \label{fig:RECAP_result_attempts}
        \end{subfigure}
        \vspace{-0.105in}
        \caption{\textbf{Results of RLDX-1 on the \textit{Light Bulb Twisting} task.} We report the box-and-whisker plot by (a) frames and (b) number of attempts, for expert demonstration, RLDX-1 after adopting imitation learning (BC), and reinforcement learning (RECAP$_1$ to RECAP$_3$).}
        \label{fig:RECAP_result}
    \end{minipage}
\end{figure}

\paragraph{Effect of RL Application}
To validate the effectiveness of RL training, we train RLDX-1 with RL and evaluate it on the \textit{Light Bulb Twisting} task, where the ALLEX humanoid must use the index finger and thumb of its right hand to twist a plastic light bulb held in a socket until the bulb lights up, as visualized in \Cref{fig:figure_lightbulbtwist_task}.
We report (a) the episode length in frames required to complete the task, and (b) the number of twist attempts to turn the bulb, where the lower values are preferred for both.
We compare expert demonstrations from teleoperators (Teleop), behavior cloning performance with a pre-trained RLDX-1 (BC), and three successive rounds of RECAP training using our task prediction critic (\ie RECAP$_1$ to RECAP$_3$).
As shown in \Cref{fig:RECAP_result}, we find that RECAP yields consistent gains over BC in both speed and robustness at every iteration. Most notably, RECAP$_3$ completes the task in $353 \pm 22$ frames with only $4.1 \pm 0.3$ attempts, a roughly $3\times$ reduction in both episode length and number of attempts compared to BC ($1056 \pm 326$ frames, $12.7 \pm 3.0$ attempts), and even surpasses the human teleoperation baseline.
The markedly smaller standard deviations of RECAP$_2$ and RECAP$_3$ further indicate improved trial-to-trial consistency.
These results show that RL-based refinement can substantially improve performance in challenging real-world manipulation tasks where imitation learning alone remains insufficient.

\begin{table}[t]
\centering
\caption{\textbf{Inference latency analysis.} We report the median latency (p50) over 
300 iterations after 100 warmup forward passes on a desktop with an  NVIDIA RTX~5090 and an 
Intel Core Ultra~7 265K.
Setup: ALLEX robot, dual-view $192{\times}256$ images, 
4 frames, action horizon 40, and 4 Euler denoising steps. 
Speedups are computed against PyTorch Eager.}
%\vspace{-0.5em}
\small
\begin{tabular}{lcc}
\toprule
\textbf{Inference Stack} & \textbf{w/o physics \& memory} & \textbf{All-modality} \\
\midrule
PyTorch Eager            & 67.0\,ms              & 71.2\,ms \\
\midrule
\multicolumn{3}{l}{\textit{Optimization-level}} \\
CUDA Graph + Torch.Compile                         & 56.9\,ms (1.18$\times$)          & 59.6\,ms (1.19$\times$) \\
\quad + Static Graph Conversion                    & 46.2\,ms (1.45$\times$)          & 48.9\,ms (1.46$\times$) \\
\quad +  Kernel Optimization      & \textbf{41.6\,ms (1.61$\times$)} & \textbf{43.7\,ms (1.63$\times$)} \\
\bottomrule
\end{tabular}
\label{tab:realtime-inference-performance}
\end{table}

\paragraph{Inference Optimization Comparison}
We analyze our inference optimization pipeline on a desktop equipped with an NVIDIA RTX~5090 and an Intel Core Ultra~7 265K.
All measurements are taken under the ALLEX configuration (dual-view 192$\times$256 images, 4 frames, action horizon 40, 4 Euler denoising steps), reporting the median over 300 iterations after 100 warmup forward passes.
PyTorch eager execution~\citep{paszke2019pytorch} and \texttt{Torch Compile}~\citep{ansel2024pytorch} serve as baselines, while our method is applied incrementally in two stages: static graph conversion, followed by operator fusion on top of the static graph.
We evaluate two RLDX-1 variants.
The first variant excludes the physical sensing and memory modules, while the second is the full all-modality model.
As shown in \Cref{tab:realtime-inference-performance}, \texttt{Torch Compile} yields suboptimal speedup over eager execution, with a ratio of 1.18$\times$ and 1.19$\times$ for the two variants.
This is because \texttt{Torch Compile} captures fragmented subgraphs rather than a single full graph, and applies only partial operator fusion.
As a result, the dominant launch and memory movement overheads remain.
Static graph conversion removes the dynamic execution overhead by capturing the forward pass as a single CUDA Graph, and delivers a 1.45$\times$ and 1.46$\times$ speedup over eager execution.
Operator fusion on top of this substrate further reduces kernel boundaries and HBM round-trips, bringing the cumulative speedup to 1.61$\times$ and 1.63$\times$.
The resulting per-step latency is 41.6\,ms for the variant without physics and memory, and 43.7\,ms for the all-modality model.
We also observe that the speedup ratios are nearly identical across the two variants.
This indicates that our optimization is robust to the additional memory module and physics stream introduced by the full model.
Overall, the ablation confirms that reducing runtime fragmentation, first at the graph level and then at the kernel level, is essential for meeting the per-step latency budget required for real-time policy serving with RLDX-1.
\section{Related Work}
\label{sec:related-work}

\paragraph{Foundation Models in Robotics}
Building human-like generalist robotic policies that can perceive, reason, and act across diverse tasks and environments has been a long-standing goal in robotics.
To this end, one line of work has explored modular approaches \citep{brohan2022rt,liang2023code,driess2023palm,huang2023voxposer,singh2022progprompt,team2025gemini}, where they utilize the Vision-Language Model (VLM; \citealt{touvron2023llama,beyer2024paligemma,bai2025qwen3,team2025gemma}) as a high-level task planner, and use additional specialized models (\eg visual tracking models) and low-level policies to generate actions.
However, recently, end-to-end Vision-Language-Action models (VLAs; \citealt{zitkovich2023rt,wu2023unleashing,team2024octo,kim2024openvla,black2024pi_0,bjorck2025gr00t,bu2025univla,intelligence2025pi_,nvidia2025gr00t,zheng2025x,nvidia2025gr00t16,community2026starvla}) have emerged as a promising paradigm, which unify visual perception, language-conditioned reasoning, and action generation within a single model, rather than exploiting only high-level semantic information from the VLM.
They either fine-tune VLMs to autoregressively generate discretized action tokens within the VLM vocabulary~\citep{kim2024openvla,pertsch2025fast}, or use a dedicated action decoder with a flow-matching transformer to generate actions conditioned on internal VLM representations such as hidden states or KV-cache features~\citep{black2024pi_0,bjorck2025gr00t}.
More recent works have explored extending VLAs with additional future video/state predictive objectives to improve physical generalization \citep{won2025dual,kim2026cosmos,ye2026world}.
While they have advanced toward generalist robotic policies by primarily focusing on versatility, we emphasize that versatility represents only one aspect of generalist intelligence.
For real-world manipulation, a generalist robotic policy must also acquire functional capabilities, including motion awareness, long-term memory, and physical sensing.
To this end, we propose a unified end-to-end VLA system that integrates these functional capabilities with versatile language-conditioned action generation, moving toward VLAs for practical real-world robot manipulation.

\paragraph{Synthetic Data for Robot Learning}
Although large-scale data is central to large model training, robotics lacks access to internet-scale data comparable to that in vision and language~\citep{brown2020language, radford2021learning}.
To address this limitation, previous work has explored the generation of synthetic data in simulation~\citep{li2023behavior,tao2024maniskill3, nasiriany2024robocasa}, where robot trajectories are obtained by rolling out policies learned through imitation learning~\citep{mandlekar2023mimicgen,jiang2025dexmimicgen} or reinforcement learning~\citep{laskin2020reinforcement} across diverse simulated environments.
However, visual and physical discrepancies between simulation and the real world introduce a sim-to-real gap~\citep{zhao2020sim}, limiting the effectiveness of simulator-generated data for real-world policy learning.
Recent advances in video generative models offer an alternative by synthesizing visually realistic robot videos~\citep{jang2025dreamgen,team2025gigaworld,kim2026robocurate} and then annotating them with actions using inverse dynamics models~\citep{baker2022video}.
While such video-based synthetic data offers a scalable way to augment robot datasets, prior work has largely been limited to tabletop pick-and-place tasks, leaving open whether this paradigm transfers to specialized humanoid hardware on realistic task scenarios.
In contrast, we propose a synthetic data framework for practical dexterous manipulation on the ALLEX humanoid, targeting real-world scenarios such as pot pouring and bulb twisting, and further find that such data remains beneficial even when combined with torque-annotated real data, despite containing no torque signals of its own.

\paragraph{Functionalities for Robot Manipulation}
Real-world robot manipulation requires diverse functional capabilities for dynamic, long-horizon, and contact-rich environments.
It requires motion awareness to reason about temporal changes, long-term memory to preserve task context over extended interactions, and physical sensing to capture contact and force information beyond vision.
Prior work has investigated these capabilities from different perspectives.
For motion awareness, many works leverage multi-frame inputs by using the full temporal context to generate actions \citep{wu2023unleashing,team2024octo,cheang2024gr,huang2025otter,wang2025unified,torne2025learning,li2025cronusvla}.
Moreover, several works leverage them efficiently by selecting key-frames \citep{wen2021keyframe}, summarizing entire sequences into high-level visualizations \citep{sundaresan2024rt,zheng2024tracevla}, or compressing past observations \citep{wen2020fighting,seo2023regularized,jang2025contextvla}.
For long-term memory, some approaches process past observations using a VLM and then use the resulting feature vectors to generate actions~\citep{sridhar2025memer,intelligence2026pi}. More closely related to our approach, another line of work introduces a memory module that stores past observations in a memory queue~\citep{fang2025sam2act,shi2025memoryvla,koo2025hamlet}.
For physical sensing, most prior work focuses on leveraging a single physical signal to improve performance on contact-rich tasks~\citep{yu2025forcevla,zhang2025ta,huang2025tactile,bi2025vla,cheng2025omnivtla,zhang2025vtla}. More recent approaches extend this direction by integrating multiple physical signals~\citep{lee2026modular}.
However, these approaches are designed to address individual functionalities, whereas real-world manipulation often requires robots to use multiple functionalities jointly. 
In contrast, our work integrates these three capabilities into a unified end-to-end policy, enabling a single model to combine complementary functional capabilities for real-world manipulation.
\section{Conclusion}
\label{sec:conclusion}

In this paper, we have presented \textbf{RLDX-1}, a general-purpose Vision-Language-Action model (VLA) for human-like dexterous manipulation in real-world environments.
RLDX-1 goes beyond versatile intelligence by integrating key functional capabilities into a unified architecture for real-world manipulation, including motion awareness, long-term memory, and physical sensing.
To make these capabilities practical for real-world deployment, RLDX-1 further combines synthetic data augmentation for dexterous robot data, three-stage training, and inference optimization for responsive control.
Our experiments demonstrate that RLDX-1 substantially improves both functional-capability-specific and general manipulation performance.
In particular, on ALLEX humanoid tasks that demand diverse functional capabilities, RLDX-1 achieves a success rate of approximately 90\%, while frontier VLAs remain around 40\%.
RLDX-1 also consistently outperforms these baselines on every simulation benchmark and real humanoid benchmark requiring versatile intelligence, showing that the proposed design generalizes beyond capability-specific tasks.
We hope that these results position RLDX-1 as a meaningful step toward general-purpose robot policies capable of human-like dexterous manipulation in complex real-world environments.
%\newpage

\makeatletter
\renewcommand\@biblabel[1]{[#1]}
\let\@bibsetup\NAT@bibsetnum
\makeatother

\newpage
\bibliographystyle{plainnat}
\bibliography{reference}

\newpage

\appendix
\crefalias{section}{appendix}
\crefname{appendix}{Appendix}{Appendices}
\Crefname{appendix}{Appendix}{Appendices}

\addcontentsline{toc}{section}{Appendix}
\addtocontents{toc}{\protect\setcounter{tocdepth}{-1}}

\section{Contributors and Acknowledgments}
\label{sec:contributors}

Within each role, names are listed alphabetically by first name and then by last name.
All Project Leads contributed equally.

\subsection{Main Contributors}

\textbf{Project Leads}
\vspace{-1.0em}
\begin{itemize}[leftmargin=*,itemsep=0mm]
    \item Dongyoung Kim: Sub-Led Data \& Model Training.
    \item Huiwon Jang: Sub-Led Data \& Architecture.
    \item Myungkyu Koo: Led Model Architecture.
    \item Suhyeok Jang: Led Synthetic Data.
    \item Taeyoung Kim: Led Model Training.
\end{itemize}

\textbf{Research Leads}
\vspace{-1.0em}
\begin{itemize}[leftmargin=*,itemsep=0mm]
    \item Dongyoung Kim, Huiwon Jang, Jinwoo Shin: Led the overall research direction and orchestrated cross-area coordination among contributors.
\end{itemize}

\subsection{Core Contributors}
Grouped by area of contribution. Names within each group are listed alphabetically by first name.
Multi-role members appear in multiple groups, with role-specific descriptions per group.

\textbf{Model Architecture}
\vspace{-1.0em}
\begin{itemize}[leftmargin=*,itemsep=0mm]
    \item Daewon Choi: Contributed to the early-stage design of MSAT.
    \item Heeseung Kwon: Designed and implemented the motion module.
    \item Jimin Lee: Designed and implemented physical sensory signal integration.
    \item Kyungmin Lee: Contributed to the early-stage design of MSAT.
    \item Seungcheol Park: Supported the validation of physical sensory signal integration.
\end{itemize}

\textbf{Synthetic Data}
\vspace{-1.0em}
\begin{itemize}[leftmargin=*,itemsep=0mm]
    \item Byungjun Yoon: Designed task augmentation and video quality filtering in the synthetic data pipeline.
    \item Changsung Jang: Supported synthetic data generation by training an inverse dynamics model, generating synthetic data, and implementing motion-consistency-based filtering on ALLEX.
    \item John Won: Supported scene augmentation and GR-1 synthetic data generation.
    \item Junyoung Sung\footnote{KAIST-affiliated intern}: Supported ALLEX synthetic data generation.
    \item Minseong Han: Supported ALLEX synthetic data generation and evaluation, and processed in-house datasets for video model fine-tuning on ALLEX.
    \item Sejune Joo: Designed real-world experiments for validation of ALLEX synthetic data and contributed to the early-stage design of ALLEX synthetic data generation.
    \item Seungku Kim: Designed motion-consistency-based filtering in the synthetic data pipeline.
\end{itemize}

\textbf{Model Training Pipeline}
\vspace{-1.0em}
\begin{itemize}[leftmargin=*,itemsep=0mm]
    \item Byungjun Yoon: Supported VLM/VLA training by optimizing the multi-node training pipeline.
    \item Jaehyun Kang: Drove codebase refactoring and maintenance.
    \item Sejune Joo: Supported the ALLEX demo by training baseline models; implemented training and asynchronous inference pipelines to enable deployment of baseline models on ALLEX.
    \item Seungjun Moon: Integrated baseline models into the real ALLEX evaluation pipeline and automated ALLEX demo evaluation.
\end{itemize}

\newpage

\textbf{Real-World Robot Evaluation \& Demonstration}
\vspace{-1.0em}
\begin{itemize}[leftmargin=*,itemsep=0mm]
    \item Beomjun Kim: Supported the OpenArm evaluation.
    \item Heeseung Kwon: Designed real-world motion-specific experiments.
    \item Jaehyun Kang: Designed memory-related experiments on ALLEX.
    \item Jaekyoung Bae: Led the production-grade research direction and real-world demo task design and evaluation for ALLEX.
    \item Jimin Lee: Designed real-world experiments for physical sensing evaluation on FR3.
    \item Joonwoo Ahn: Designed teleoperation-based data collection scenarios for ALLEX demo tasks, defined task-specific data distribution strategies considering consistency and variance, and improved task performance through targeted data refinement.
    \item Junhyeong Park\footnote{KAIST-affiliated intern}: Supported the OpenArm evaluation.
    \item Seungcheol Park: Designed real-world experiments for validating physical sensory signals on ALLEX.
\end{itemize}

\textbf{Reinforcement Learning}
\vspace{-1.0em}
\begin{itemize}[leftmargin=*,itemsep=0mm]
    \item Donguk Lee: Developed RECAP training and inference functionality; built the real-world RL evaluation pipeline; conducted evaluations on the real ALLEX robot.
    \item Seonil Son: Designed and developed the value model for RECAP; produced and analyzed ALLEX experimental results.
    \item Yonghoon Dong: Supported RECAP training.
    \item Yongjin Cho: Designed the RL task and led the team's effort on it; developed RECAP training and inference functionality.
\end{itemize}

\textbf{Inference Optimization \& Deployment}
\vspace{-1.0em}
\begin{itemize}[leftmargin=*,itemsep=0mm]
    \item Dongsu Han: Led the research direction for the inference optimization pipeline.
    \item Hojin Jeon: Supported inference optimization integration and evaluation.
    \item Jaehyun Kang: Built the asynchronous inference pipeline.
    \item Jihyuk Lee: Built and designed the inference optimization pipeline for the RLDX-1 architecture, including static-graph conversion, CUDA Graph integration, and Triton kernel optimization.
    \item Minsung Yoon: Supported kernel optimization and real-time chunking evaluation.
    \item Seunggeun Cho: Supported kernel evaluation for the inference optimization pipeline.
    \item Youngchan Kim: Designed and implemented the Fused Attention kernel for the RLDX-1 inference pipeline, designed GraphSafe static-graph conversion by analyzing graph-break sources, and implemented and analyzed static-graph conversion and Triton kernel optimizations.
\end{itemize}

\subsection{Contributors}

Grouped by area of contribution. Names within each group are listed alphabetically by first name.
Multi-role members appear in multiple groups, with role-specific descriptions per group. All contributors are affiliated with RLWRLD unless otherwise noted.

\textbf{Training Infrastructure}
\vspace{-1.0em}
\begin{itemize}[leftmargin=*,itemsep=0mm]
    \item Jaeheon Jung: Built end-to-end infra for training; built a teleoperation-based data collection system using leader devices across real \& simulation environments; built an inference and evaluation system for multiple VLA models, supporting both real \& simulation environments.
    \item Joochul Chang: Managed GPU infrastructure for training and experiment; built dataset management system and pipeline automation.
\end{itemize}

\textbf{Robot Control System}
\vspace{-1.0em}
\begin{itemize}[leftmargin=*,itemsep=0mm]
    \item Jaewoo Kim: Led the overall robot system, including tele-operation, human data, and system-level code refactoring.
    \item Junhyeok Park: Built the low-level robot control interfaces; designed and implemented the teleoperation control stack; built the data-collection pipeline for fine-tuning, especially for RB-Y1.
    \item Seunghyun Kim: Built the low-level robot control interfaces; designed and implemented the teleoperation control stack; built the data-collection pipeline for fine-tuning, especially for OpenArm.
\end{itemize}

\textbf{Robot Hardware Research \& Prototyping}
\vspace{-1.0em}
\begin{itemize}[leftmargin=*,itemsep=0mm]
    \item Chang Hwan Kim: Sourced/modified upper body robot platform for data collection; maintained hardware systems including ALLEX; analyzed/integrated various upper body, arms, dexterous hands and sensors, especially for OpenArm.
    \item Sungryol Yang: Built OpenArm data collection hardware platform; maintained ALLEX; fabricated benchmark task objects.
\end{itemize}

\textbf{Teleoperation System}
\vspace{-1.0em}
\begin{itemize}[leftmargin=*,itemsep=0mm]
    \item Jaeheon Jung: Built teleoperation systems supporting robot data collection.
    \item Kwanghoon Kim: Built the camera processing, video streaming, and VR integration components of the teleoperation system; troubleshot performance issues; built the operating environment for ALLEX; supported researchers in operating ALLEX.
    \item Seungyup Ka: Developed the teleoperation, data collection and inference backbone codebases for ALLEX; built the operating environment for ALLEX; supported researchers in operating ALLEX.
\end{itemize}

\textbf{Teleoperation Data Collection}
\vspace{-1.0em}
\begin{itemize}[leftmargin=*,itemsep=0mm]
    \item Hyunsoo Choi, Hyunsoo Shin, Jinwook Kim, Sangjun Kim, Seungjun Lee, Yeonwoo Bae: Collected teleoperation data for ALLEX.
    \item Dohyeon Kim, Jungwoo Park, Seunghoon Shim, Wook Jung: Collected teleoperation data for OpenArm and Franka.\footnote{Dohyeon Kim, Jungwoo Park, and Seunghoon Shim are KAIST-affiliated interns; Wook Jung is affiliated with KAIST}
    \item  Yashu Shukla: Established an automated training-inference-evaluation pipeline for ALLEX simulations.
\end{itemize}

\textbf{Human Data Pipeline}
\vspace{-1.0em}
\begin{itemize}[leftmargin=*,itemsep=0mm]
    \item Joochul Chang: Built the human data collection pipeline.
    \item Kyoungwhan Choe: Conducted early research on human data processing and retargeting; established the foundation for integrating human demonstrations.
    \item Sangwoo Kim: Built the human data collection pipeline, establishing the foundation for incorporating human demonstration data into the system.
    \item Sejune Joo: Designed and deployed human data training pipeline; contributed to the early-stage development of the human data collection pipeline.
\end{itemize}

\textbf{Live Demo Production}
\vspace{-1.0em}
\begin{itemize}[leftmargin=*,itemsep=0mm]
    \item Heecheol Kim, Heewon Lee, Junghun Park, Manoj Bhadu, Nayoung Oh, Yeonjae Lee: Collected live demo-task data and fine-tuned RLDX-1 for live demonstrations.
    \item Joonsoo Kim: Led live demo-task data collection and fine-tuning of RLDX-1 for live demonstrations.
\end{itemize}

\textbf{Dexterity Benchmark for Industry Conversion}
\vspace{-1.0em}
\begin{itemize}[leftmargin=*,itemsep=0mm]
    \item Hensen Ahn: Designed and curated DexBench, the industry-oriented dexterity benchmark for evaluating RLDX-1 in industry conversion scenarios.
    \item Junho Cho: Contributed to DexBench narrative and figure direction; aligned cross-team outputs with the dexterity benchmark team.
    \item Kangwook Lee: Defined the dexterity standards for industry conversion.
\end{itemize}

\textbf{Project Coordination \& Public Release}
\vspace{-1.0em}
\begin{itemize}[leftmargin=*,itemsep=0mm]
    \item Hyungkyu Ryu: Led overall project coordination and management.
    \item Junho Cho: Authored technical narrative; wrote and polished tech-blog content; produced and scripted demo videos; directed figures and infographics; aligned cross-team outputs.
    \item Junwon Lee: Coordinated end-to-end release planning; managed licensing, including data, code, and model weights, for public distribution.
\end{itemize}

\subsection{Acknowledgments}

We thank all members of RLWRLD whose support made the development and release of RLDX-1 possible. While this report credits direct research and product contributions, this work would not have been possible without the broader company's shared belief in our mission.

\textbf{Executive Sponsorship}
\vspace{-1.0em}
\begin{itemize}[leftmargin=*,itemsep=0mm]
    \item Junghee Ryu (CEO)
\end{itemize}

\paragraph{Operational Support}
We thank the Operational Support team for the day-to-day support that enabled the project to run smoothly:
\vspace{-1.0em}
\begin{itemize}[leftmargin=*,itemsep=0mm]
    \item Changhyun Hong, Jeeye Lee, Jeongwan Choi, Jihye Song, Jiyeon Olivia Chun, Junsang Yoo, Sara Jang, Sunjeong Kim, Sunyoung Jeon, Youngwoong Cho
\end{itemize}

\paragraph{Business \& Partnerships}
We thank the Business \& Partnerships team---Strategy, Sales Partnership, Robotics Deployment, Branch operations, and Business Research---for partner relations, market alignment, and field-side coordination that surrounded RLDX-1's release:
\vspace{-1.0em}
\begin{itemize}[leftmargin=*,itemsep=0mm]
    \item Amine Benari, Beopryong Kim, Carl Choi, Donghyun Kim, Eunkyu Ko, Hahyun Park, Hajin Kim, Hayeon Kim, Hayoung Jin, Hina Koizumi, Hoon Lee, Hyunji Song, Ilgyu Shin, Inseok Lee, Iris Cho, Jaewon Lee, Jehoon Kim, Jihoon Lee, Maako Hori, Mark Lee, Namho Kim, Seeun Sung, Seonggyeong Kim, Seungbeen Jeon, Seungwoo Choi, Shogo Koda, Suhwan Choi, Suyeon So, Taehyun Jung, Yiroom Yum, Yongjae Byeon, Younghoon Shin, Younseo Kim
\end{itemize}

\paragraph{Academic Advisors}
We thank our academic advisors and the researchers of their labs for the invaluable guidance and discussions that shaped RLDX-1:
\vspace{-1.0em}
\begin{itemize}[leftmargin=*,itemsep=0mm]
    \item Hanbyul Joo (Seoul National University), Jongwoo Lim (Seoul National University), Minsu Cho (POSTECH), Sungjoon Choi (Korea University)
\end{itemize}

\paragraph{Hardware Partners}
We thank WIRobotics---including Yong-Jae Kim---for the ALLEX robot platform and ongoing hardware support throughout the development of RLDX-1.

\paragraph{Final Note}
We also thank every member of RLWRLD and the researchers of ALIN-Lab (Algorithmic Intelligence Lab, KAIST) under Prof. Jinwoo Shin — whether named here or not — whose work, discussions, and support made this release possible.

\newpage

\section{Datasets Details}
Here, we describe the details of the datasets used throughout pre-training and mid-training, covering image preprocessing (\Cref{appendix:data:image-preprocessing}), public real-world data (\Cref{sec:appendix:data:real}), and synthetic data (\Cref{sec:synth-data-details}).

\subsection{Image Preprocessing}
\label{appendix:data:image-preprocessing}
To reduce the vision-token budget during training, we resize all input frames offline following Qwen3-VL's \texttt{smart\_resize} procedure~\citep{bai2025qwen3}. Given a source image of resolution $H \times W$, we compute target dimensions $(H', W')$ such that $H' \cdot W' \leq 256^2$ while preserving the original aspect ratio as closely as possible, with both $H'$ and $W'$ constrained to be integer multiples of $\text{factor} = \text{patch\_size} \times \text{spatial\_merge\_size} = 16 \times 2 = 32$. Concretely, if the aspect-preserved rounded resolution already satisfies the pixel budget, we round each dimension to the nearest multiple of $32$. This yields at most $(256 / 32)^2 = 64$ vision tokens per frame. 

\subsection{Public Real-World Real Data Details}
\label{sec:appendix:data:real}

In \Cref{tab:oxe-mixture}, we describe the detailed dataset ratios of the Open-X-Embodiment (OXE; \citealt{o2024open}) mixture used for pre-training RLDX-1. We follow the OXE dataset mixture commonly referred to as OXE Magic Soup, as used in Octo \citep{team2024octo} and OpenVLA \citep{kim2024openvla}.

\begin{table}[ht]
\centering
\caption{Detailed composition of our Open-X-Embodiment (OXE) subset.}
\label{tab:oxe-mixture}
\small
\setlength{\tabcolsep}{6pt}
\begin{tabular}{lr@{\hskip 3em}lr}
\toprule
\textbf{Dataset} & \textbf{Ratio (\%)} & \textbf{Dataset} & \textbf{Ratio (\%)} \\
\midrule
Fractal                     & 16.00 & DobbE                       & 2.00 \\
Kuka                        & 16.00 & TOTO                        & 2.00 \\
BridgeV2                    & 16.00 & VIOLA                       & 1.20 \\
BC-Z                        & \phantom{0}8.00 & Berkeley Autolab UR5        & 1.20 \\
FMB                         & \phantom{0}8.00 & IAMLab CMU Pickup-Insert    & 1.00 \\
Language Table              & \phantom{0}5.00 & NYU Franka Play             & 0.80 \\
Stanford HYDRA              & \phantom{0}5.00 & Berkeley Fanuc Manipulation & 0.80 \\
TACO Play                   & \phantom{0}3.36 & Jaco Play                   & 0.60 \\
FurnitureBench              & \phantom{0}2.80 & Berkeley Cable Routing      & 0.32 \\
RoboTurk                    & \phantom{0}2.60 & Austin BUDS                 & 0.24 \\
Austin Sailor               & \phantom{0}2.40 & CMU Stretch                 & 0.16 \\
UT Austin MUTEX             & \phantom{0}2.40 & DLR EDAN Shared Control     & 0.08 \\
Austin Sirius               & \phantom{0}2.00 & UCSD Kitchen                & 0.04 \\
\bottomrule
\end{tabular}
\end{table}

\newpage

\subsection{Synthetic Data Details}
\label{sec:synth-data-details}

To generate diverse synthetic data for specialized real-world scenarios, we use pre-trained video generative models~\citep{nvidia2025cosmospredict2, ali2025world} to generate novel trajectories from initial images that are unseen in existing datasets. Since off-the-shelf video diffusion models are insufficient to capture egocentric robot viewpoints and manipulation dynamics, we fine-tune them on robot demonstrations, training dedicated models for the GR-1 and ALLEX humanoids.

\paragraph{GR-1 Video Fine-tuning}
We build the GR-1 video model on top of Cosmos-Predict2-14B~\citep{nvidia2025cosmospredict2} and fine-tune it on two publicly available sources: 3,027 videos from ActionNet~\citep{fourier2025actionnet} and 92 videos from the public NVIDIA GR-1 dataset~\citep{jang2025dreamgen}. We re-caption all videos with Qwen3-VL 8B Instruct~\citep{bai2025qwen3} into short, medium, and long variants, from which one is randomly sampled during training. We apply LoRA~\citep{hu2022lora} fine-tuning with rank 32, training on 93-frame clips at 432$\times$768 resolution for 10K steps with a batch size of 4.

\paragraph{ALLEX Video Fine-tuning}
We build the ALLEX video model on top of Cosmos-Predict2.5-2B~\citep{ali2025world}. Since ALLEX data alone may be limited in task coverage, we mix in demonstrations from other embodiments to expose the model to a broader set of manipulation motions: we combine our in-house ALLEX and OpenArm data with the 3,027 ActionNet episodes used above, at a 2:1:1 ratio. To provide more detailed task descriptions, we re-caption all videos with Qwen3-VL 8B Instruct or Qwen3-VL 30B-A3B Instruct~\citep{bai2025qwen3} into short, medium, and long variants, and randomly sample one during training. During training, we sample 93-frame clips with two complementary strategies: \emph{uniform sampling} selects 93 frames at evenly spaced intervals to cover the full episode, while \emph{random sampling} takes 93 consecutive frames from a random start time at 16 FPS for fine-grained temporal continuity. We fully fine-tune the model at 432$\times$768 resolution for 20K steps with a batch size of 8.

\paragraph{Action Annotation}
Since the generated videos lack ground-truth action labels, we annotate them using an Inverse Dynamics Model (IDM) that predicts the action between two observations.
Following \citet{jang2025dreamgen,kim2026robocurate}, we use a 0.1B Diffusion Transformer with a SigLIP-2~\citep{tschannen2025siglip} vision encoder as our IDM, trained with a flow-matching objective that denoises the sequence of intermediate actions conditioned on a pair of input frames.
For GR-1, we use a publicly available pre-trained checkpoint\footnote{\url{https://huggingface.co/seonghyeonye/IDM_gr1}}, while for ALLEX we train the IDM from scratch on our in-house teleoperation data (see \Cref{subsec:in-house-datasets}) with an action horizon of $H+1=20$, a batch size of $256$, and $60$K gradient steps.
This produces synthetic data that pairs generated videos with pseudo-action labels for downstream policy training.

\paragraph{Video Quality Filtering}
We use the Gemini API \citep{googleGeminiAPI} to assess each generated video on two criteria, feeding 16 uniformly sampled frames for instruction following and 8 frames for trajectory plausibility.

\paragraph{Motion-Consistency Filtering}
We train the attentive probe with a binary cross-entropy loss on 16-frame clips at $256{\times}256$ resolution with a temporal stride of 4. We use the AdamW optimizer \citep{loshchilov2017decoupled} with a batch size of 32 and a learning rate of $10^{-4}$.

\newpage

\section{Reinforcement Learning (RL) Details}
\label{appendix:RL}

\paragraph{Critic Training}
We use \texttt{gemma3-4b-it}~\citep{team2025gemma} as the VLM backbone for progress estimation (critic). 
We apply LoRA~\citep{hu2022lora} to all trainable components (full target), including the vision encoder, with rank $r = 128$.
We train the model for 1 epoch on target-task success demonstrations only, with a learning rate of $1 \times 10^{-4}$.

\paragraph{Policy Training}
We train a pre-trained RLDX-1 for up to 30K steps (approximately 3 epochs) using AdamW optimizer \citep{loshchilov2017decoupled} with a learning rate of $1\times10^{-4}$ with a global batch size of 128 and a cosine schedule preceded by linear warmup over the first 5\% of training.
We freeze the vision encoder throughout policy training. We train the policy to generate an action chunk of size 40.

\begin{algorithm}[h]
    \caption{RECAP Post-Training with VLM Critic for RLDX-1}
    \label{alg:vlm_advantage_training}
    \begin{algorithmic}[1]
        \Require Demonstration dataset $\mathcal{D}_l$, number of iterations $N$
        \Ensure  Pre-trained policy $\pi$, VLM critic $V$
        \State Train $V$ on $\mathcal{D}_l$            \Comment{Initial value training}
        \State $A \leftarrow V(\mathcal{D}_l)$         \Comment{Annotate advantages}
        \State Train $\pi$ on $\mathcal{D}_l$ with advantage labels $A$
        \For{$i = 1, \cdots, N$}
            \State $\mathcal{D}_l \leftarrow \mathcal{D}_l \cup \pi.\texttt{rollout}()$
                \Comment{Collect rollouts}
            \State $\mathcal{D}_{\text{succ}} \leftarrow \{(\tau, y) \in \mathcal{D}_l \mid y = \text{success}\}$
            \State Train $V$ on $\mathcal{D}_{\text{succ}}$
                \Comment{Refine value on successes}
            \State $A \leftarrow V(\mathcal{D}_l)$     \Comment{Re-annotate trajectories}
            \State Train $\pi$ on $\mathcal{D}_l$ with $A$
                \Comment{Refine policy}
        \EndFor
    \end{algorithmic}
\end{algorithm}

\paragraph{Policy Inference}
At inference time, we execute only the first 20 predicted actions in each chunk, and then we re-query the policy to generate an action chunk.
We apply re-planning via the Real Time Chunking (RTC; \citealt{black2025real}) under the same configuration used in all other experiments in this work.

\newpage

\section{Parameter-Efficient Fine-Tuning}

We investigate how RLDX-1 can be adapted to downstream tasks with less compute, using a parameter-efficient fine-tuning approach.
\Cref{tab:peft_lora} reports success rate, trainable parameter count, and peak VRAM for full fine-tuning and three PEFT variants on Robocasa Kitchen, where ``Full FT'' refers to fully fine-tuning the top-4 layers of the backbone VLM as established in the main paper.
First, applying LoRA to the action model alone, while fully fine-tuning the top-4 backbone layers, recovers full fine-tuning performance, matching the 62.67\% success rate at rank 64 while halving the trainable parameter count (1{,}150.5M vs.\ 2{,}376.0M).
Second, freezing the backbone severely limits adaptation: the best frozen configuration reaches only 36.42\%, a 26.25-point gap from full fine-tuning, indicating that backbone updates are essential when the downstream task distribution diverges from pre-training.
Third, applying LoRA to both the top-4 backbone layers and the action model offers the best efficiency-and-performance trade-off: at rank 64, it achieves 55.33\% success while training only 5.72\% of total parameters and using 41.3\% of the batch-32 VRAM of full fine-tuning (35.93 GiB vs.\ 87.10 GiB). 
At batch size 1, the same configuration fits within 24 GiB, making single-GPU adaptation on consumer-grade hardware practical.

\begin{table}[ht]
    \centering
    \caption{\textbf{Parameter-efficient fine-tuning evaluation.}
        We compare full fine-tuning against PEFT configurations applying Low-Rank Adaptation (LoRA;~\citealt{hu2022lora}) to the learnable top-4 layers of the backbone VLM and the action model.
        Here, ``Full FT'' denotes fully fine-tuning the top-4 backbone layers, consistent with the setting used throughout the main paper.
        Success rate is averaged over 24 tasks $\times$ 50 episodes with 3 multi-frame observation views and video length 4 (12 frames total).
        Experiments are conducted on a single NVIDIA H200 GPU.
    }
    \label{tab:peft_lora}
    \begin{tabular}{llcccc}
        \toprule
        Backbone VLM & Action Model & Success Rate & Train Params ($\times 10^6$) & VRAM (batch 32) & VRAM (batch 1) \\
        \midrule
        \textbf{Full FT} & \textbf{Full FT} & \textbf{62.67\%} & \textbf{2{,}376.0} & \textbf{87.10 GiB} & \textbf{56.77 GiB} \\
        \midrule
        Frozen & LoRA, rank=8 & 21.25\% & \phantom{0,}361.7 & 27.22 GiB & 22.94 GiB \\
        Frozen & LoRA, rank=64 & 36.42\% & \phantom{0,}378.4 & 27.48 GiB & 23.31 GiB \\
        \midrule
        Full FT & LoRA, rank=8 & 60.17\% & 1{,}133.7 & 76.33 GiB & 36.83 GiB \\
        Full FT & LoRA, rank=64 & 62.67\% & 1{,}150.5 & 76.60 GiB & 37.20 GiB \\
        \midrule
        LoRA, rank=8 & LoRA, rank=8 & 45.75\% & \phantom{0,}364.1 & 35.39 GiB & 23.11 GiB \\
        \textbf{LoRA, rank=64} & \textbf{LoRA, rank=64} & \textbf{55.33\%} & \phantom{0,}\textbf{397.8} & \textbf{35.93 GiB} & \textbf{23.71 GiB} \\
        \bottomrule
    \end{tabular}
\end{table}

\newpage

\section{Test-time Sampling}
\subsection{Test-time Sampling Technique}
Beyond RL training for last-mile performance, a natural complementary direction is inference-time reasoning via Best-of-$N$ (BoN) sampling: drawing multiple action chunk candidates from the policy and executing the one scored highest by a critic~\citep{chen2023offline,nakamoto2024steering}. DEAS~\citep{kim2025deas} instantiates this pattern by offline-training a chunk-level critic on Vision-Language-Action model (VLA) features.
We investigate whether such a test-time critic can provide an orthogonal boost on top of the RL post-training.
Since we do not perform additional on-policy fine-tuning, the critic must remain stable under the distribution mismatch between its offline training data and on-policy samples from the post-RL policy.
To this end, we adopt an Implicit Q-Learning (IQL; \citealt{kostrikov2022offline})-style objective that biases the critic toward conservative, in-distribution values without requiring full action-space coverage.
We further extend the critic by conditioning VLA vision-language features and scoring action chunks $\rva_{t:t+H}$ instead of single-step actions~\citep{zhao2023learning}.
We use both an in-chunk discount $\gamma_1$ and a chunk-wise discount $\gamma_2$ following the temporally-extended critic formulation of~\citep{li2025qchunking}.

At inference, we raise RLDX-1's noise sampling temperature to 1.5--2.0 to diversify candidate chunks and execute the one with the highest $Q$-value among the $N$ samples. 
For the VLA experiments, we set the expectile to $\tau{=}0.7$ and use discounts $\gamma_1{=}0.9$ (in-chunk) and $\gamma_2{=}0.99$ (chunk-level), with a universal support type for the distributional critic head and an action chunk length $H+1=16$ matched to the GR00T~N1.6 chunk size.
We optimize with AdamW at learning rate $1{\times}10^{-4}$ and batch size $64$, for $30$K steps on RoboCasa and $10$K on the real-robot setup.

\subsection{Implementation Details}

\paragraph{Architecture}
We instantiate the value function $V$ and Q-function $Q$ on top of a frozen RLDX-1 backbone post-trained on RECAP$_1$.
For each observation, we average-pool the final-layer backbone token states along the sequence axis to obtain a single visual-language (VL) representation.
We then project it into a shared 64-dimensional VL embedding using an embodiment-conditioned 4-layer MLP.
The resulting representation is concatenated with the proprioceptive state $\rvs_t$ to form the input to the value network.
The value function $V$ processes this concatenated vector with a BRONet consisting of 4 pre-norm residual blocks of width 256.
The Q-function $Q$ additionally takes an action chunk $\rva_{t:t+H}$, enabling evaluation of temporally extended actions.
We employ a twin-critic setup and aggregate predictions via $\min(Q_1,Q_2)$ for both TD targets and BoN scoring.
Critically, $Q$ uses the same normalized action representations, ensuring consistency between training and inference without requiring additional encoding or decoding.
The embodiment-conditioned VL encoder supports up to 36 embodiment slots, shared across tasks.

\paragraph{Optimization}
Backbone features are extracted once from the post-RECAP$_1$ RLDX-1 checkpoint and cached on disk, so critic training never backpropagates through the VLM.
We optimize with AdamW at learning rate $3{\times}10^{-5}$ (chosen from a sweep over ${1,3}{\times}{10^{-5},10^{-4}}$), weight decay $10^{-5}$, batch size 64, $10\text{k}$ gradient steps, $5\%$ linear warmup, and gradient clipping at 1.0.

\paragraph{RL hyperparameters}
We follow DEAS: in-chunk discount $\gamma_1{=}0.9$, chunk-level discount $\gamma_2{=}0.99$, expectile $\tau{=}0.7$, and target-critic Polyak rate $0.005$.
Rewards are shifted to $r\leftarrow r-1$ so that the HL-Gaussian critic head covers $[-1/(1-\gamma_2),,0]=[-100,0]$ with $101$ atoms and Gaussian smoothing $\sigma{=}0.75$ bin-widths; the categorical logits are decoded to scalar $Q$ and $V$ values by taking the expectation against the bin centers.

\paragraph{BoN inference}
For BoN inference, the policy's flow-matching initial noise is multiplicatively rescaled by a temperature $T\in[1.5,2.0]$ (sweet spot $T{=}1.5$); the backbone and action head are run once on a $B{\cdot}N$-tiled observation batch so that all $N$ candidate chunks share a single backbone forward pass.
Concretely, following SAC-style action sampling, we draw the flow-matching initial noise from $\mathcal{N}(0, T^{2}I)$ instead of the standard $\mathcal{N}(0, I)$.

\newpage

\subsection{Experimental results}
\begin{figure}[ht]
    \centering
    \includegraphics[width=1\linewidth]{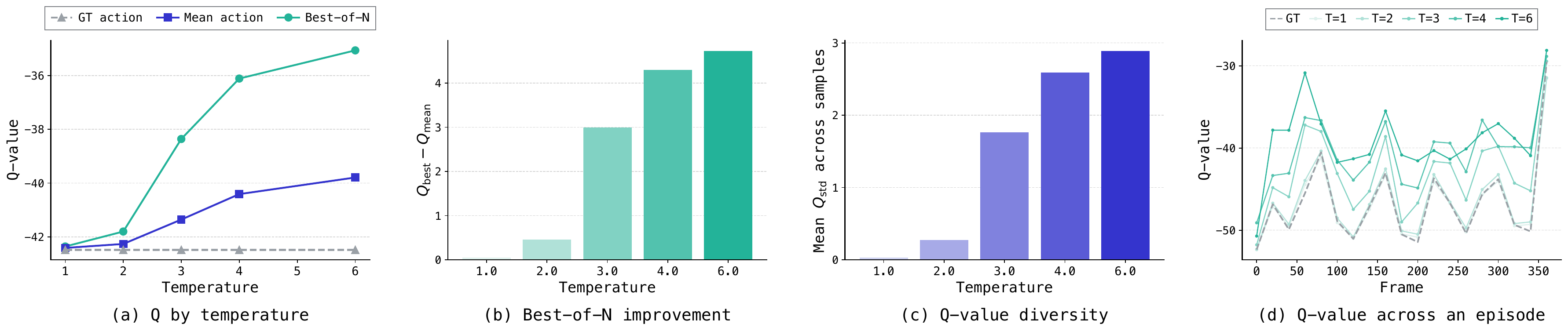}
    \vspace{-1.5em}
    \caption{
    On the offline dataset, we sampled $N=10$ from the RECAP$_3$ policy and measured the Q value to see whether test-time sampling would be effective. 
    (a) Picking the best action chunk according to Q-value makes a difference, and (b), (c) the gap becomes clear as temperature increases. 
    (d) Q-value versus frame of an \textit{Light Bulb Twisting} episode shows that Q improvement by temperature takes place uniformly within an episode.}
    \label{fig:episode_avg_Q_vs_temperature}
\end{figure}

We investigate whether test-time compute can be traded for task performance by pairing the RLDX-1 policy with a learned Q-critic and performing best-of-$N$ action selection: at every decision step, we sample $N$ candidate action chunks from the policy at temperature $T$, score each with the critic, and execute the top-ranked one.
To calibrate the sampling hyperparameters, we first conducted an offline analysis on the RECAP$_3$ policy, measuring both the action diversity across the $N{=}10$ samples and the resulting spread in Q-values as a function of temperature.
In \Cref{fig:episode_avg_Q_vs_temperature}, we find that action diversity grew steadily with $T$, but the Q-value gap between the best-of-$N$ chunk and the mean only became clearly separated at $T > 2$, and per-episode traces confirmed that this improvement was distributed fairly uniformly across frames rather than being driven by a few outlier steps.
Operating the physical robot at such high temperatures was, however, unsafe due to occasional out-of-distribution actions, so we fixed $T = 1.5$ as a practical compromise between exploitable diversity and safe execution.

\begin{figure}[ht]
    \centering
    \includegraphics[width=0.9\linewidth]{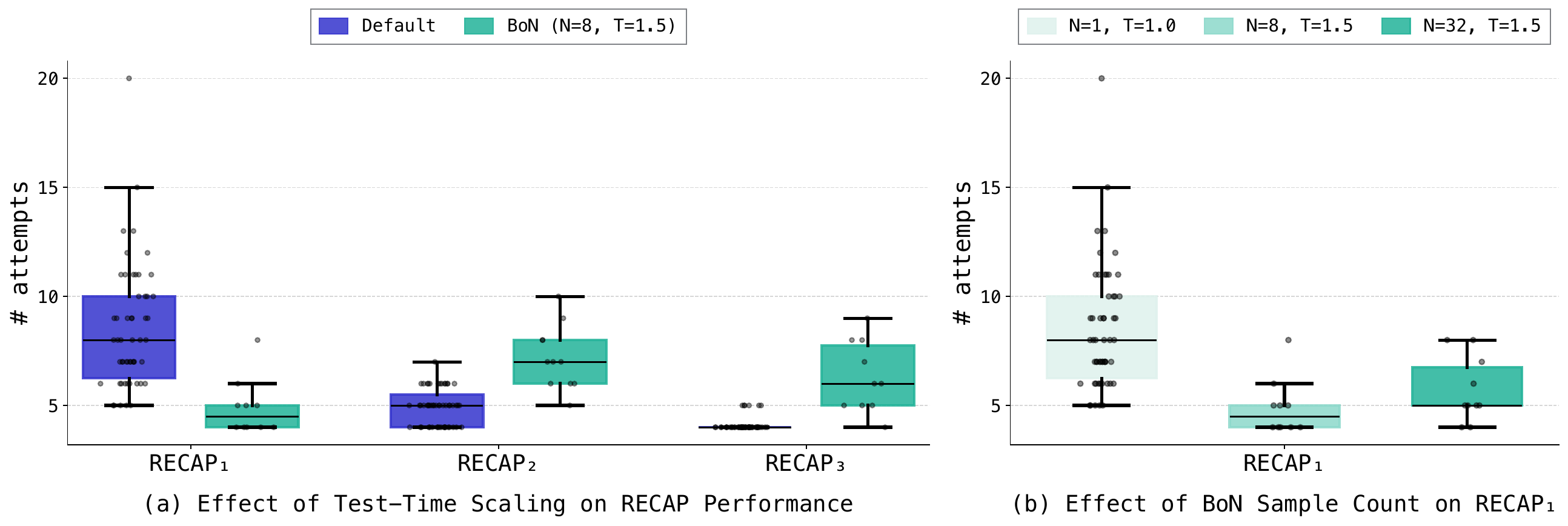}
    \vspace{-0.75em}
    \caption{
    Test-time sampling result of RL-trained RLDX-1 checkpoints (RECAP$_1$ to RECAP$_3$).
    We measure the number of attempts to complete \textit{Light Bulb Twisting} task.
    Test-time sampling ($N=8$, and $T=1.5$) improves RECAP$_1$ policy nearly to RECAP$_2$ policy performance, while it degrades the performance of RECAP$_2$ and RECAP$_3$.
    We checked that increasing the number of samples from 8 to 32 for RECAP$_1$ does not further improve the performance.
    We find that test-time-sampling helps explore more which might harm near-converged policies.}
    \label{fig:BoN}
\end{figure}

We evaluate BoN sampling ($N = 8$, $T = 1.5$) across RECAP$_1$ to RECAP$_3$ on the \textit{Light Bulb Twisting} task.
As shown in \Cref{fig:BoN} (a), the effect strongly depends on the degree of policy convergence. For RECAP$_1$, test-time sampling reduces the mean number of attempts from $8.5 \pm 2.8$ to $4.9 \pm 1.3$ ($-3.6$), performing better than RECAP$_2$.
In contrast, applying the same procedure to RECAP$_2$ and RECAP$_3$ degrades performance, increasing attempts by $+2.3$ and $+2.2$, respectively. 
In addition, as shown in \Cref{fig:BoN} (b), scaling $N$ from $8$ to $32$ on RECAP$_1$ does not yield further gains ($4.9 \rightarrow 5.7$ attempts), indicating that the benefit saturates quickly and is not simply a matter of wider search.
Taken together, these results suggest that test-time sampling acts primarily as an exploration mechanism.
It improves performance for less converged policies by recovering missing modes that the critic can exploit, but degrades performance for well-converged policies by introducing stochasticity that moves actions away from the optimum.
This behavior is consistent with similar observations in other domains~\citep{yue2025does}.
BoN is therefore complementary to, rather than a substitute for, additional RL training.

\newpage

\section{Kernel Optimization}
\label{appendix:kernel_opt_list}
In \Cref{tab:kernel_optimization_table}, we describe the detailed kernel optimization results.

\begin{table}[ht]
\caption{
Fused kernels used in our system.
Here, $\rvh_{\mathrm{in}}^{(\ell)}$ and $\rvh_{\mathrm{out}}^{(\ell)}$
denote the input and output hidden states of layer $\ell$, respectively.
We compare the original unfused operator sequence with the corresponding fused kernel.
}
\label{tab:kernel_optimization_table}
\centering
\small
\renewcommand{\arraystretch}{1.3}

\resizebox{\linewidth}{!}{%
\begin{tabular}{
@{}>{\centering\arraybackslash}p{0.27\textwidth}
>{\centering\arraybackslash}p{0.27\textwidth}
>{\centering\arraybackslash}p{0.46\textwidth}@{}}
\toprule
\textbf{Kernel} & \textbf{Unfused Operations} & \textbf{Fused Operations} \\
\midrule

\texttt{fused\_vision\_attention} &
$\begin{gathered}
\rvq' = \text{RoPE}(\rvq) \\
\rvk' = \text{RoPE}(\rvk) \\
\rvh_{\mathrm{out}}^{(\ell)} = \text{Attn}(\rvq', \rvk', \rvv)
\end{gathered}$ &
$\rvh_{\mathrm{out}}^{(\ell)} =
\text{FusedVisAttn}(\rvq,\rvk,\rvv)$
\\[4pt]\cmidrule(lr){1-3}

\texttt{fused\_llm\_attention} &
$\begin{gathered}
\rvq' = \text{RoPE}(\text{RMSNorm}(\rvq)) \\
\rvk' = \text{RoPE}(\text{RMSNorm}(\rvk)) \\
\rvh_{\mathrm{out}}^{(\ell)} = \text{Attn}(\rvq', \rvk', \rvv)
\end{gathered}$ &
$\rvh_{\mathrm{out}}^{(\ell)} =
\text{FusedLLMAttn}(\rvq,\rvk,\rvv)$
\\[4pt]\cmidrule(lr){1-3}

\texttt{fused\_add2\_layernorm} &
$\begin{gathered}
\rvh_{\mathrm{res}}^{(\ell)} =
\rvh_{\mathrm{out}}^{(\ell)} + \rvh_{\mathrm{in}}^{(\ell)} \\
\rvh_{\mathrm{in}}^{(\ell+1)} =
\text{LayerNorm}(\rvh_{\mathrm{res}}^{(\ell)})
\end{gathered}$ &
$\rvh_{\mathrm{in}}^{(\ell+1)} =
\text{FusedAddLayerNorm}(\rvh_{\mathrm{out}}^{(\ell)},\rvh_{\mathrm{in}}^{(\ell)})$
\\[4pt]\cmidrule(lr){1-3}

\texttt{fused\_add2\_rmsnorm} &
$\begin{gathered}
\rvh_{\mathrm{res}}^{(\ell)} =
\rvh_{\mathrm{out}}^{(\ell)} + \rvh_{\mathrm{in}}^{(\ell)} \\
\rvh_{\mathrm{in}}^{(\ell+1)} =
\text{RMSNorm}(\rvh_{\mathrm{res}}^{(\ell)})
\end{gathered}$ &
$\rvh_{\mathrm{in}}^{(\ell+1)} =
\text{FusedAddRMSNorm}(\rvh_{\mathrm{out}}^{(\ell)},\rvh_{\mathrm{in}}^{(\ell)})$
\\[4pt]\cmidrule(lr){1-3}

\texttt{fused\_add3\_rmsnorm} &
$\begin{gathered}
\rvh_{\mathrm{res}}^{(\ell)} =
\rvh_{\mathrm{out}}^{(\ell)} + \rvh_{\mathrm{in}}^{(\ell)} + \rvh_{\mathrm{ds}}^{(\ell)} \\
\rvh_{\mathrm{in}}^{(\ell+1)} =
\text{RMSNorm}(\rvh_{\mathrm{res}}^{(\ell)})
\end{gathered}$ &
$\rvh_{\mathrm{in}}^{(\ell+1)} =
\text{FusedAdd3RMSNorm}(\rvh_{\mathrm{out}}^{(\ell)},\rvh_{\mathrm{in}}^{(\ell)},\rvh_{\mathrm{ds}}^{(\ell)})$
\\[4pt]\cmidrule(lr){1-3}

\texttt{fused\_memory\_attention} &
$\begin{gathered}
\rvq' = \text{RoPE}(\rvq) \\
\rvk' = \text{RoPE}(\rvk) \\
\rvh_{\mathrm{out}}^{(\ell)} = \text{Attn}(\rvq', \rvk', \rvv)
\end{gathered}$ &
$\rvh_{\mathrm{out}}^{(\ell)} =
\text{FusedMemAttn}(\rvq,\rvk,\rvv)$
\\[4pt]\cmidrule(lr){1-3}

\texttt{grouped\_swiglu} &
$\begin{gathered}
\rvy_1 = \text{SwiGLU}(\rvz_1) \\
\rvy_2 = \text{SwiGLU}(\rvz_2)
\end{gathered}$ &
$(\rvy_1,\rvy_2) =
\text{GroupedSwiGLU}(\rvz_1,\rvz_2)$
\\[4pt]\cmidrule(lr){1-3}

\texttt{fused\_mlp\_swiglu} &
$\rvy = \text{SwiGLU}(\rvz)$ &
$\rvy = \text{FusedSwiGLU}(\rvz)$
\\

\bottomrule
\end{tabular}}
\end{table}

\newpage

\section{Evaluation \& Analysis}

\subsection{Simulation Benchmark Details}
\label{appendix:simul_setup}

We evaluate our method on the following simulation benchmarks.

\vspace{-0.5em}
\begin{itemize}[leftmargin=*,itemsep=0mm]
\item \textbf{LIBERO} \citep{liu2023libero} is a single-arm tabletop manipulation benchmark built on a Franka Research 3 robotic arm with a parallel gripper. It contains 40 tasks grouped into four sub-benchmarks (Spatial, Object, Goal, and Long). 
We use a fixed front-view camera and a wrist camera of 256$\times$256 resolution. We train RLDX-1 on the concatenated training datasets of each sub-benchmark, evaluate the model 50 times for each task, and report the average success rates. 
For baselines, we reproduce GR00T N1.6 under the same setup as RLDX-1 by following its official implementation; other results are taken from the respective official repositories.

\item \textbf{LIBERO-Plus} \citep{fei2025libero} is a benchmark built on top of LIBERO that measures the robustness under a diverse set of perturbations. Specifically, it evaluates robustness to changes in object layout, camera viewpoint, robot initial state, language instruction, light condition, background texture, and sensor noise. 
Results for $\pi_0$ and $\pi_0$-FAST are taken from the original paper~\citet{fei2025libero}; the remaining methods, including RLDX-1, are evaluated on all 10{,}300 perturbation tasks using the same checkpoints as in LIBERO.

\item \textbf{SIMPLER Google-VM/VA} \citep{li2024evaluating} is a single-arm gripper benchmark based on the Google robot with an ego-centric camera view. It includes 4 tasks (Move Near, Pick Coke Can, Open Drawer, and Close Drawer). We follow the common setup in \citet{li2024evaluating}, where RLDX-1 is trained on the Fractal dataset \citep{brohan2022rt} and evaluated on each task. For Visual Matching (VM), we evaluate the model 200 times for each task by varying the random seed of the benchmark, and report the average success rates. For Variant Aggregation (VA), we evaluate RLDX-1 100 times per task perturbation and report the average success rates. Baseline scores borrow from \citet{community2026starvla,yang2025vlaser,nvidia2025gr00t16}.

\item \textbf{SIMPLER WidowX} \citep{li2024evaluating} is a single-arm gripper benchmark based on the WidowX robot with a single third-person camera view. It includes 4 tasks (Spoon on Towel, Carrot on Plate, Stack Cube, and Put Eggplant in Basket). We follow the common setup in \citet{li2024evaluating} that trains RLDX-1 on the BridgeV2 dataset \citep{walke2023bridgedata}, evaluates the model 200 times for each task by varying the random seed of the benchmark, and reports the average success rate. Baseline scores borrow from \citet{community2026starvla,chen2025villa,nvidia2025gr00t16}.

\item \textbf{RoboCasa Kitchen} \citep{nasiriany2024robocasa} is a single-arm kitchen manipulation benchmark built on a mobile Panda manipulator (PandaOmron) with a gripper. It includes 24 tasks about kitchen skills such as pick-and-place, opening and closing doors and drawers, and appliance control. We use two fixed external camera views, left and right, together with a wrist camera, all at a resolution of 256$\times$256.
We train RLDX-1 on a dataset constructed by concatenating 300 machine-generated demonstrations from each task, and report the average success rate over 50 episodes per task. For baselines, GR00T N1.6 results are taken from the official repository, and the other scores borrow from \citet{kim2025contrastive,jang2025contextvla}.

\item \textbf{GR-1 Tabletop} \citep{bjorck2025gr00t} is a GR-1 humanoid tabletop manipulation benchmark. The benchmark contains 24 tabletop tasks, including 18 object rearrangement tasks and 6 articulated object manipulation tasks. We use an ego-centric camera of 256$\times$256 resolution, following \citet{bjorck2025gr00t}. We train RLDX-1 on a dataset constructed by concatenating 1,000 machine-generated demonstrations from each task, and report the average success rates over 50 episodes per task. For baselines, GR00T-variant results are taken from the official repository, while $\pi_0$-variant results are reproduced by strictly following their official implementation under the same setup as RLDX-1.

\item \textbf{RoboCasa365} \citep{nasiriany2026robocasa365} is a large-scale household manipulation benchmark that extends RoboCasa Kitchen to a broader distribution of everyday kitchen tasks and environments. Following the official RoboCasa365 leaderboard protocol, we train RLDX-1 on the 300-task human pre-training dataset, which provides 100 demonstrations per task. We then evaluate the model on the 50-task multi-task benchmark across Atomic-Seen, Composite-Seen, and Composite-Unseen splits, and report the average success rate. For baselines, $\pi_{0}$-FAST results are reproduced by strictly following their official implementation under the same setup as RLDX-1, and the other scores borrow from the RoboCasa365 leaderboard \citep{nasiriany2026robocasa365}.
\end{itemize}

\newpage

\paragraph{Implementation Details}

We evaluate the pre-trained RLDX-1 by fine-tuning it on each benchmark.
Following the pre-training implementation, we freeze the vision encoder and the Large-Language-model (LLM) backbone, except for the top four layers of the LLM backbone.
Unless otherwise specified, we train the model for 60K steps with a global batch size of 1024 using AdamW optimizer \citep{loshchilov2017decoupled} with a learning rate of $1\times10^{-4}$ and a cosine schedule preceded by linear warmup over the first 5\% of training. 
Since a single hyperparameter setting may not be optimal across simulation benchmarks with different dataset scales, trajectory lengths, task distributions, and robot embodiments, we follow \citet{intelligence2025pi_,nvidia2025gr00t16} and perform benchmark-specific hyperparameter tuning. For each benchmark, we report the best-performing configuration. 
Specifically, for the number of iterations and batch size, we use benchmark-specific settings for three cases: LIBERO (and LIBERO-Plus) uses a global batch size of 256, SIMPLER Google-VM/VA is trained for 20K steps, and RoboCasa365 follows the official implementation with 250K training steps and a global batch size of 196. 
We additionally observe that proprioceptive state dropout is beneficial for SIMPLER Google-VM/VA and GR-1 Tabletop. Accordingly, we set the state dropout ratio to 0.5 for these benchmarks. For evaluation, we find that using more denoising steps is helpful for SIMPLER WidowX and RoboCasa Kitchen, and therefore use 10 denoising steps for action generation. In addition, we use fixed denoising timesteps with interval $1/T$ for all benchmarks except GR-1 Tabletop. For GR-1 Tabletop, we find that sampling timesteps from the same timestep distribution used during training, \ie $\mathrm{Beta}(1.5, 1.0)$, leads to better performance (58.7\% vs 58.0\%), while yielding comparable or slightly lower performance on the other benchmarks (within 1\%). We further use only a partial action horizon from the 16 generated action chunks: 2 chunks for SIMPLER and 8 chunks for LIBERO, LIBERO-PLUS, and GR-1 Tabletop.
We summarize the benchmark-specific hyperparameter setup for each simulation benchmark in \Cref{tab:simul-hparams}.

\begin{table}[ht]
\centering
\caption{Benchmark-specific training and evaluation hyperparameters for simulation experiments.}
\label{tab:simul-hparams}
\resizebox{\linewidth}{!}{%
\begin{tabular}{lcccccc}
\toprule
 & \shortstack{LIBERO \& \\LIBERO-Plus} & \shortstack{SIMPLER\\Google-VM \& VA} & \shortstack{SIMPLER\\WidowX} & \shortstack{RoboCasa\\Kitchen} & \shortstack{GR-1\\Tabletop} & \shortstack{\\RoboCasa365} \\
\midrule
\textit{Training} \\
Iterations & 60,000 & 20,000 & 60,000 & 60,000 & 60,000 & 250,000\\
Batch size & 256 & 1024 & 1024 & 1024 & 1024 & 192 \\
State dropout ratio & 0.0 & 0.5 & 0.0 & 0.0 & 0.5 & 0.0 \\
\midrule
\textit{Inference} \\
Denoising steps & 4 & 4 & 10 & 10 & 4 & 4 \\
Denoising timestep sampling & Fixed & Fixed & Fixed & Fixed & Beta(1.5, 1.0) & Fixed \\
Action execution horizon & 8 & 2 & 2  & 16 & 8 & 16 \\
\bottomrule
\end{tabular}}
\end{table}

\newpage

\subsection{OpenArm Experiments Details}
\label{appendix:openarm_details}

\paragraph{Task and Evaluation Details}
All tasks are formulated as language-conditioned pick-and-place tasks in a three-object tabletop scene. For training, we collect pick-and-place demonstrations from four object categories, including bottle, snack, cup, and doll, with four object instances per category. In each episode, a target object and two distractor objects from different categories are placed at three possible initial locations, and the robot is instructed to move the target object to a specified goal location. Some configurations require bimanual coordination, especially when the target object and goal location are placed on opposite sides of the workspace. For evaluation, we conduct 24 trials per task. For \textit{Directional PnP (Shelf)}, we evaluate two different target object categories with 24 trials per category, resulting in 48 evaluation trials in total.

\paragraph{Training Demonstrations}
We collect demonstrations for four task families, each associated with one target object category: \textit{Bottle to Shelf}, \textit{Snack to Shelf}, \textit{Cup to Dish Rack}, and \textit{Doll to Box}. 
The first two task families correspond to \textit{Directional PnP (Shelf)}, while \textit{Cup to Dish Rack} and \textit{Doll to Box} correspond to \textit{Directional PnP (Dish Rack)} and \textit{Basic PnP}, respectively.
For \textit{Bottle to Shelf}, we use four bottle instances, three initial table locations, four target shelf locations, and two trials per configuration, resulting in 96 demonstrations. 
\textit{Snack to Shelf} follows the same data collection protocol using four snack instances, yielding 96 demonstrations. 
For \textit{Cup to Dish Rack}, we use four cup instances, three initial table locations, three target dish rack slots, and three trials per configuration, resulting in 108 demonstrations. 
For \textit{Doll to Box}, we use four doll instances, three initial table locations, and four trials per configuration, resulting in 48 demonstrations. 
In total, the training set contains 348 demonstrations across the four task families.

\paragraph{Hardware Setup}
We use the OpenArm humanoid\footnote{\url{https://openarm.dev/}}, which consists of two 7-DoF arms. Each arm is equipped with a 6-DoF Inspire RH56F1 hand, and the robot head is mounted on a 2-DoF neck that supports pitch and yaw motions. A ZED 2i stereo camera is mounted on the neck, and its left and right egocentric views are used as visual inputs throughout all experiments. For data collection, we use ExoArm-7, a wearable exoskeleton compatible with OpenArm, to control the two arms of the robot, and we use the Manus Pro gloves to capture finger-tip positions, from which finger joint angles are computed via inverse kinematics. The neck joint is kept fixed throughout data collection.

\paragraph{Implementation Details}
We evaluate the pre-trained RLDX-1 by fine-tuning it on the concatenated training datasets of each in-domain task. Following the pre-training implementation, we freeze the vision encoder and the Large-Language-model (LLM) backbone, except for the top four layers of the LLM backbone. 
We train the model for 30K steps with a global batch size of 64 using AdamW optimizer~\citep{loshchilov2017decoupled} with a learning rate of $1\times 10^{-4}$ and a cosine schedule preceded by linear warmup over the first 5\% of training. We set the state dropout ratio to 0.3.
For the baselines, we strictly follow the official implementations and train them for 30K steps with a global batch size of 64, matching the training steps and batch size used for RLDX-1.

\newpage

\subsection{ALLEX Experimental Details}
\label{appendix:allex_details}

\paragraph{Task and Evaluation Details} 
All tasks are formulated as language-conditioned manipulation tasks with target objects and task-specific environments.
For training, we exclude factors that are not directly relevant to the functional capabilities evaluated in each task, such as spatial generalization or language following.
For evaluation, we conduct 24 trials per task. For \textit{Conveyor Pick-and-Place}, we evaluate each of the four conveyor speeds with 6 trials.
For \textit{Object-in-Box Selection}, we evaluate each of the three positions with 8 trials.
For \textit{Card Slide-and-Pick}, we evaluate all trials at the lower of the two training positions to ensure hardware safety.

\paragraph{Training Demonstrations}
We collect demonstrations for each task under task-specific conditions.
For \textit{Conveyor Pick-and-Place}, we collect 40 demonstrations using two different speed settings (S1, S4).
For \textit{Object-in-Box Selection}, we collect 90 demonstrations, with 30 demonstrations for each of the three boxes.
For \textit{Card Slide-and-Pick}, we collect 72 demonstrations using two different desk height settings (80cm, 75cm).
For \textit{Pot-to-Cup Pouring}, we collect 62 demonstrations. For safety, small plastic balls are used instead of liquid (\eg coffee) in the pouring task.

\paragraph{Hardware Setup}
For ALLEX, a high-DoF upper-body humanoid, we use a total of 48 joints for manipulation, including dual 7-DoF arms, two 15-DoF five-finger hands, a 2-DoF waist, and a 2-DoF neck.
A head-mounted stereo camera providing egocentric views is the only camera used in the benchmark, without wrist or side cameras.
Joint torques estimated from motor currents are used as physical signals, using values from all 48 DoF.

\paragraph{Implementation Details}
We fine-tune the RLDX-1 model initialized from the mid-training stage for 30K steps per task with a global batch size of 128 using AdamW optimizer with a learning rate of $1\times10^{-4}$.
A cosine learning rate schedule with 5\% warmup is used.
For each task, only the module relevant to the target functionality is enabled during fine-tuning and evaluation, while the remaining modules are disabled.
We set the state dropout ratio to 0.8.
To facilitate smooth action execution of the ALLEX humanoid, real-time chunking~\citep{black2025real} is used as the default inference strategy.
For the baselines, we follow the training recipes of the official implementations and fine-tune the models for 30K steps under the same training setup as RLDX-1.

\newpage

\subsection{Franka Research 3 Experimental Details}
\label{appendix:franka_details}

\paragraph{Task and Evaluation Details}
All tasks are formulated as language-conditioned manipulation tasks with target objects and task-specific environments.
% For training,
For evaluation, we conduct 24 trials per task, except for \textit{Spin Tracking} and \textit{Pong Game}.
For \textit{Spin Tracking}, we assess performance over 96 interactions involving responses to clockwise or counterclockwise motion, with 24 trials for each of the 4 initial positions, and measure the success rate.
For \textit{Pong Game}, we assess performance over 54 interactions, each involving a response to the ball reaching the right wall, and measure the success rate; all evaluation interactions are unseen during training.
For \textit{Cup Swapping}, we measure the full success rate over twelve trials for each initial position.
For \textit{Shell Game}, we measure the selection success rate over 8 trials per position across 3 shell positions.
For \textit{Plug Insertion} and \textit{Egg Pick-and-Place}, we evaluate at 3 different positions and measure the success rate over 8 trials per position.

\paragraph{Training Demonstrations}
We collect demonstrations for each task under task-specific conditions.
For \textit{Spin Tracking}, we collect 64 demonstrations.
Each demonstration contains three interaction sequences with either clockwise or counterclockwise motion.
During training, we use four different initial object positions.
For \textit{Pong Game}, we collect 50 demonstrations.
Each demonstration contains 5 interaction sequences where the ball reaches the right wall.
For \textit{Cup Swapping}, we collect 60 demonstrations for 2 different initial positions. Each demonstration involves overlapping trajectories that can be confused without memory. 
For \textit{Shell Game}, we collect 60 demonstrations, with 20 demonstrations for each target cup.
For \textit{Plug Insertion}, we collect 100 demonstrations, with varying pick-up positions the plug. 
For \textit{Egg Pick-and-Place}, we collect 60 demonstrations with varying object positions.

\paragraph{Hardware Setup}
We follow the DROID \citep{khazatsky2024droid} setup that uses a 7-Degree-of-Freedom Franka Research 3 robotic arm equipped with a Robotiq 2F-85 gripper. Here, we mount an
AnySkin tactile sensor \citep{bhirangi2025anyskin} on the gripper, while the robot arm itself supports torque sensing at each joint. 
Specifically, for tactile signals, we use the signals from the left gripper that provide a 15-dimensional feature vector, obtained from five sensing units, with each measuring a 3-dimensional force vector (\ie $x, y,z$). For the torque signals, the robot provides a 7-dimensional vector (one scalar torque value per joint) measured in Newton-meters. 
We visualize the hardware setup in {\Cref{figure:hardware_overview}}.

\paragraph{Implementation Details}
We post-train the mid-trained RLDX-1 for 30K steps per task, with a global batch size of 64 using the AdamW optimizer with a learning rate of $1\times10^{-4}$. A cosine learning rate schedule with 5\% warmup is used.
For each task, only the module relevant to the target functionality is enabled during fine-tuning and evaluation, while the remaining modules are disabled.
We set the state dropout ratio to 0.3. For the baselines, we strictly follow the official implementations and train them for 30K steps with a global batch size of 64, matching the training steps and batch size used for RLDX-1.

\newpage

\subsection{Additional ALLEX Experiments}
\label{appendix:additional_allex_exp}
\paragraph{Task and Evaluation Details}
We evaluate on the grasping subtask of \textit{Pot-to-Cup Pouring}, a bimanual manipulation task where ALLEX humanoid sequentially grasps a paper cup and a coffeepot from a table.
To assess robustness to spatial variation, we evaluate across 3 cup positions (spaced ${\sim}$5\,cm apart) and 4 coffeepot positions (spaced ${\sim}$9\,cm apart), with 2 trials per configuration, for a total of 24 trials. Each trial is scored as a full success (1.0) if the robot grasps both the cup and the coffeepot, a partial success (0.5) if it grasps the cup only, and a failure (0.0) otherwise.

\paragraph{Training Data}
We first collect real \textit{Pot-to-Cup Pouring} demonstrations via teleoperation, with the cup and coffeepot placed at positions sampled from a Gaussian distribution ($\sigma{\approx}$10\,cm) around a nominal location. We additionally generate synthetic data through multi-turn rollouts of our ALLEX video generative model (see \Cref{sec:synth-data-details} for training details), where each turn reuses the final frames of the preceding turn as conditioning. We use two rounds of generation with subtask-specific prompts (\textit{grasp-and-tilt}, then \textit{pour-and-return}) to mitigate the length gap from real demonstrations while preserving realistic motion speed. We note that only video quality filtering is applied to this synthetic data, without task or scene augmentation.

\paragraph{Implementation Details}
We fine-tune the pre-trained RLDX-1 model for 30K steps with a global batch size of 128, using the AdamW optimizer with a learning rate of $1 \times 10^{-4}$ and a cosine learning rate schedule with 5\% warmup. We set the state dropout ratio to 0.4 and the action horizon to 40. For the co-finetuning configuration, we mix real and synthetic data at a 50/50 ratio and route each dataset to its corresponding embodiment slot via per-dataset embodiment tagging.

\paragraph{Results}
As shown in \Cref{tab:additional_allex_coffeepot_exp}, we find that co-finetuning with our synthetic data improves the overall success rate from 66.7\% to 83.3\%. The gain is primarily driven by a substantial reduction in failures (7 $\to$ 2), accompanied by an increase in full successes (9 $\to$ 14).
This indicates that our synthetic data enhances the spatial generalization of the policy by augmenting successful trajectories, enabling stable grasping across diverse positions of the cup and coffeepot.

\begin{table}[ht]
\centering
\caption{Effect of synthetic data on the \textit{Pot-to-Cup Pouring} grasping task. Each of the 24 trials is scored as 1.0 (full), 0.5 (partial), or 0.0 (failure); the \textit{Score} column reports the sum of trial scores.}
\label{tab:additional_allex_coffeepot_exp}
\begin{tabular}{lccccc}
\toprule
Fine-tuning Data & Full & Partial & Fail & Score & Success Rate \\
\midrule
Real           & 9  & 8  & 7 & 16.0\,/\,24 & 66.7\% \\
Real + Synth. & 14 & 8  & 2 & 20.0\,/\,24 & \textbf{83.3\%} \\
\bottomrule
\end{tabular}
\end{table}

\newpage

\subsection{Additional Simulation Results / Full Results}
\label{appendix:full_results}

\paragraph{Effect of Batch Size}
We here investigate how the training batch size affects the performance of simulation benchmarks. As shown in \Cref{tab:batch_size_ablation}, we find that training RLDX-1 with a larger batch size improves the performance.

\begin{table}[ht]
\centering
\caption{Performance comparison under different batch sizes.}
\begin{tabular}{c c cc}
\toprule
 Batch Size & LIBERO & Robocasa Kitchen & GR-1 Tabletop \\
\midrule
64   & 97.4 & 66.9 & 36.8 \\
256  & \textbf{97.8} & 69.6 & 53.2 \\
1024 & \textbf{-} & \textbf{70.6} & \textbf{58.7} \\
\bottomrule
\end{tabular}
\label{tab:batch_size_ablation}
\end{table}

\paragraph{Full Results}

In \Cref{tab:libero-appendix,tab:simpler-appendix,tab:robocasa-kitchen,tab:gr1-tabletop,tab:rc365}, we provide full results of RLDX-1 on simulation benchmarks.

\begin{table}[ht]
\centering
\caption{Full results of RLDX-1 on LIBERO \citep{liu2023libero} and LIBERO-Plus \citep{fei2025libero} benchmarks.}
\label{tab:libero-appendix}
\resizebox{\textwidth}{!}{%
\begin{tabular}{lccccccccccccc}
\toprule
& \multicolumn{5}{c}{LIBERO} & \multicolumn{8}{c}{LIBERO-Plus} \\
\cmidrule(lr){2-6}\cmidrule(lr){7-14}
Model & Long & spatial & object & goal & \textbf{Avg} & Camera & Robot & Language & Light & Background & Noise & Layout & \textbf{Total} \\
\midrule
\textbf{RLDX-1 (Ours)} & 95.3 & 98.0 & 99.3 & 98.4 & \textbf{97.8} & 75.0 & 91.8 & 92.2 & 97.1 & 95.5 & 80.3 & 80.7 & \textbf{86.7} \\
\bottomrule
\end{tabular}}
\end{table}

\begin{table}[ht]
\centering
\caption{Full results of RLDX-1 on SIMPLER benchmark \citep{li2024evaluating}.}
\label{tab:simpler-appendix}
\resizebox{\textwidth}{!}{%
\begin{tabular}{lccccc ccccc ccccc}
\toprule
& \multicolumn{5}{c}{Google-VM} & \multicolumn{5}{c}{Google-VA} & \multicolumn{5}{c}{WidowX}\\
\cmidrule(lr){2-6}\cmidrule(lr){7-11} \cmidrule(lr){12-16}
\shortstack{\\Model} & \shortstack{Move\\Near} & \shortstack{Close\\Drawer} & \shortstack{Open\\Drawer} & \shortstack{Pick\\Coke Can} & \shortstack{\\\textbf{Avg.}} & \shortstack{Move\\Near} & \shortstack{Close\\Drawer} & \shortstack{Open\\Drawer} & \shortstack{Pick\\Coke Can} & \shortstack{\\\textbf{Avg.}} & \shortstack{Carrot\\on Plate} & \shortstack{Put Eggplant\\in Basket} & \shortstack{Spoon\\on Towel} & \shortstack{Stack\\Cube} & \shortstack{\\\textbf{Avg.}} \\
\midrule
\textbf{RLDX-1 (Ours)} & 92.0 & 78.5 & 58.5 & 97.0& \textbf{81.5} & 90.9 & 93.2 & 32.3 & 93.4 & \textbf{77.4} & 83.0 & 52.0 & 88.5 & 64.0 & \textbf{71.9}\\
\bottomrule
\end{tabular}
}
\end{table}

\newpage

\begin{minipage}[ht]{0.48\textwidth}
  \centering
  \captionof{table}{Per-task success rates of RLDX-1 on the RoboCasa
  Kitchen benchmark~\citep{nasiriany2024robocasa}.}
  \label{tab:robocasa-kitchen}
  \vspace{-0.5em}
  \begin{tabular}{lc}
  \toprule
  Task & \textbf{RLDX-1 (Ours)} \\
  \midrule
  \multicolumn{2}{l}{\textit{Pick and place (8 tasks)}} \\
  \midrule
  Cabinet to Counter   & \phantom{0}44.0 \\
  Counter to Cabinet   & \phantom{0}64.0 \\
  Counter to Microwave & \phantom{0}38.0 \\
  Counter to Sink      & \phantom{0}72.0 \\
  Counter to Stove     & \phantom{0}66.0 \\
  Microwave to Counter & \phantom{0}26.0 \\
  Sink to Counter      & \phantom{0}76.0 \\
  Stove to Counter     & \phantom{0}56.0 \\
  \midrule
  \multicolumn{2}{l}{\textit{Open or close (6 tasks)}} \\
  \midrule
  Close Double Door & \phantom{0}94.0  \\
  Close Drawer      & \phantom{0}98.0  \\
  Close Single Door & 100.0 \\
  Open Double Door  & \phantom{0}62.0  \\
  Open Drawer       & \phantom{0}80.0  \\
  Open Single Door  & \phantom{0}80.0  \\
  \midrule
  \multicolumn{2}{l}{\textit{Others (10 tasks)}} \\
  \midrule
  Coffee Press Button  & \phantom{0}98.0  \\
  Coffee Serve Mug     & \phantom{0}82.0  \\
  Coffee Setup Mug     & \phantom{0}44.0  \\
  Turn Off Microwave   & \phantom{0}96.0  \\
  Turn Off Sink Faucet & \phantom{0}96.0  \\
  Turn Off Stove       & \phantom{0}22.0  \\
  Turn On Microwave    & \phantom{0}86.0  \\
  Turn On Sink Faucet  & 100.0 \\
  Turn On Stove        & \phantom{0}24.0  \\
  Turn Sink Spout      & \phantom{0}90.0  \\
  \midrule
  \textbf{Total (24 tasks)} & \phantom{0}\textbf{70.6} \\
  \bottomrule
  \end{tabular}
  \end{minipage}
  \hfill
  \begin{minipage}[ht]{0.48\textwidth}
  \centering
  \captionof{table}{Per-task success rates of RLDX-1 on the GR-1 Tabletop
benchmark~\citep{bjorck2025gr00t}.
}
\vspace{1.75em}
  \label{tab:gr1-tabletop}
  \begin{tabular}{lc}
  \toprule        
  Task & \textbf{RLDX-1 (Ours)} \\
  \midrule
  \multicolumn{2}{l}{\textit{Object rearrangement (18 tasks)}} \\
  \midrule
  Cutting Board to Pot           & 60.0 \\
  Cutting Board to Basket        & 66.0 \\
  Cutting Board to Tiered Basket & 46.0 \\
  Cutting Board to Pan           & 72.0 \\
  Cutting Board to Cardboard Box & 64.0 \\
  Placemat to Bowl               & 72.0 \\
  Placemat to Plate              & 70.0 \\
  Placemat to Basket             & 56.0 \\
  Placemat to Tiered Shelf       & 24.0 \\
  Plate to Pan                   & 48.0 \\
  Plate to Cardboard Box         & 64.0 \\
  Plate to Bowl                  & 56.0 \\
  Plate to Plate                 & 74.0 \\
  Tray to Tiered Shelf           & 36.0 \\
  Tray to Tiered Basket          & 54.0 \\
  Tray to Plate                  & 64.0 \\
  Tray to Cardboard Box          & 62.0 \\
  Tray to Pot                    & 74.0 \\
  \midrule
  \multicolumn{2}{l}{\textit{Articulated object manipulation (6 tasks)}}
  \\
  \midrule        
  Wine to Cabinet         & 56.0 \\
  Place Bottle to Cabinet & 64.0 \\
  Place Milk to Microwave & 48.0 \\
  Potato to Microwave     & 44.0 \\
  Cup to Drawer           & 60.0 \\
  Can to Drawer           & 74.0 \\
  \midrule
  \textbf{Total (24 tasks)} & \textbf{58.7} \\
  \bottomrule
  \end{tabular}
  \end{minipage}

\begin{table}[ht]
  \centering
  \caption{Per-task success rates of RLDX-1 on the RoboCasa365 benchmark \citep{nasiriany2026robocasa365}.}
  \label{tab:rc365}
  \vspace{-0.5em}
  \begin{subtable}[t]{0.32\textwidth}
  \centering
  \caption{Atomic-Seen}
  \label{tab:rc365-atomic-seen}
  \vspace{-0.5em}
  \resizebox{\linewidth}{!}{%
  \begin{tabular}{lc}
  \toprule
  Task & \textbf{RLDX-1 (Ours)} \\
  \midrule
  Close Blender Lid       & 22.0 \\
  Close Fridge            & 88.0 \\
  Close Toaster Oven Door & 60.0 \\
  Coffee Setup Mug        & 40.0 \\
  Navigate Kitchen        & 72.0 \\
  Open Cabinet            & 54.0 \\
  Open Drawer             & 74.0 \\
  Open Stand Mixer Head   & 92.2 \\
  PnP Counter to Cabinet  & 68.0 \\
  PnP Counter to Stove    & 74.0 \\
  PnP Drawer to Counter   & 76.0 \\
  PnP Sink to Counter     & 86.0 \\
  PnP Toaster to Counter  & 92.0 \\
  Slide Dishwasher Rack   & 76.0 \\
  Turn Off Stove          & 30.0 \\
  Turn On Electric Kettle & 80.0 \\
  Turn On Microwave       & 54.0 \\
  Turn On Sink Faucet     & 72.5 \\
  \midrule
  \textbf{Total (18 tasks)} & \textbf{67.3} \\
  \bottomrule
  \end{tabular}
  }
  \end{subtable}
  \hfill
  \begin{subtable}[t]{0.32\textwidth}
  \centering
  \caption{Composite-Seen}
  \label{tab:rc365-comp-seen}
  \vspace{0.88em}
  \resizebox{\linewidth}{!}{%
  \begin{tabular}{lc}
  \toprule
  Task & \textbf{RLDX-1 (Ours)} \\
  \midrule
  Deliver Straw           & 10.0 \\
  Get Toasted Bread       & \phantom{0}0.0  \\
  Kettle Boiling          & 18.0 \\
  Load Dishwasher         & 50.0 \\
  Pack Identical Lunches  & \phantom{0}4.0  \\
  Prepare Coffee          & \phantom{0}0.0  \\
  Pre-Soak Pan            & 34.0 \\
  Rinse Sink Basin        & 28.0 \\
  Scrub Cutting Board     & 48.0 \\
  Searing Meat            & \phantom{0}2.0  \\
  Set Up Cutting Station  & 24.0 \\
  Stack Bowls Cabinet     & 30.0 \\
  Steam in Microwave      & 14.0 \\
  Stir Vegetables         & 18.0 \\
  Store Leftovers in Bowl & \phantom{0}4.0  \\
  Wash Lettuce            & 20.0 \\
  \midrule
  \textbf{Total (16 tasks)} & \textbf{19.0} \\
  \bottomrule
  \end{tabular}
  }
  \end{subtable}
  \hfill
  \begin{subtable}[t]{0.32\textwidth}
  \centering
  \caption{Composite-Unseen}
  \label{tab:rc365-comp-unseen}
  \vspace{1.06em}
  \resizebox{\linewidth}{!}{%
  \begin{tabular}{lc}
  \toprule
  Task & \textbf{RLDX-1 (Ours)} \\
  \midrule
  Arrange Bread Basket    & \phantom{0}6.0  \\
  Arrange Tea             & \phantom{0}2.0  \\
  Bread Selection         & 14.0 \\
  Categorize Condiments   & \phantom{0}0.0  \\
  Cutting Tool Selection  & \phantom{0}9.8  \\
  Garnish Pancake         & 16.0 \\
  Gather Tableware        & \phantom{0}0.0  \\
  Heat Kebab Sandwich     & \phantom{0}0.0  \\
  Make Ice Lemonade       & \phantom{0}8.0  \\
  Pan Transfer            & \phantom{0}0.0  \\
  Portion Hot Dogs        & \phantom{0}2.0  \\
  Recycle Bottles by Type & \phantom{0}6.0  \\
  Separate Freezer Rack   & \phantom{0}0.0  \\
  Waffle Reheat           & \phantom{0}8.0  \\
  Wash Fruit Colander     & 14.0 \\
  Weigh Ingredients       & \phantom{0}4.0  \\
  \midrule
  \textbf{Total (16 tasks)} & \phantom{0}\textbf{5.6} \\
  \bottomrule
  \end{tabular}
  }
  \end{subtable}

\end{table}

\end{document}